%% file: main.tex
\documentclass[conference,onecolumn]{IEEEtran}
\IEEEoverridecommandlockouts
\usepackage{cite}
\usepackage{amsmath,amssymb,amsfonts}
\usepackage{algorithm,algorithmic}
\usepackage{graphicx}
\usepackage{textcomp}
\usepackage{xcolor}
\def\BibTeX{{\rm B\kern-.05em{\sc i\kern-.025em b}\kern-.08em
    T\kern-.1667em\lower.7ex\hbox{E}\kern-.125emX}}
    
\usepackage{amsthm,dsfont,mathrsfs,bm}
\usepackage[caption = false]{subfig}

\usepackage{pgfplots}
\pgfplotsset{compat=newest}
\usetikzlibrary{plotmarks}
\usetikzlibrary{arrows.meta}
\usepgfplotslibrary{patchplots}
\usepackage{grffile}
\usepackage{amsmath}

\DeclareMathOperator{\prob}{Pr}
\DeclareMathOperator{\Expect}{\mathbb{E}}

\newtheorem{assumption}{Assumption}

\newtheorem{remark}{Remark}
    
\begin{document}

\title{Detecting adversaries in Crowdsourcing\\

\thanks{Work in this paper was supported by NSF grants 1901134, 2126052, 2128593 and ARO-STIR grant 00093896.}
}

\author{\IEEEauthorblockN{Panagiotis A. Traganitis and Georgios B. Giannakis}

\IEEEauthorblockA{\textit{Dept. of Electrical and Computer Engineering and Digital Technology Center} \\
\textit{University of Minnesota, MN, USA}\\
emails: \{traga003,georgios\}@umn.edu}
}

\maketitle

\begin{abstract}
Despite its successes in various machine learning and data science tasks, crowdsourcing can be susceptible to attacks from dedicated adversaries. This work investigates the effects of adversaries on crowdsourced classification, under the popular Dawid and Skene model. The adversaries are allowed to deviate arbitrarily from the considered crowdsourcing model, and may potentially cooperate. To address this scenario, we develop an approach that leverages the structure of second-order moments of annotator responses, to identify large numbers of adversaries, and mitigate their impact on the crowdsourcing task. 
The potential of the proposed approach is empirically demonstrated on synthetic and real crowdsourcing datasets.
\end{abstract}

\begin{IEEEkeywords}
Crowdsourcing, Classification, Adversaries, Ensemble learning.
\end{IEEEkeywords}

\section{Introduction}
\label{sec:intro}
Crowdsourcing has emerged as a powerful paradigm for tackling various machine learning, data mining, and data science tasks. Crowdsourcing, via services such as Amazon's Mechanical Turk~\cite{MTurk} and Figure8~\cite{fig8} enlists  inexpensive crowds of human workers, or annotators, to accomplish any given task. 
The focus of much research on crowdsourcing is centered on properly aggregating the noisy annotator labels, to obtain results as close to the ground-truth as possible. This is challenging due to the sparsity of annotator responses and the variability in annotator ability and effort.
To add insult to injury, crowdsourcing is vulnerable to attacks by determined and coordinated adversaries, which provide erroneous responses aiming to reduce the performance of the overall system, or cause misclassification of specific data. 

This paper puts forth a novel method for detecting arbitrary adversaries in crowdsourced classification. As a first step, we first analyze the structure of the correlation matrix of annotator responses, under the popular Dawid and Skene model. Afterwards, a subspace clustering-based approach is developed to split annotators into two groups. Honest and adversarial annotator groups are then distinguished by utilizing some additional side information. In this work, two types of side information are considered: a.) Knowledge of one trusted annotator, or b.) the assumption that the majority ($>50\%$) of annotators are honest. 
Finally, a heuristic approach to prudently aggregate annotator responses is provided.
Compared to other state-of-the-art approaches, the proposed method is based on the more general Dawid and Skene model, and can handle a potentially much larger number of adversaries.

The rest of the paper is organized as follows: Sec.~\ref{sec:problem_statement} states the problem, provides the necessary preliminaries and recaps the prior art; Sec.~\ref{sec:anno_properties} studies the structure of the second-order moments of annotator responses, whereas Sec.~\ref{sec:Adv_ID} introduces our novel method for crowdsourced classification in the presence of adversaries; Sec.~\ref{sec:numerical_tests} presents numerical tests to evaluate the performance of the proposed algorithm, and finally Sec.~\ref{sec:conclusions} concludes with a brief discussion on future research directions.

\noindent\textbf{Notation}. Unless otherwise noted, lowercase bold letters, $\bm{x}$, denote
column vectors, uppercase bold letters, $\mathbf{X}$, represent matrices, and
calligraphic uppercase letters, $\mathcal{X}$, stand for sets. The
$(i,j)$th entry of matrix $\mathbf{X}$ is denoted by
$[\mathbf{X}]_{ij}$; $\text{\rm vec}(\mathbf{X})$ denotes a vector consisting of the stacked columns of $\mathbf{X}$. The Frobenius and nuclear norms of a matrix $\mathbf{X}$ are denoted by $\|\mathbf{X}\|_F$ and $\|\mathbf{X}\|_*$ respectively. The rank of a matrix $\mathbf{X}$ is denoted by $\text{\rm rank}(\mathbf{X})$ and $\text{diag}(\bm{x})$ denotes a diagonal matrix with the vector $\bm{x}$ on its diagonal. $\text{tr}(\mathbf{X})$ denotes the trace of matrix $\mathbf{X}$, that is the sum of the values on its diagonal. $\prob$ denotes probability, or the probability mass function; $\sim$ denotes  "distributed as;"  ${}^\top$ represents transpose; $\text{card}(\mathcal{A})$ denotes the cardinality, i.e. the number of elements, of set $\mathcal{A}$; $\Expect[\cdot]$ denotes
expectation, and $\mathds{1}({\mathcal{A}})$ is the indicator function for the event $\mathcal{A}$, that takes value $1$ when $\mathcal{A}$ occurs, and $0$ otherwise.

\section{Problem statement and preliminaries}
\label{sec:problem_statement}
Consider a dataset consisting of $N$ independent and identically distributed (i.i.d.) data $\{x_n\}_{n=1}^{N}$ each belonging to one of $K$ possible classes with corresponding labels $\{y_n\}_{n=1}^{N}$, e.g. $y_n = k$ if $x_n$ belongs to class $k$. Class prior probabilities are collected in $\bm{\pi}:= [\pi_1,\ldots,\pi_K]^{\top} = [\prob(y_n= 1),\ldots,\prob(y_n= K)]^{\top}.$ An ensemble of $M$ annotators or workers observe $\{x_n\}_{n=1}^{N}$, and provide noisy estimates of labels, with $g_m(x_n)\in\{1,\ldots,K\}$ denoting the label assigned to the $n$-th datum by annotator $m$. When an annotator does not provide a response for a datum $x_n$, we encode this by $g_m(x_n) = 0.$
% in the $M\times N$ matrix $\mathbf{F} = [\mathbf{f}(x_1),\ldots,\mathbf{f}(x_N)]$, that has entries $[\mathbf{F}]_{mn} = f_m(x_n)$, and whose columns $\mathbf{f}(x_n)$, collect annotator responses for datum $x_n$. 
Given only the annotator responses $\{g_m(x_n), m=1,\ldots,M\}_{n=1}^{N}$, \emph{crowdsourced classification} seeks to properly aggregate information contained in annotator responses and estimate the ground-truth labels of the data $\mathbf{y} := [y_1,\ldots,y_N]^\top$.

Next, we outline a popular probabilistic model for tackling the aforementioned crowdsoucring task, that this work is based on.

\subsection{The Dawid and Skene model for crowdsourcing}
\label{ssec:dawidandskene}
The Dawid and Skene (DS) model \cite{dawid1979maximum} asserts that annotators have constant behavior and are conditionally independent given the (unknown) true label of a datum $y_n$, that is their errors are independent. Note that, this is reminiscent of a Naive Bayes model, conditioned on the true labels $y_n$. An annotators response $g_m(x_n)$ for a datum $x_n$, depends only on that datum and only through its label $y_n$.  Further, under the DS model annotators are characterized by a so-called \emph{confusion matrix}, that captures the statistical behavior of an annotator when presented with a datum from each class. For an annotator $m$ the $K\times K$ confusion matrix is denoted by $\mathbf{H}_m$, and has entries $h_{m,k,c}:=[\mathbf{H}_m]_{k,c} = \prob(g_m(x_n) = k |y_n = c).$ Clearly, entries of $\mathbf{H}_m$ are non-negative and its columns sum up to $1$. Since responses of different annotators per datum $n$ are presumed conditionally independent, given the ground-truth label $y_n$, the joint pmf of annotator responses for datum $x_n$ is $\prob\left(g_1(x_n) = k_1,\ldots,g_M(x_n)=k_M | y_n = c\right) 
= \prod_{m=1}^{M} \prob\left(g_M(x_n) = k_m | y_n = c\right) = \prod_{m=1}^{M}h_{m,k_m,c}. $

If annotator confusion matrices and class priors are known, the label of datum $n$ can be estimated using a maximum a posteriori (MAP) classifier, as 
\begin{align}
	\label{eq:max_label_bayes_log}
	\hat{y}_n 
	& = \underset{{c\in\{1,\ldots,K\}}}{\arg\max} \log\pi_c + \sum_{m=1}^{M}\log(h_{m,g_m(x_n),c}) 
\end{align}
where we used the conditional independence of the annotators, and the monotonicity of the logarithm. In realistic crowdsourcing scenaria, however, confusion matrices $\{\mathbf{H}_m \}_{m=1}^{M}$ and class priors $\bm{\pi}$ are unknown and have to be estimated.

\subsection{Crowdsourcing with adversaries}
\label{ssec:crowd_w_adv}
Suppose now, that a subset $\mathcal{H}\subseteq\mathcal{M}=\{1,\ldots,M\}$, of $M_\mathcal{H} :=\text{card}(\mathcal{H})$ annotators are honest, and a subset $\mathcal{A}\subset\mathcal{M}$, of $M_\mathcal{A} :=\text{card}(\mathcal{A})$ are adversaries. The adversaries in $\mathcal{A}$ also observe the data $\{x_n\}_{n=1}^{N}$ and seek to undermine the crowdsourcing task.
In order to ensure robust estimation of the ground-truth labels, one has to \emph{detect the presence of these adversaries and take mitigating steps}. 
The following assumptions hold throughout the rest of the paper:
% \begin{assumption}
% \label{as:constant_behavior}
% The behavior of all annotators remains invariant over the observation window.
% \end{assumption}
\begin{assumption}
\label{as:honest_ds}
Honest annotators adhere to the Dawid and Skene model; that is, given the ground-truth label $y_n$ of a datum $x_n$, the responses of annotators are conditionally independent.
% \begin{align}
% 	   & \prob\left( \left\{ g_m(x_n) = k_m \right\}_{m\in\mathcal{H}}     \Bigg| y_n=k\right) = \prod_{m\in\mathcal{H}}\prob\left(g_m(x_n) = k_m | y_n=k\right).\notag
% \end{align}
Additionally, honest annotators are better than random.
\end{assumption}
\begin{assumption}
\label{as:num_honest}
The number of  honest annotators $\text{\rm card}(\mathcal{H})$ is strictly greater than $K^2$, and they are distinct. 
\end{assumption}
\begin{assumption}
\label{as:adversaries}
Adversarial annotators observe $\{x_n\}_{n=1}^{N}$ and deviate arbitrarily from the Dawid and Skene model.
\end{assumption}

Under the aforementioned assumptions, in this work we seek to \emph{identify adversarial annotators} and if possible \emph{mitigate} their impact on the crowdsourcing classification task, using only the available annotator responses.

Assumptions~\ref{as:honest_ds} is fairly standard in crowdsourcing and enables estimating annotator parameters and label aggregation. As will be shown later, Assumption~\ref{as:num_honest} is necessary in this context for distinguishing honest annotators from adversaries.
Further, this assumption also indicates that the proposed approach can, in principle, tolerate up to $M_\mathcal{A} = M - K^2$ adversaries, which depending on $M$ and $K$, may be much larger than the $M/2$ number of adversaries that is allowed by competing alternatives. Nevertheless, the number of tolerated adversaries will depend on the additional information employed in Sec.~\ref{sec:Adv_ID}. Assumption~\ref{as:adversaries} implies that adversaries take into consideration only the data, and not the responses of honest annotators. It does not place any further restriction on the behavior of the adversaries, and suggests that their behavior is captured by an \emph{unknown} conditional pmf 
\begin{align*}
    p_{\mathcal{A}}:=\prod_{n=1}^{N}\prob\left( \{ g_m(x_n) = k_{m,n} \}_{m\in\mathcal{A}} \Bigg| \{y_{n'}= k_{n'}\}_{n'=1}^{N} \right).
\end{align*}
Note here that adversaries are not necessarily conditionally independent with each other, and their responses may depend on all observed data. However, since they only observe the available data, they are considered conditionally independent from the honest workers.

\subsection{Prior art}
\label{ssec:prior_art}
\noindent\textbf{Crowdsourced classification.} 
The simplest method for aggregating crowdsourced labels is majority voting, where the estimated label for a specific data point is the one most annotators agree on. This however, assumes that all annotators are of equal ability, which is, in many cases, unrealistic. 
The seminal paper of Dawid and Skene \cite{dawid1979maximum} proposed the aforementioned model and introduced an  expectation maximization (EM) algorithm for estimating annotator confusion matrices and class priors, that is guaranteed to converge to a local optimum. Recent spectral methods use second- and third-order moments of annotator responses to infer confusion matrices and are often used to initialize the EM algorithm~\cite{jaffe2015estimating,zhang2014spectral,traganitis2018,kargas_xiao}, while \cite{minimax_crowd} finds confusion matrices by maximizing the entropy of the joint pmf of annotators, labels and their responses. Other works, advocate simpler, but less expressive models, such as the "one-coin" model, where each annotator is characterized by a single parameter~\cite{Ghosh,EigenRatio,KOS}. The work of \cite{pgd} considered crowdsourced classification under the one-coin model as a rank-one matrix completion problem. A bayesian approach to crowdsourced classification was advocated in \cite{BCC_Kim}, while other works take advantage of the data structure to aid the label aggregation task \cite{Rodrigues2014,traganitis_seq_networked}. 

\noindent\textbf{Crowdsourcing under adversarial attacks.}
Regarding adversarial attacks in crowdsourcing, \cite{raykar_spammers} modified the EM algorithm of \cite{dawid1979maximum} to detect and eliminate spammers during the label aggregation phase, whereas \cite{traganitis_spammers} proposed a spectral algorithm for detecting spammers before the aggregation phase. In the binary classification setting, \cite{adv_unrel_jmlr} proposed a penalty based algorithm for detecting adversaries, and  \cite{crowd_arbitrary_adv} considers arbitrary adversaries under the one-coin model. Recently, \cite{MMSR} introduced a rank-$1$ matrix completion algorithm for aggregating labels in the presence of adversaries, under the one-coin model. However, the three aforementioned works assumed that most annotators ($>50\%$) are honest.
Compared to current adversarial crowdsourcing approaches, this work introduces an algorithm that is based on the general Dawid and Skene model, and can potentially detect a large number of adversaries.

\section{Annotator Correlation}
\label{sec:anno_properties}
Based on the Dawid and Skene model of Sec.~\ref{ssec:dawidandskene}, we will first examine the structure of the second-order moments of honest and adversarial annotator responses. As annotator responses are categorical variables, the  measure of correlation considered here is the probability of agreement, or agreement rate between two annotators $\sigma_{m,m'}:=\prob(g_m(x_n) = g_{m'}(x_n)), m,m'\in\mathcal{M}$. For the remainder of this section, we will also assume without loss of generality, that the first $M_\mathcal{H}$ annotators are honest, and the remaining are adversarial.

\subsection{Correlation of honest annotators}
\label{ssec:hon_properties}
Let $\mathbf{g}_m(x_n)$ denote the response of annotator $m$ when observing datum $x_n$, in ``one-hot'' format, that is, if $g_m(x_n) = k$ then $\mathbf{g}_m(x_n) = \mathbf{e}_k$, where $\mathbf{e}_k$ denotes the canonical $K\times 1$ vector that has a one in its' $k$-th entry and zeroes elsewhere.

Invoking the law of total probability, the assumed conditional independence of annotators, and the definitions of Sec.~\ref{ssec:dawidandskene}, the $K\times K$ co-occurrence matrix between annotators $m,m'\in\mathcal{H}$ is \cite{traganitis2018}
\begin{align}
     \mathbf{R}_{m,m'} &:= \Expect[\mathbf{g}_m(x_n)\mathbf{g}_{m'}(x_n)] 
     = \sum_{k,k'}^{K}\mathbf{e}_k\mathbf{e}_{k'}\prob(g_m(x_n) = k,g_{m'}(x_n) = k')  \notag\\
     & = \sum_{k,k'}^{K}\mathbf{e}_k\mathbf{e}_{k'}\sum_{c=1}^{K}h_{m,k,c}h_{m',k',c}\pi_c   = \mathbf{H}_m\text{diag}(\bm{\pi})\mathbf{H}_{m'}^{\top}.
\end{align}
From $\mathbf{R}_{m,m'}$, the probability of agreement $\sigma_{m,m'}$ between annotators $m,m'\in\mathcal{H}$ is 
\begin{align}
    \sigma_{m,m'} & = \prob(g_m(x_n) = g_{m'}(x_n)) 
    = \sum_{k=1}^{K}\prob(g_m(x_n) = g_{m'}(x_n) = k) 
      = \text{tr}\left(\mathbf{R}_{m,m'}\right) = \text{tr}\left( \mathbf{H}_m\text{diag}(\bm{\pi})\mathbf{H}_{m'}^{\top} \right).
\end{align}
Using the properties of the trace $\sigma_{m,m'}$ can be decomposed into 
\begin{align}
     \sigma_{m,m'} & = \text{tr}(\mathbf{H}_m\text{diag}(\bm{\pi})^{1/2}\text{diag}(\bm{\pi})^{1/2}\mathbf{H}_{m'}^{\top})      \label{eq:kappa_decomp}
\\ & =  \text{vec}(\text{diag}(\bm{\pi})^{1/2}\mathbf{H}_{m}^{\top})^{\top}\text{vec}(\text{diag}(\bm{\pi})^{1/2}\mathbf{H}_{m'}^{\top}) = \bm{v}_m^{\top}\bm{v}_{m'} \notag
\end{align}
where $\bm{v}_m:= \text{vec}(\text{diag}(\bm{\pi})^{1/2}\mathbf{H}_{m}^{\top})$ is a $K^2\times 1$ vector. Eq.~\ref{eq:kappa_decomp} in turn implies that the $M_{\mathcal{H}}\times M_{\mathcal{H}}$ agreement matrix between honest annotators $\mathbf{\Sigma}_{\mathcal{H}}$, with entries $[\mathbf{\Sigma}_{\mathcal{H}}]_{m,m'} = \sigma_{m,m'}, m,m'\in\mathcal{H}$ has a low-rank plus diagonal form 
\begin{align}
\label{eq:kappa_mat_honest}
\mathbf{\Sigma}_\mathcal{H} =\mathbf{C}_\mathcal{H} + \mathbf{I}_{\mathcal{H}} =  \mathbf{V}^{\top}\mathbf{V} + \mathbf{I}_{\mathcal{H}},
\end{align}
where $\mathbf{I}_{\mathcal{H}}$ denotes the identity matrix of appropriate dimension, $\mathbf{C}_\mathcal{H}:=\mathbf{V}^{\top}\mathbf{V}$, and $\mathbf{V}:=[\bm{v}_1,\ldots,\bm{v}_{M_\mathcal{H}}]$ is a $K^2\times M$ matrix. 
Finally, Assumption~\ref{as:num_honest} asserts that $\text{rank}(\mathbf{C}_\mathcal{H}) = K^2.$

\subsection{Correlation between honest and adversarial annotators}
\label{ssec:moments_adv}
As mentioned in Sec. \ref{ssec:crowd_w_adv} the behavior of adversarial annotators  is captured by an \emph{unknown} joint pmf $p_{\mathcal{A}}$. Despite $p_{\mathcal{A}}$ being unknown, the conditional independence between the group of adversaries and honest annotators enables characterization of their cross-moments. Based on this conditional independence
[cf. Sec.\ref{ssec:crowd_w_adv}], the co-occurrence matrix between an honest annotator $m\in\mathcal{H}$ and an adversary $m'\in\mathcal{A}$ is
\begin{align}
    \mathbf{R}_{m,m'} & = \Expect[\mathbf{g}_m(x_n)\mathbf{g}_{m'}(x_n)] 
     =  \sum_{k}^{K}\sum_{k'=1}^{K}\mathbf{e}_k \mathbf{e}_{k'} \prob(g_m(x_n) = k,g_{m'}(x_n) = k') \notag\\
    & =   \sum_{k=1}^{K}\sum_{k'=1}^{K}\mathbf{e}_k \mathbf{e}_{k'}\sum_{c_n=1}^{K}\pi_{c_n}h_{m,k,c}\tilde{h}_{m',k',n}
     = \mathbf{H}_m\text{diag}(\bm{\pi})\tilde{\mathbf{H}}_{m'}^{\top}. \label{eq:xcorr_hon_adv}
\end{align}
where we have used the law of total probability, the fact that data are i.i.d., and defined 
$\tilde{h}_{m,k,n} := [\tilde{\mathbf{H}}_m]_{k,n} = \sum_{\mathbf{c}_{-n}}\prob(g_m(x_n) = k | \mathbf{y} = \mathbf{c})\prod_{j\neq n}\prob(y_j = c_j)$. In addition, $\mathbf{c}_{-n}$ is an $N-1\times 1$ vector containing $\{c_{j}\}_{j=1}^{N}$ except $c_n.$ It is worth noting that, for the purposes of this work, we are not interested in estimating $\tilde{\mathbf{H}}_{m}$, but are merely employing them to discover the properties of the annotator agreement matrix.

Then, \eqref{eq:xcorr_hon_adv} yields $\sigma_{m,m'}$ between an honest annotator $m\in\mathcal{H}$ and an adversarial one $m'\in\mathcal{A}$ as
\begin{align}
& \sigma_{m,m'}  = \text{tr}\left(\mathbf{H}_m\text{diag}(\bm{\pi})\tilde{\mathbf{H}}_{m'}^{\top}\right) = \bm{v}_m^{\top}\tilde{\bm{u}}_{m'}
\end{align}
with $\bm{v}_m$ as defined in Sec. \ref{ssec:hon_properties} and $\tilde{\bm{u}}_{m'} := \text{vec}\left(\text{diag}(\bm{\pi})^{1/2}\tilde{\mathbf{H}}_{m'}^{\top}\right)$. The agreement rate between all honest and adversarial annotators is then captured in the $M_{\mathcal{H}}\times M_{\mathcal{A}}$ matrix $\mathbf{C}_{\mathcal{H},\mathcal{A}} = \mathbf{C}_{\mathcal{A},\mathcal{H}}^{\top}$, with entries $[\mathbf{C}_{\mathcal{H},\mathcal{A}}]_{m,m'} = \bm{v}_m^{\top}\tilde{\bm{u}}_{m'}$ for $m\in\mathcal{H},m'\in\mathcal{A}$. Thus, $\mathbf{C}_{\mathcal{H},\mathcal{A}} = \mathbf{V}^{\top}\tilde{\mathbf{U}}$, where $\tilde{\mathbf{U}} := [\tilde{\bm{u}}_1,\ldots,\tilde{\bm{u}}_{M_{\mathcal{A}}}]$, and  $\text{rank}(\mathbf{C}_{\mathcal{H},\mathcal{A}}) \leq K^2$.

Bringing it all together, the $M\times M$ agreement matrix between all annotators, honest and adversarial, $\mathbf{\Sigma}$ has the following block form
\begin{align}
\label{eq:XCov}
\mathbf{\Sigma} = \mathbf{C} + \mathbf{I} = 
   \left[\begin{array}{c|c} \mathbf{C}_\mathcal{H}  & \mathbf{C}_{\mathcal{H},\mathcal{A}} \\ \hline \mathbf{C}_{\mathcal{A},\mathcal{H}} & \mathbf{C}_{\mathcal{A}} \end{array}\right] + 
   \left[\begin{array}{c c} \mathbf{I}_\mathcal{H} & \\ &\mathbf{I}_{\mathcal{A}} \end{array}\right]
\end{align}
where  the $M_{\mathcal{A}}\times M_{\mathcal{A}}$ matrix $\mathbf{C}_{\mathcal{A}}$ denotes the  correlation between adversaries, $\mathbf{I}_{\mathcal{A}}$ is a $M_{\mathcal{A}}\times M_{\mathcal{A}}$ identity  matrix, and $\mathbf{I}$ is a $M\times M$ identity matrix.
Note that $[\mathbf{C}_{\mathcal{H}}, \mathbf{C}_{\mathcal{H},\mathcal{A}}]^{\top} = [\mathbf{V}^{\top}\mathbf{V}, \mathbf{V}^{\top}\tilde{\mathbf{U}}]^{\top} = \left( \mathbf{V}^{\top}[\mathbf{V}, \tilde{\mathbf{U}}]\right)^{\top}.$ Thus, the $M_{\mathcal{H}}$ columns of $\mathbf{C}$ corresponding to honest workers will be of rank $K^2$, as long as $\text{rank}([\mathbf{V}, \tilde{\mathbf{U}}]) = K^2$. 
Finally, since $\text{\rm rank}\left( [\mathbf{C}_{\mathcal{H}}, \mathbf{C}_{\mathcal{H},\mathcal{A}}]^{\top}\right) \leq K^2,$ $\mathbf{C}$ is a rank deficient matrix.

\section{Identifying adversaries}
\label{sec:Adv_ID}
In this section, we will take advantage of the structure of the annotator agreement matrix, specifically $\mathbf{C}$ [cf. \eqref{eq:XCov}], in order to develop a method to distinguish honest workers from adversaries in crowdsourcing. In a nutshell, the proposed method seeks annotators whose agreement matrix fits the low-rank model discussed in the previous section. These annotators are deemed honest, while the rest are considered adversaries.

\subsection{Estimating $\mathbf{C}$}
In the crowdsourcing setup, we do not have access to $\mathbf{C}$, or even $\mathbf{\Sigma}$. Nevertheless, the empirical agreement matrix $\hat{\mathbf{\Sigma}}$ can be readily computed from $\{g_m(x_n)\}_{m=1,n=1}^{M,N}$. 
Given the $M\times M$ empirical agreement matrix $\hat{\mathbf{\Sigma}}$, we can decouple the diagonal matrix $\mathbf{D}$ and the rank deficient matrix $\mathbf{C}$, using robust principal component analysis (RPCA~\cite{rpca_candes}) or robust matrix completion (RMC~\cite{rmc}) methods, that is
\begin{align}
\label{eq:rpca}
    \{\hat{\mathbf{C}},\hat{\mathbf{S}} \} = &\arg\min_{\mathbf{C},\mathbf{S}} \|\mathbf{C}\|_* + \lambda\|\text{vec}(\mathbf{S})\|_1 \\
    & \text{subject to }  \mathbf{\Omega}\circ\hat{\mathbf{\Sigma}} = \mathbf{\Omega}\circ\left(\mathbf{C} + \mathbf{S}\right) \notag
\end{align}
where $\mathbf{S}$ is a sparse $M\times M$ matrix, $\lambda>0$,  $\mathbf{\Omega}$ is a binary $M\times M$ matrix, whose entries are equal to $1$ if the corresponding entry of $\hat{\mathbf{\Sigma}}$ is observed and $0$ otherwise, and $\circ$ denotes the Hadamard (element-wise) matrix product. Note that,  unobserved entries of $\hat{\mathbf{\Sigma}}$ are a common occurrence in crowdsourcing, since not all annotators provide responses for the same subsets of data. The nuclear norm $\|\mathbf{C}\|_*$ in the objective promotes low rank solutions for ${\mathbf{C}}$, while $\|\text{\rm vec}(\mathbf{S})\|_1$ encourages sparse solutions for $\hat{\mathbf{S}}$. The parameter $\lambda$ trades off the low rank of $\mathbf{C}$ and the sparsity of $\mathbf{S}.$ As the identity matrix $\mathbf{I}$ [cf. \eqref{eq:XCov}] generally does not adhere to the low rank structure of $\mathbf{C}$, we expect it to be captured in $\mathbf{S}$. 
Additionally, $\mathbf{S}$ may capture any spurious correlations between annotators.

The optimization problem in \eqref{eq:rpca} is a convex problem that can be solved using off-the-shelf solvers, such as CVX~\cite{cvx}. Furthermore, many efficient algorithms have been developed to tackle this problem, such as augmented lagrangian, and proximal gradient methods among others. Interested readers are directed to \cite{rmc,rpca_survey,lrslibrary2015} and references therein. Recovery of $\mathbf{C}$ and $\mathbf{S}$ with $\lambda = 1/\sqrt{\alpha M}$ was demonstrated empirically and theoretically in \cite{rpca_candes} where $\alpha$ is the percentage of observed entries in $\hat{\mathbf{\Sigma}}$, and this is the value adopted for the remainder of this paper.

\begin{remark}
 While \eqref{eq:XCov} is reminiscent of the covariance model present in Factor Analysis~\cite{fa_bertsimas}, here we opted for RPCA algorithms as they can tackle sparse corruptions that may be present in $\hat{\mathbf{\Sigma}}.$
\end{remark}
\subsection{Clustering annotators}
\label{ssec:clustering_annos}
With $\hat{\mathbf{C}}$ at hand, we now turn our attention to the task of detecting adversarial annotators. This task is equivalent to detecting the honest annotators, by identifying  the columns of $\mathbf{C}$ corresponding to honest workers.
Recall that, under Assumption~\ref{as:num_honest} $[\mathbf{C}_{\mathcal{H}}, \mathbf{C}_{\mathcal{H},\mathcal{A}}]^{\top}$, will form a low dimensional subspace of dimension at most $K^2$.
This prompts us to look into subspace clustering approaches, to segment the annotators into two groups. Subspace clustering methods are designed to group data drawn from a union of subspaces~\cite{vidal2010tutorial}. Several algorithms have been developed to tackle the subspace clustering task~\cite{elhamifar2013SSC,liu2013robust,ORGEN,asc_tsakiris}, but here we will focus on the so-called self-representation approaches.
Self-representation based approaches  take advantage of the fact that a point lying in a subspace can be written as a linear combination of other points in the same subspace. The coefficients of these linear combinations, denoted here by the $M\times M$ matrix $\mathbf{Z}$ are then typically extracted by solving the following optimization problem
\begin{align}
\label{eq:main_SC}
    \min_{\mathbf{Z}} \|\hat{\mathbf{C}} - \hat{\mathbf{C}}\mathbf{Z}\|_F^2 +  \rho r(\mathbf{Z}),
\end{align}
where $\rho >0$, and $r(\mathbf{Z})$ is a regularization function that promotes certain desirable properties on $\mathbf{Z}$, such as sparsity or low rank. Further, efficient algorithms have been proposed to solve \eqref{eq:main_SC} \cite{dyerOMP,NSN,efficientssc,sketchedSC}.
After obtaining $\mathbf{Z}$ from \eqref{eq:main_SC}, spectral clustering~\cite{spectralclustering} is performed on $|\mathbf{Z}| + |\mathbf{Z}^{\top}|$, with $|\cdot|$ denoting element-wise absolute value, to obtain cluster assignments. In this work, we opt for the Elastic Net subspace clustering algorithm \cite{ORGEN}, where $r(\mathbf{Z}) = \rho_2\|\text{\rm vec}(\mathbf{Z})\|_1 + \frac{1-\rho_2}{2}\|\mathbf{Z}\|_F^2$, and $\rho_2 = 0.95$. 

Let $\mathcal{C}_1$ and $\mathcal{C}_2$ denote the annotator indices corresponding to the two clusters that resulted from subspace clustering of $\hat{\mathbf{C}}$. In order to categorize the two formed annotator groups into honest $\hat{\mathcal{H}}$ and adversarial $\hat{\mathcal{A}}$, some additional information is required.
This information may be similar to what most prior works consider, that is, most annotators ($>50\%$) are honest~\cite{adv_unrel_jmlr,crowd_arbitrary_adv,MMSR}. In such a case, the annotators deemed as honest are the ones forming the largest group, and are collected in $\hat{\mathcal{H}}$. The annotators deemed as adversaries are collected in $\hat{\mathcal{A}} = \mathcal{M}\setminus\hat{\mathcal{H}}.$ Such an approach can tolerate up to $0.5M$ adversaries. Another type of side information, may be knowledge that one (or more) specific annotator $m_H$ is trusted, or honest. In this case, the set of annotators deemed as honest is the one that contains the index $m_H$. This type of side information has the potential to allow for greater numbers of adversaries if annotators are grouped correctly. 
Both types of side information mentioned here will be tested in Sec.~\ref{sec:numerical_tests}.

Finally, let $\hat{\mathbf{\Sigma}}_{\mathcal{C}_i}$, $\hat{\mathbf{C}}_{\mathcal{C}_i}$, $\mathbf{\Omega}_{\mathcal{C}_i}$ denote the submatrices of $\hat{\mathbf{\Sigma}}$, $\hat{\mathbf{C}}$, and $\mathbf{\Omega}$ respectively, formed by selecting the rows and columns in $\mathcal{C}_i$. The parameter $\rho$ in \eqref{eq:main_SC} is tuned using grid search to achieve the smallest $\varepsilon(\mathcal{C}_i) = \frac{1}{\text{\rm card}(\mathcal{C}_i)}\| \tilde{\mathbf{\Omega}}_i\circ(\hat{\mathbf{\Sigma}}_{\mathcal{C}_i} - \mathbf{Q}_i\mathbf{\Lambda}_i\mathbf{Q}_i^{\top}) \|_F^2/\|\mathbf{\Omega}_i\circ\hat{\mathbf{\Sigma}}_{\mathcal{C}_i}\|_F^2$, where  $\mathbf{\Lambda}_i$ and $\mathbf{Q}_i$  contain the $K^2$ largest eigenvalues and corresponding eigenvectors of $\hat{\mathbf{C}}_{\mathcal{C}_i}$, respectively, and $\tilde{\mathbf{\Omega}}_{\mathcal{C}_i}$ is equal to $\mathbf{\Omega}_{\mathcal{C}_i}$, but with zeroes in the diagonal.

\begin{algorithm}[tb]
\caption{Crowdsourcing with adversaries}
\label{alg:main_alg}
\begin{algorithmic}[1]
\STATE {\bfseries Input:} Annotator correlation matrix $\hat{\mathbf{\Sigma}}$, number of classes $K$, $\rho$.
\STATE {\bfseries Output:} $\hat{\mathbf{y}}$, estimated annotator groups $\hat{\mathcal{H}},\hat{\mathcal{A}}$
\STATE Extract $\hat{\mathbf{C}}$ from $\hat{\mathbf{\Sigma}}$ using \eqref{eq:rpca}.
\STATE Cluster annotators by solving \eqref{eq:main_SC}, obtain two groups of annotator indices $\mathcal{C}_1,\mathcal{C}_2.$
\STATE Determine sets of  honest $\hat{\mathcal{H}}$ and adversarial workers $\hat{\mathcal{A}} = \mathcal{M}\setminus\hat{\mathcal{H}}$ using side information [cf. Sec.~\ref{ssec:clustering_annos}].
% \STATE Set the set of honest workers as $\hat{\mathcal{H}} = \mathcal{C}_i$, where $i = \arg\min\varepsilon(\mathcal{C}_i),$ and the set of adversarial ones as $\hat{\mathcal{A}} = \mathcal{M}\setminus\hat{\mathcal{H}}.$
\STATE Aggregate labels of adversaries $\{g_m(x_n)\}_{n=1,m\in\hat{\mathcal{A}}}^{N}$. Denote fused labels as $\{\hat{t_n}\}_{n=1}^{N}$. 
\STATE Aggregate labels of honest workers and fused adversary labels $\{ g_m(x_n)\}_{n=1,m\in\hat{\mathcal{H}}}^{N}\cup\{\hat{t}_n\}_{n=1}^{N}$ to form the final estimated data labels $\{\hat{y}_n\}_{n=1}^{N}.$
\end{algorithmic}
\end{algorithm}

\subsection{Aggregating labels}
Upon grouping annotators into honest and adversarial the final step involves aggregating the noisy labels.
% One potential way to address this, is to prune all annotators deemed as adversaries, effectively disregarding their labels, and aggregate the labels provided by the remaining annotators. Nevertheless, adversaries may still provide some useful label information, which would be lost by pruning them. 
To extract any label information that may be present in the adversarial responses, a two-step heuristic label aggregation approach is outlined below.

First, responses from annotators deemed adversarial (with indices in $\hat{\mathcal{A}}$) are aggregated using standard crowdsourcing techniques, e.g. the EM algorithm of \cite{dawid1979maximum}. This yields the aggregated labels $\{\hat{t}_n \}_{n=1}^{N}$, with $\hat{t}_n \in\{1,\ldots,K\}.$ This step approximates the unknown pmf of adversaries as a Dawid and Skene model, and condenses their effect into the estimated $\hat{t}_n$'s.
Second, to produce the final aggregated labels the responses of annotators deemed honest (with indices in $\hat{\mathcal{H}}$) alongside $\{\hat{t}_n \}_{n=1}^{N}$, from the first step, are aggregated using standard crowdsourcing algorithms, to produce the final estimated labels $\{\hat{y}_n\}_{n=1}^{N}$. 
In order to minimize the effect of misclassified annotators from the clustering stage, estimated labels are provided for data that have received a response from at least $K$ different annotators.
The entire algorithm for detecting adversaries and aggregating labels is tabulated in Alg.~\ref{alg:main_alg}.

\section{Numerical tests}
\label{sec:numerical_tests}
The performance of the proposed algorithm is validated in this section using synthetic and real datasets. Alg. \ref{alg:main_alg} is compared against majority voting, denoted as \emph{MV}, the EM algorithm of \cite{dawid1979maximum}, denoted as \emph{DS}, the matrix completion method for one-coin models of \cite{pgd}, denoted as \emph{PGD}, and the state-of-the-art method for adversarial crowdsourcing of \cite{MMSR}, denoted as \emph{MMSR}. For these numerical tests, Alg.~\ref{alg:main_alg} uses \emph{DS} at the aggregation stage. When considering that most annotators are honest, we will denote our approach as \emph{Alg. \ref{alg:main_alg} - H + DS}, whereas when we consider \emph{one} trusted annotator, we will denote our approach as \emph{Alg. \ref{alg:main_alg} - TA + DS}. For \emph{Alg. 1 - TA + DS}  one randomly chosen honest worker is deemed trusted. The parameter $\rho$ of Alg.~\ref{alg:main_alg} is selected using grid search [cf. Sec.~\ref{ssec:clustering_annos}] from the set $\{1.1, 2, 5, 10, 20, 100, 500, 800, 1000\}$. In all cases, \emph{DS} is initialized using \emph{MV}. \emph{MV}, \emph{DS}, and \emph{PGD} are chosen to represent non-adversarially robust crowdsourcing algorithms, since, most algorithms not designed for adversaries exhibit similar behavior in the setup considered here, as is shown from the experiments of \cite{MMSR}.
All algorithms are compared in terms of classification accuracy, that is, the percentage of correctly classified data: $\text{Accuracy} = \frac{1}{N}\sum\mathds{1}(\hat{y}_n = y_n)$. For the synthetic data tests, Alg. \ref{alg:main_alg} is also evaluated on the performance of detecting adversaries, in terms of sensitivity (a.k.a. true positive rate, or recall) and specificity (true negative rate) \cite{powers2011evaluation}, as well as clustering accuracy, that is, how accurately the groups of annotators are recovered. Such a comparison cannot be done with \emph{MMSR} as it does not assign a label to adversaries, rather it appropriately weighs their responses. All algorithms were evaluated using MATLAB~\cite{MATLAB:2020} and all results represent the averages of $20$ runs, and shaded areas indicate standard deviation around the average. 

In this section, adversaries adopt strategies similar to the ones used in the numerical tests of \cite{MMSR}. Per run, $M_{\mathcal{A}} = \lfloor Mp_{\rm adv} \rfloor$ annotators are randomly selected to act as adversaries, with $p_{\rm adv}$ denoting the percentage of adversaries. A percentage of $p_{\rm corr}$ data are randomly selected to be corrupted by the adversaries. In this work, two types of adversaries are considered, and are referred to as \emph{Type A} and \emph{Type B} adversaries. \emph{Type A} adversaries provide the same wrong response for the corrupted data, and for the remaining  $1-p_{\rm corr}$ percentage of data they provide the ground-truth label. Note here that \emph{Type A} adversaries are the ones considered in \cite{MMSR}. \emph{Type B} adversaries also provide the same wrong response for the corrupted data, however, they adopt the Dawid and Skene model for the remaining $1-p_{\rm corr}$ percentage of data, that is, their responses are conditionally independent: each adversary $m$ generates a random confusion matrix $\mathbf{H}_m$ satisfying Assumption~\ref{as:honest_ds}, and for data that are not going to be corrupted $g_m(x_n)$'s are generated according to this $\mathbf{H}_m$. Both types of adversaries are colluding in this setup, however Type B adversaries attempt to mimic the behavior of honest ones. Finally, each adversary provides a response for each datum with probability $p_{\rm obs}.$

\subsection{Synthetic data}
Here, a synthetic dataset with $M=60$ annotators, $K=3$ classes and $N=5,000$ data points was randomly generated; labels $y$ were drawn i.i.d. from $\bm{\pi} = 1/K\mathbf{1}$, and honest annotator confusion matrices $\{\mathbf{H}_m\}$ were randomly generated to satisfy Asumption~\ref{as:honest_ds}. Using these confusion matrices honest annotator responses were generated, i.e. if $y_n = k$ $g_m(x_n)\sim \bm{h}_{m,k}$, with $\bm{h}_{m,k}$ denoting the $k$-th column of $\mathbf{H}_m$. All annotators (honest and adversarial) provide responses for data with probability $p_{\rm obs} = 0.2.$  First, \emph{Type A} adversaries are considered. Fig.~\ref{fig:synth_data_per_adv}  shows the results for this synthetic dataset as the percentage of adversaries $p_{\rm adv}$ varies. In this figure, the percentage of corrupted data is $p_{\rm corr} = 0.5.$ As the number of adversaries increases, overall classification performance drops. The classification accuracy \emph{DS} and \emph{PGD} drops quickly to $0.5$, whereas \emph{MV} and \emph{MMSR} decline more gracefully, and provide better classification performance than \emph{DS} and \emph{PGD} up until $50\%$ of the annotators are adversaries. For up to $p_{\rm adv} = 0.5$,  \emph{Alg. 1 - H} is the more robust of all algorithms. Impressively \emph{Alg. 1 - TA} outperforms other algorithms for most percentages of adversaries, even though only \emph{one} annotator is known a priori to be honest.   Fig.\ref{fig:synth_data_p_adv} shows results for the same dataset, but now the percentage of adversaries is fixed $p_{\rm adv}=0.3$, and the percentage of corrupted data $p_{\rm corr}$ varies. As the number of corrupted data increases, the performance of \emph{DS} and \emph{MV} decreases, while interestingly the performance of \emph{MMSR} increases after approximately $60\%$ are corrupted. We conjecture that large numbers of corrupted data, under this adversarial model, enable their detection, as their corresponding correlation increases. The performance of \emph{Alg. \ref{alg:main_alg}} remains almost constant throughout, with both variants achieving high classification accuracy. In both scenaria, increasing number of adversaries and increasing number of corrupted data, adversaries are almost perfectly detected.
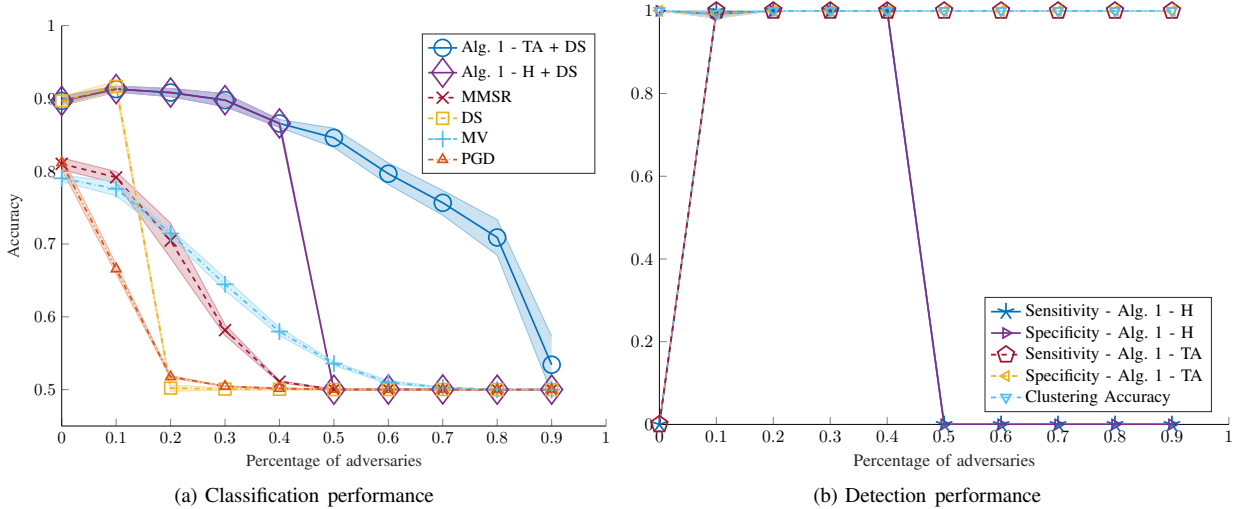
\begin{figure}[tb]
\centering{
\subfloat[Classification performance]{
\resizebox{0.45\columnwidth}{!}{\input{fig/tikz/synthetic_data/per_adv_synth_data_acc_groundtruth_shaded_acc.tikz}}}
\subfloat[Detection performance]{
\resizebox{0.45\columnwidth}{!}{\input{fig/tikz/synthetic_data/per_adv_synth_data_detect_shaded_groundtruth.tikz}}}
}
\caption{Results for a synthetic dataset with varying percentage of \emph{Type A} adversaries}
\label{fig:synth_data_per_adv}
\end{figure}
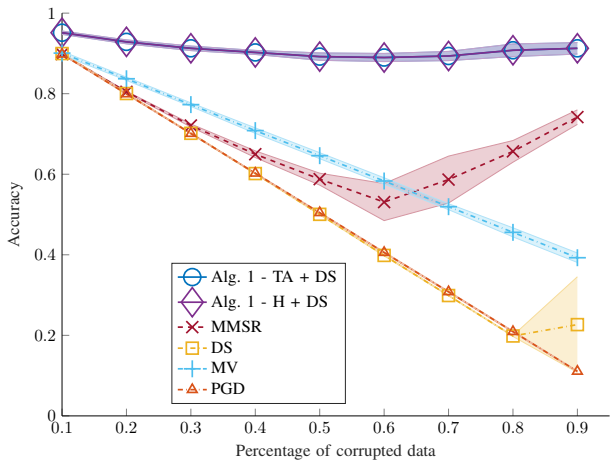
\begin{figure}[tb]
     \centering
     \resizebox{0.45\columnwidth}{!}{\input{fig/tikz/synthetic_data/p_adv_synth_data_acc_groundtruth_shaded_acc.tikz}
     }
        \caption{Results for a synthetic dataset with \emph{Type A} adversaries and varying percentage of corrupted data}
        \label{fig:synth_data_p_adv}
\end{figure}

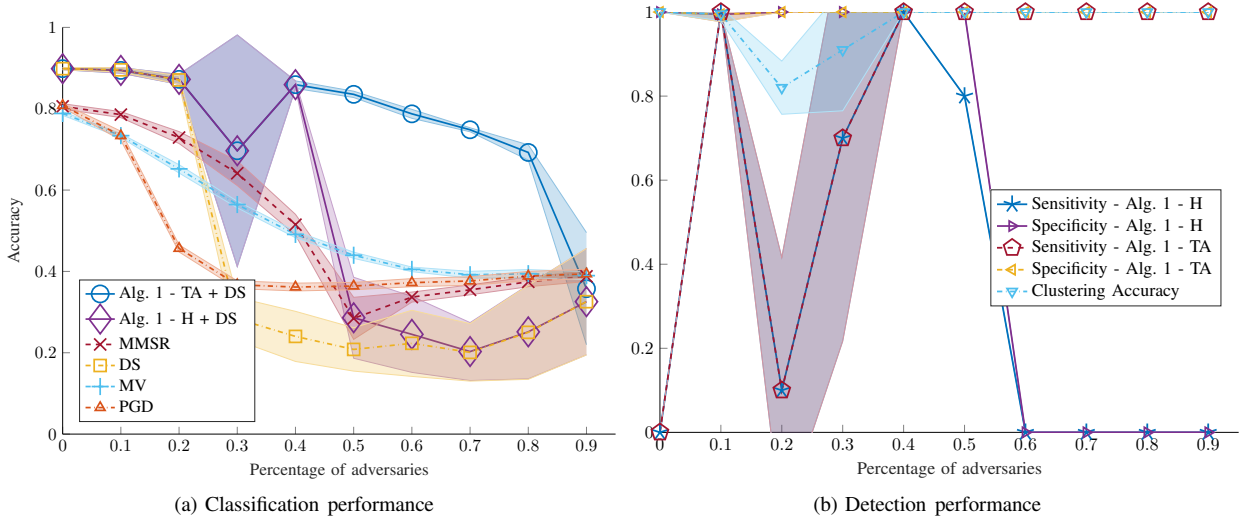
\begin{figure}[htb]
\centering{
\subfloat[Classification performance]{
\resizebox{0.45\columnwidth}{!}{\input{fig/tikz/synthetic_data/per_adv_synth_data_acc_DS_shaded_acc.tikz}}}
\subfloat[Detection performance]{
\resizebox{0.45\columnwidth}{!}{\input{fig/tikz/synthetic_data/per_adv_synth_data_detect_shaded_DS.tikz}}}
}
\caption{Results for a synthetic dataset with varying percentage of \emph{Type B} adversaries}
\label{fig:synth_data_per_adv_DS}
\end{figure}

\begin{figure}[htb]
\centering{
\subfloat[Classification performance]{
\resizebox{0.45\columnwidth}{!}{\input{fig/tikz/synthetic_data/p_adv_synth_data_acc_DS_shaded_acc.tikz}}}
\subfloat[Detection performance]{
\resizebox{0.45\columnwidth}{!}{\input{fig/tikz/synthetic_data/p_adv_synth_data_detect_shaded_DS.tikz}}}
}
\caption{Results for a synthetic dataset with \emph{Type B} adversaries and varying percentage of corrupted data}
\label{fig:synth_data_p_adv_DS}
\end{figure}
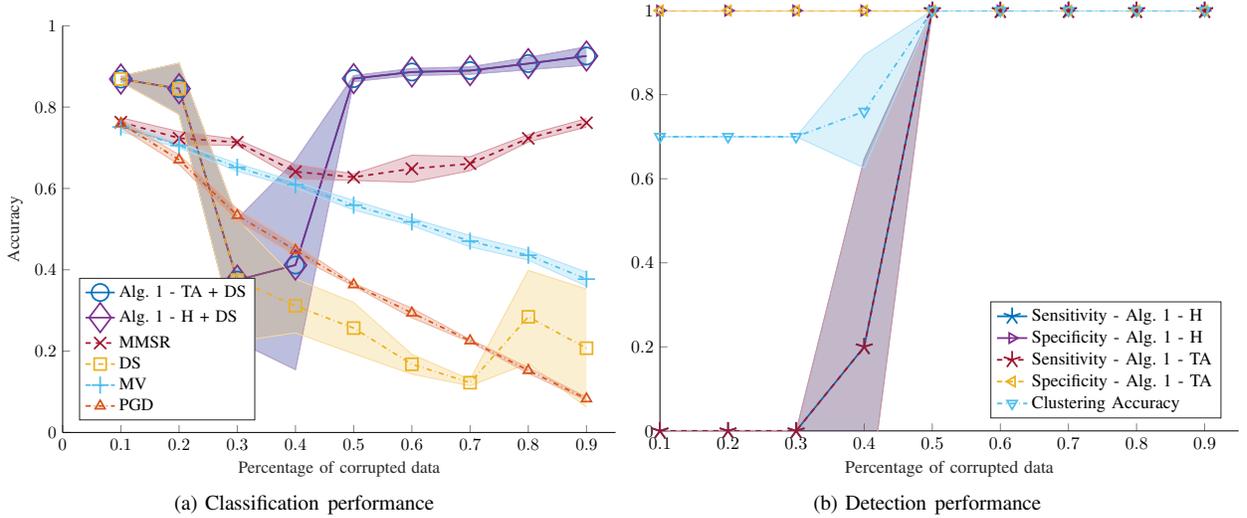

Results for the same dataset, with \emph{Type B} adversaries are shown in Fig.~\ref{fig:synth_data_per_adv_DS}, for a varying percentage of adversaries. Here, $p_{\rm corr} = 0.5$.
\emph{DS} remains robust for up to $0.2M$ adversarial annotators. After that point  
\emph{PGD, MV} and \emph{MMSR} outperform \emph{DS}, which exhibits the lowest accuracy. The proposed approaches \emph{Alg.~\ref{alg:main_alg} - H + DS} and \emph{Alg.~\ref{alg:main_alg} - TA + DS} still exhibit high levels of accuracy even under this adversarial model. Note that, despite the annotator misclassification by Alg.~\ref{alg:main_alg} when adversaries are approximately $0.2M$, data classification performance remains high. Fig.~\ref{fig:synth_data_p_adv_DS} shows results when $p_{\rm adv}=0.3$ and $p_{\rm corr}$ varies. Here \emph{MMSR} exhibits relatively stable accuracy for all values of $p_{\rm corr}$, whereas the performance of \emph{MV,DS} and \emph{PGD} decreases as the number of corrupted data increase. Large amounts of corrupted data are beneficial for Alg.~\ref{alg:main_alg} as they enable accurate detection of adversaries, whereas for small values of $p_{\rm corr}$ Alg.~\ref{alg:main_alg} assigns all annotators to one cluster, i.e. does not detect the presence of adversaries. Overall, \emph{Type B} adversaries can be accurately detected by Alg.~\ref{alg:main_alg} when their population is large and/or they corrupt large amounts of data.

\subsection{Real data}
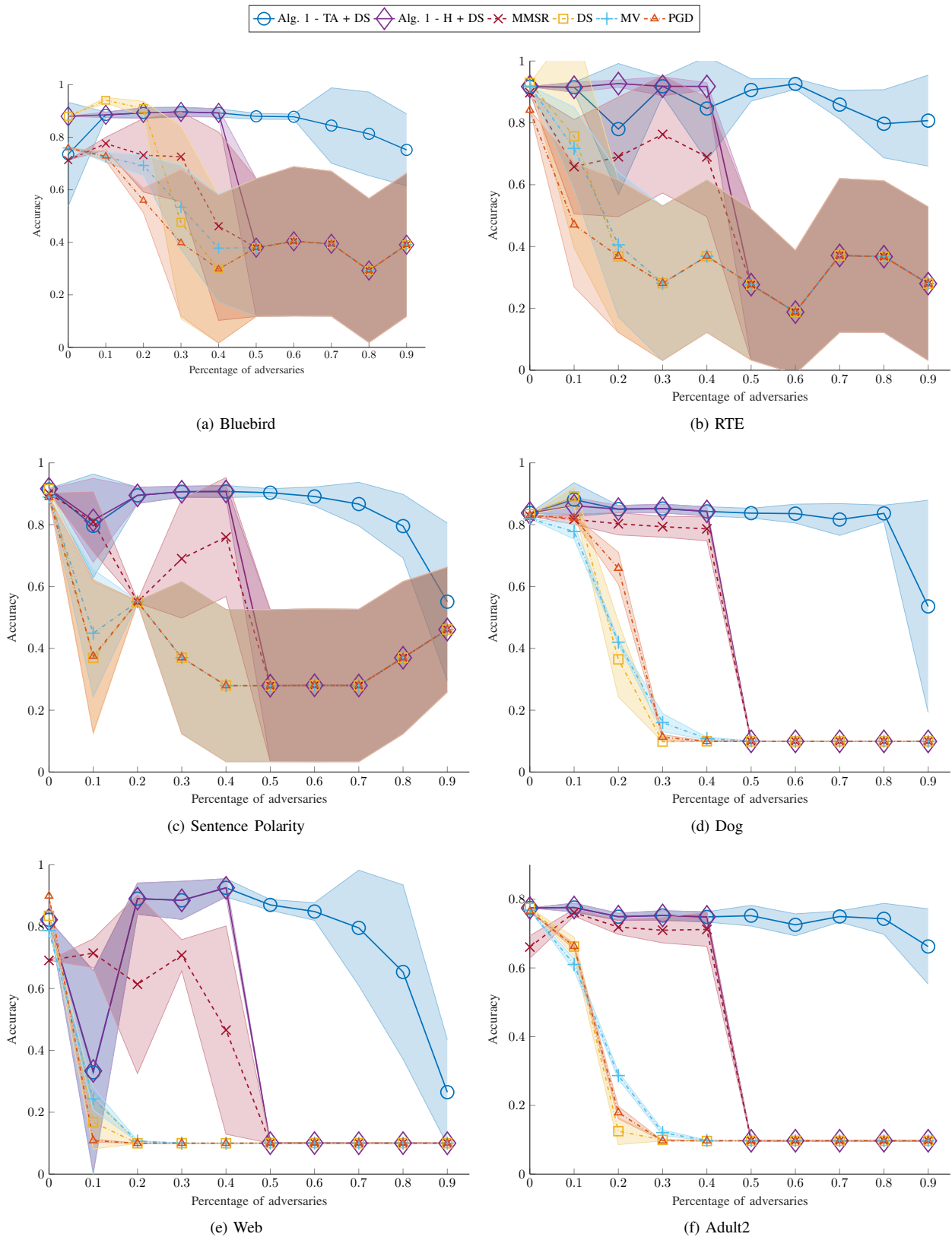
\begin{figure*}[tb]
\centering
\hspace*{2cm}\subfloat{\centering
\resizebox{0.55\textwidth}{!}{\input{fig/tikz/legend_acc.tikz}}}
\\
\subfloat[Bluebird]{
\resizebox{0.45\columnwidth}{!}{\input{fig/tikz/real_data/per_adv_groundtruth_bluebird_shaded_acc.tikz}}
}
\subfloat[RTE]{
\resizebox{0.45\columnwidth}{!}{\input{fig/tikz/real_data/per_adv_groundtruth_rte_shaded_acc.tikz}}
}
\\
\subfloat[Sentence Polarity]{
\resizebox{0.45\columnwidth}{!}{\input{fig/tikz/real_data/per_adv_groundtruth_sen_polarity_shaded_acc.tikz}}
}
\subfloat[Dog]{
\resizebox{0.45\columnwidth}{!}{\input{fig/tikz/real_data/per_adv_groundtruth_dog_shaded_acc.tikz}}
}
\\
\subfloat[Web]{
\resizebox{0.45\columnwidth}{!}{\input{fig/tikz/real_data/per_adv_groundtruth_web_shaded_acc.tikz}}
}
\subfloat[Adult2]{
\resizebox{0.45\columnwidth}{!}{\input{fig/tikz/real_data/per_adv_groundtruth_Adult2_shaded_acc.tikz}}
}
\caption{Classification results for real crowdsourcing datasets with varying percentage of \emph{Type A} adversaries}\label{fig:real_data_per_adv}
\end{figure*}

Further tests were conducted on real crowdsourcing datasets, namely the Bluebird~\cite{multidim_wisdom}, RTE~\cite{cheapnfast}, Sentence polarity~\cite{musicgenre_senpoldata}, Dog~\cite{imagenet}, Web~\cite{minimax_crowd}, and Adult2  datasets, whose properties are tabulated in Tab. \ref{tab:realdata}. The average number of responses per annotator is denoted by $\delta$. Following the numerical test strategy of \cite{MMSR}, for all datasets, the percentage of corrupted data is fixed to $p_{\rm corr}$, and we vary the number of adversaries. Further, annotators that provided the same response for all data were removed from the datasets. Fig. \ref{fig:real_data_per_adv} shows the classification accuracy, as the number of \emph{Type A} adversaries increases for the considered datasets, with $p_{\rm corr} = 0.9$. Adversaries provide responses for data with probability $p_{\rm obs} = 0.3.$ Trends similar to those of the synthetic data tests can be observed. \emph{MV, DS,} and \emph{PGD} tolerate very few adversaries before starting to lose accuracy, in all datasets. \emph{MMSR} outperforms the non-adversarially robust methods for approximately up to $p_{\rm adv} = 0.5$. Both variants of Alg.~\ref{alg:main_alg} achieve high classification accuracy, with \emph{Alg. \ref{alg:main_alg} - TA} outperforming all other algorithms for the range of $p_{\rm adv}$ considered. \emph{Alg. \ref{alg:main_alg} - H} performs similarly to \emph{Alg. \ref{alg:main_alg} - TA} for up to $p_{\rm adv} = 0.5$, as expected, and exhibits higher classification accuracy than \emph{MMSR} in most cases. This is probably due to the use of the Dawid and Skene model in the derivation of \emph{Alg. 1} instead of the one-coin model used in \emph{MMSR}. Interestingly, we observe a performance ``dip'' for Alg. \ref{alg:main_alg} at $p_{\rm adv} = 0.1$ for some datasets. This may be attributed to the subspace clustering algorithm used; alternative subspace clustering techniques, that can handle imbalanced datasets~\cite{exemplar_SC}, may yield improved results. 

\begin{figure*}[tb]
\centering
\hspace*{2cm}\subfloat{\centering
\resizebox{0.55\textwidth}{!}{\input{fig/tikz/legend_acc.tikz}}}
\\
\subfloat[Bluebird]{
\resizebox{0.45\columnwidth}{!}{\input{fig/tikz/real_data/per_adv_DS_bluebird_shaded_acc.tikz}}
}
\subfloat[RTE]{
\resizebox{0.45\columnwidth}{!}{\input{fig/tikz/real_data/per_adv_DS_rte_shaded_acc.tikz}}
}
\\
\subfloat[Sentence Polarity]{
\resizebox{0.45\columnwidth}{!}{\input{fig/tikz/real_data/per_adv_DS_sen_polarity_shaded_acc.tikz}}
}
\subfloat[Dog]{
\resizebox{0.45\columnwidth}{!}{\input{fig/tikz/real_data/per_adv_DS_dog_shaded_acc.tikz}}
}
\\
\subfloat[Web]{
\resizebox{0.45\columnwidth}{!}{\input{fig/tikz/real_data/per_adv_DS_web_shaded_acc.tikz}}
}
\subfloat[Adult2]{
\resizebox{0.45\columnwidth}{!}{\input{fig/tikz/real_data/per_adv_DS_Adult2_shaded_acc.tikz}}
}
\caption{Classification results for real crowdsourcing datasets with varying percentage of \emph{Type B} adversaries}\label{fig:real_data_per_adv_DS}
\end{figure*}
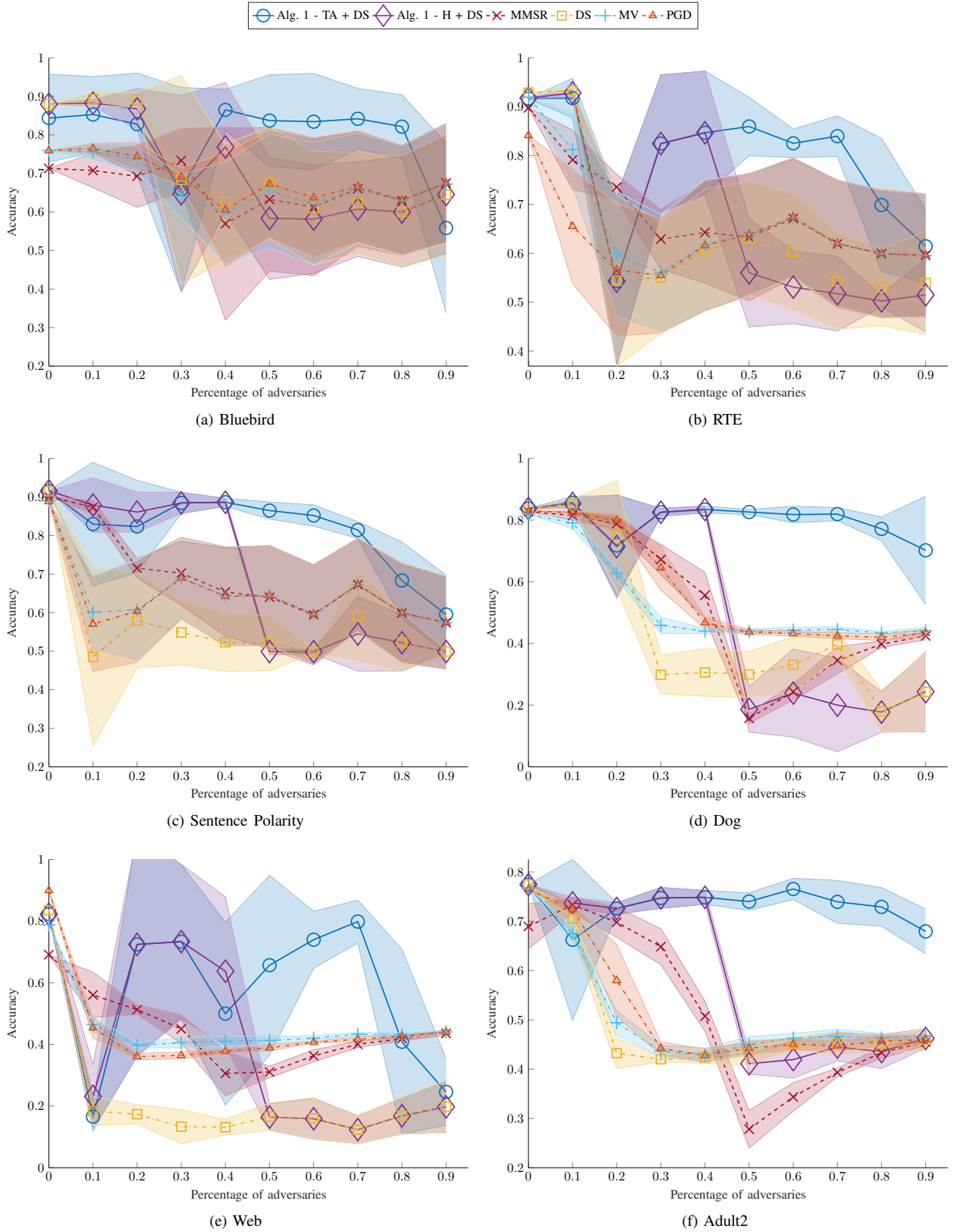

Results for the same datasets, but with \emph{Type B} adversaries and $p_{\rm corr} = 0.5$ are shown in Fig.~\ref{fig:real_data_per_adv_DS}. As expected, for all datasets the performance of \emph{DS, MV} and \emph{PDG} declines as the number of adversaries increases. Furthermore, \emph{MMSR} seems more sensitive to this type of adversarial attack as its accuracy also decreases. Despite increased variability in results, possibly caused by the subspace clustering step, Alg.~\ref{alg:main_alg} exhibits similar performance trends as with the \emph{Type A} adversaries, and manages to outperform the competing alternatives for most percentages of adversaries.
\begin{table*}[tb]
\caption{Real dataset properties}
\label{tab:realdata}
\centering
\begin{tabular}{|c||c|c|c|c|c|c|}
\hline
Dataset & Bluebird & RTE  & Sentence Polarity  & Dog  &  Web & Adult2    \\ \hline
$N$       & $108$ & $800$  & $5,000$ & $807$  & $2,665$ &  $333$  \\ \hline
$M$       & $39$  & $164$  & $203$   & $109$  & $177$   &   $269$ \\ \hline
$K$       & $2$   & $2$    & $2$     & $5$    & $5$     &  $4$   \\ \hline
$\delta$  & $108$ & $48.78$& $136.68$& $74.03$& $87.94$ &  $12.33$ \\ \hline
\end{tabular}
\end{table*}

\addtocounter{subfigure}{-1}

\addtocounter{subfigure}{-1}
\section{Conclusions}
\label{sec:conclusions}
This paper investigated crowdsourcing under adversarial attacks. A subspace clustering based algorithm was developed to detect adversaries and perform label aggregation, and its performance was evaluated on synthetic and real data. 

Future research will involve theoretical analysis of the proposed method, alongside algorithms that can handle more advanced adversaries, enhanced label aggregation methods in the presence of adversaries, and online variants of the algorithm to handle streaming annotators and data.

\bibliographystyle{IEEEtranS}
\bibliography{icdm2021}

\end{document}

%% file: fig/tikz/synthetic_data/per_adv_synth_data_acc_groundtruth_shaded_acc.tikz
\definecolor{mycolor1}{rgb}{0.00000,0.44700,0.74100}%
\definecolor{mycolor2}{rgb}{0.49400,0.18400,0.55600}%
\definecolor{mycolor3}{rgb}{0.63500,0.07800,0.18400}%
\definecolor{mycolor4}{rgb}{0.92900,0.69400,0.12500}%
\definecolor{mycolor5}{rgb}{0.30100,0.74500,0.93300}%
\definecolor{mycolor6}{rgb}{0.85000,0.32500,0.09800}%
\begin{tikzpicture}

\begin{axis}[%
width=4.464in,
height=3.286in,
at={(0.815in,0.76in)},
scale only axis,
xmin=0,
xmax=1,
xlabel style={font=\color{white!15!black}},
xlabel={Percentage of adversaries},
ymin=0.45,
ymax=1,
ylabel style={font=\color{white!15!black}},
ylabel={Accuracy},
axis background/.style={fill=white},
axis x line*=bottom,
axis y line*=left,
legend style={legend cell align=left, align=left, draw=white!15!black}
]

\addplot[area legend, draw=none, fill=mycolor1, fill opacity=0.2, forget plot]
table[row sep=crcr] {%
x	y\\
0	0.890285710364508\\
0.1	0.908796619484115\\
0.2	0.902438928638493\\
0.3	0.888713681489912\\
0.4	0.860256543126364\\
0.5	0.832817235392785\\
0.6	0.78176205678246\\
0.7	0.739725161221208\\
0.8	0.684636638138315\\
0.9	0.495384030127336\\
0.9	0.573070607484549\\
0.8	0.733383642181854\\
0.7	0.773525493510631\\
0.6	0.811386706722069\\
0.5	0.85903481100078\\
0.4	0.871044229249964\\
0.3	0.906576965414493\\
0.2	0.913668074897378\\
0.1	0.916867741578622\\
0	0.902894974269603\\
}--cycle;
\addplot [color=white!55!mycolor1, forget plot]
  table[row sep=crcr]{%
0	0.890285710364508\\
0.1	0.908796619484115\\
0.2	0.902438928638493\\
0.3	0.888713681489912\\
0.4	0.860256543126364\\
0.5	0.832817235392785\\
0.6	0.78176205678246\\
0.7	0.739725161221208\\
0.8	0.684636638138315\\
0.9	0.495384030127336\\
};
\addplot [color=white!55!mycolor1, forget plot]
  table[row sep=crcr]{%
0	0.902894974269603\\
0.1	0.916867741578622\\
0.2	0.913668074897378\\
0.3	0.906576965414493\\
0.4	0.871044229249964\\
0.5	0.85903481100078\\
0.6	0.811386706722069\\
0.7	0.773525493510631\\
0.8	0.733383642181854\\
0.9	0.573070607484549\\
};
\addplot [color=mycolor1, line width=1.0pt, mark size=5.0pt, mark=o, mark options={solid, mycolor1}]
  table[row sep=crcr]{%
0	0.896590342317055\\
0.1	0.912832180531368\\
0.2	0.908053501767936\\
0.3	0.897645323452202\\
0.4	0.865650386188164\\
0.5	0.845926023196782\\
0.6	0.796574381752265\\
0.7	0.75662532736592\\
0.8	0.709010140160084\\
0.9	0.534227318805943\\
};
\addlegendentry{Alg. 1 - TA + DS}

\addplot[area legend, draw=none, fill=mycolor2, fill opacity=0.2, forget plot]
table[row sep=crcr] {%
x	y\\
0	0.890285710364508\\
0.1	0.908796619484115\\
0.2	0.902438928638493\\
0.3	0.888713681489912\\
0.4	0.860256543126364\\
0.5	0.49919926886455\\
0.6	0.499899051622082\\
0.7	0.500009809921946\\
0.8	0.499885128162542\\
0.9	0.499888242853552\\
0.9	0.500191885348369\\
0.8	0.499994767730131\\
0.7	0.500470670597065\\
0.6	0.500381341062426\\
0.5	0.500600979884091\\
0.4	0.871044229249964\\
0.3	0.906576965414493\\
0.2	0.913668074897378\\
0.1	0.916867741578622\\
0	0.902894974269603\\
}--cycle;
\addplot [color=white!55!mycolor2, forget plot]
  table[row sep=crcr]{%
0	0.890285710364508\\
0.1	0.908796619484115\\
0.2	0.902438928638493\\
0.3	0.888713681489912\\
0.4	0.860256543126364\\
0.5	0.49919926886455\\
0.6	0.499899051622082\\
0.7	0.500009809921946\\
0.8	0.499885128162542\\
0.9	0.499888242853552\\
};
\addplot [color=white!55!mycolor2, forget plot]
  table[row sep=crcr]{%
0	0.902894974269603\\
0.1	0.916867741578622\\
0.2	0.913668074897378\\
0.3	0.906576965414493\\
0.4	0.871044229249964\\
0.5	0.500600979884091\\
0.6	0.500381341062426\\
0.7	0.500470670597065\\
0.8	0.499994767730131\\
0.9	0.500191885348369\\
};
\addplot [color=mycolor2, line width=1.0pt, mark size=8.6pt, mark=diamond, mark options={solid, mycolor2}]
  table[row sep=crcr]{%
0	0.896590342317055\\
0.1	0.912832180531368\\
0.2	0.908053501767936\\
0.3	0.897645323452202\\
0.4	0.865650386188164\\
0.5	0.49990012437432\\
0.6	0.500140196342254\\
0.7	0.500240240259506\\
0.8	0.499939947946336\\
0.9	0.50004006410096\\
};
\addlegendentry{Alg. 1 - H + DS}

\addplot[area legend, draw=none, fill=mycolor3, fill opacity=0.2, forget plot]
table[row sep=crcr] {%
x	y\\
0	0.802229439012998\\
0.1	0.783626094202713\\
0.2	0.680777557944397\\
0.3	0.574438984159126\\
0.4	0.509663929363164\\
0.5	0.499577774000619\\
0.6	0.499899051622082\\
0.7	0.49990838024983\\
0.8	0.499885128162542\\
0.9	0.499888242853552\\
0.9	0.500191885348369\\
0.8	0.499994767730131\\
0.7	0.500492004160262\\
0.6	0.500381341062426\\
0.5	0.500782907174127\\
0.4	0.512308118961971\\
0.3	0.58910648088581\\
0.2	0.728606657631249\\
0.1	0.799553830579495\\
0	0.818181269217876\\
}--cycle;
\addplot [color=white!55!mycolor3, forget plot]
  table[row sep=crcr]{%
0	0.802229439012998\\
0.1	0.783626094202713\\
0.2	0.680777557944397\\
0.3	0.574438984159126\\
0.4	0.509663929363164\\
0.5	0.499577774000619\\
0.6	0.499899051622082\\
0.7	0.49990838024983\\
0.8	0.499885128162542\\
0.9	0.499888242853552\\
};
\addplot [color=white!55!mycolor3, forget plot]
  table[row sep=crcr]{%
0	0.818181269217876\\
0.1	0.799553830579495\\
0.2	0.728606657631249\\
0.3	0.58910648088581\\
0.4	0.512308118961971\\
0.5	0.500782907174127\\
0.6	0.500381341062426\\
0.7	0.500492004160262\\
0.8	0.499994767730131\\
0.9	0.500191885348369\\
};
\addplot [color=mycolor3, dashed, line width=1.0pt, mark size=5.0pt, mark=x, mark options={solid, mycolor3}]
  table[row sep=crcr]{%
0	0.810205354115437\\
0.1	0.791589962391104\\
0.2	0.704692107787823\\
0.3	0.581772732522468\\
0.4	0.510986024162568\\
0.5	0.500180340587373\\
0.6	0.500140196342254\\
0.7	0.500200192205046\\
0.8	0.499939947946336\\
0.9	0.50004006410096\\
};
\addlegendentry{MMSR}

\addplot[area legend, draw=none, fill=mycolor4, fill opacity=0.2, forget plot]
table[row sep=crcr] {%
x	y\\
0	0.890285710364508\\
0.1	0.909457633265641\\
0.2	0.497841740730895\\
0.3	0.499264762658909\\
0.4	0.499525385546015\\
0.5	0.499532303072093\\
0.6	0.499840690004285\\
0.7	0.499871781200647\\
0.8	0.499885128162542\\
0.9	0.499888242853552\\
0.9	0.500191885348369\\
0.8	0.499994767730131\\
0.7	0.500448555187044\\
0.6	0.500279590600166\\
0.5	0.500588105689189\\
0.4	0.501395182955281\\
0.3	0.50225650232093\\
0.2	0.50620188770639\\
0.1	0.923256324861252\\
0	0.902894974269603\\
}--cycle;
\addplot [color=white!55!mycolor4, forget plot]
  table[row sep=crcr]{%
0	0.890285710364508\\
0.1	0.909457633265641\\
0.2	0.497841740730895\\
0.3	0.499264762658909\\
0.4	0.499525385546015\\
0.5	0.499532303072093\\
0.6	0.499840690004285\\
0.7	0.499871781200647\\
0.8	0.499885128162542\\
0.9	0.499888242853552\\
};
\addplot [color=white!55!mycolor4, forget plot]
  table[row sep=crcr]{%
0	0.902894974269603\\
0.1	0.923256324861252\\
0.2	0.50620188770639\\
0.3	0.50225650232093\\
0.4	0.501395182955281\\
0.5	0.500588105689189\\
0.6	0.500279590600166\\
0.7	0.500448555187044\\
0.8	0.499994767730131\\
0.9	0.500191885348369\\
};
\addplot [color=mycolor4, dashdotted, line width=1.0pt, mark size=3.5pt, mark=square, mark options={solid, mycolor4}]
  table[row sep=crcr]{%
0	0.896590342317055\\
0.1	0.916356979063447\\
0.2	0.502021814218642\\
0.3	0.500760632489919\\
0.4	0.500460284250648\\
0.5	0.500060204380641\\
0.6	0.500060140302225\\
0.7	0.500160168193845\\
0.8	0.499939947946336\\
0.9	0.50004006410096\\
};
\addlegendentry{DS}

\addplot[area legend, draw=none, fill=mycolor5, fill opacity=0.2, forget plot]
table[row sep=crcr] {%
x	y\\
0	0.785854708514237\\
0.1	0.766283207729322\\
0.2	0.708683184918427\\
0.3	0.635958195752696\\
0.4	0.574526343096477\\
0.5	0.533189355389077\\
0.6	0.507945101905008\\
0.7	0.50217616321886\\
0.8	0.500085285809478\\
0.9	0.499888242853552\\
0.9	0.500191885348369\\
0.8	0.500195042669731\\
0.7	0.503669795291229\\
0.6	0.512355174060978\\
0.5	0.538362719940431\\
0.4	0.585115222428701\\
0.3	0.653403437887802\\
0.2	0.720491403324737\\
0.1	0.785424455393688\\
0	0.794747309079583\\
}--cycle;
\addplot [color=white!55!mycolor5, forget plot]
  table[row sep=crcr]{%
0	0.785854708514237\\
0.1	0.766283207729322\\
0.2	0.708683184918427\\
0.3	0.635958195752696\\
0.4	0.574526343096477\\
0.5	0.533189355389077\\
0.6	0.507945101905008\\
0.7	0.50217616321886\\
0.8	0.500085285809478\\
0.9	0.499888242853552\\
};
\addplot [color=white!55!mycolor5, forget plot]
  table[row sep=crcr]{%
0	0.794747309079583\\
0.1	0.785424455393688\\
0.2	0.720491403324737\\
0.3	0.653403437887802\\
0.4	0.585115222428701\\
0.5	0.538362719940431\\
0.6	0.512355174060978\\
0.7	0.503669795291229\\
0.8	0.500195042669731\\
0.9	0.500191885348369\\
};
\addplot [color=mycolor5, dashdotted, line width=1.0pt, mark size=5.0pt, mark=+, mark options={solid, mycolor5}]
  table[row sep=crcr]{%
0	0.79030100879691\\
0.1	0.775853831561505\\
0.2	0.714587294121582\\
0.3	0.644680816820249\\
0.4	0.579820782762589\\
0.5	0.535776037664754\\
0.6	0.510150137982993\\
0.7	0.502922979255045\\
0.8	0.500140164239605\\
0.9	0.50004006410096\\
};
\addlegendentry{MV}

\addplot[area legend, draw=none, fill=mycolor6, fill opacity=0.2, forget plot]
table[row sep=crcr] {%
x	y\\
0	0.802568254687586\\
0.1	0.659278145797955\\
0.2	0.516086652794123\\
0.3	0.50404301383859\\
0.4	0.500622881681307\\
0.5	0.499551003580334\\
0.6	0.499899051622082\\
0.7	0.49990838024983\\
0.8	0.499885128162542\\
0.9	0.499888242853552\\
0.9	0.500191885348369\\
0.8	0.499994767730131\\
0.7	0.500492004160262\\
0.6	0.500381341062426\\
0.5	0.50104999805591\\
0.4	0.502538583274678\\
0.3	0.505087029605257\\
0.2	0.519193377105789\\
0.1	0.672210566727548\\
0	0.821126058244801\\
}--cycle;
\addplot [color=white!55!mycolor6, forget plot]
  table[row sep=crcr]{%
0	0.802568254687586\\
0.1	0.659278145797955\\
0.2	0.516086652794123\\
0.3	0.50404301383859\\
0.4	0.500622881681307\\
0.5	0.499551003580334\\
0.6	0.499899051622082\\
0.7	0.49990838024983\\
0.8	0.499885128162542\\
0.9	0.499888242853552\\
};
\addplot [color=white!55!mycolor6, forget plot]
  table[row sep=crcr]{%
0	0.821126058244801\\
0.1	0.672210566727548\\
0.2	0.519193377105789\\
0.3	0.505087029605257\\
0.4	0.502538583274678\\
0.5	0.50104999805591\\
0.6	0.500381341062426\\
0.7	0.500492004160262\\
0.8	0.499994767730131\\
0.9	0.500191885348369\\
};
\addplot [color=mycolor6, dashdotted, line width=1.0pt, mark size=3.3pt, mark=triangle, mark options={solid, mycolor6}]
  table[row sep=crcr]{%
0	0.811847156466194\\
0.1	0.665744356262752\\
0.2	0.517640014949956\\
0.3	0.504565021721923\\
0.4	0.501580732477993\\
0.5	0.500300500818122\\
0.6	0.500140196342254\\
0.7	0.500200192205046\\
0.8	0.499939947946336\\
0.9	0.50004006410096\\
};
\addlegendentry{PGD}

\end{axis}
\end{tikzpicture}%

%% file: fig/tikz/synthetic_data/per_adv_synth_data_detect_shaded_groundtruth.tikz
\definecolor{mycolor1}{rgb}{0.00000,0.44700,0.74100}%
\definecolor{mycolor2}{rgb}{0.49400,0.18400,0.55600}%
\definecolor{mycolor3}{rgb}{0.63500,0.07800,0.18400}%
\definecolor{mycolor4}{rgb}{0.92900,0.69400,0.12500}%
\definecolor{mycolor5}{rgb}{0.30100,0.74500,0.93300}%
\begin{tikzpicture}

\begin{axis}[%
width=4.521in,
height=3.286in,
at={(0.758in,0.76in)},
scale only axis,
xmin=0,
xmax=1,
xlabel style={font=\color{white!15!black}},
xlabel={Percentage of adversaries},
ymin=0,
ymax=1,
axis background/.style={fill=white},
axis x line*=bottom,
axis y line*=left,
legend style={at={(0.97,0.03)}, anchor=south east, legend cell align=left, align=left, draw=white!15!black}
]

\addplot[area legend, draw=none, fill=mycolor1, fill opacity=0.2, forget plot]
table[row sep=crcr] {%
x	y\\
0	0\\
0.1	1\\
0.2	1\\
0.3	1\\
0.4	1\\
0.5	0\\
0.6	0\\
0.7	0\\
0.8	0\\
0.9	0\\
0.9	0\\
0.8	0\\
0.7	0\\
0.6	0\\
0.5	0\\
0.4	1\\
0.3	1\\
0.2	1\\
0.1	1\\
0	0\\
}--cycle;
\addplot [color=white!55!mycolor1, forget plot]
  table[row sep=crcr]{%
0	0\\
0.1	1\\
0.2	1\\
0.3	1\\
0.4	1\\
0.5	0\\
0.6	0\\
0.7	0\\
0.8	0\\
0.9	0\\
};
\addplot [color=white!55!mycolor1, forget plot]
  table[row sep=crcr]{%
0	0\\
0.1	1\\
0.2	1\\
0.3	1\\
0.4	1\\
0.5	0\\
0.6	0\\
0.7	0\\
0.8	0\\
0.9	0\\
};
\addplot [color=mycolor1, line width=1.0pt, mark size=5.0pt, mark=star, mark options={solid, mycolor1}]
  table[row sep=crcr]{%
0	0\\
0.1	1\\
0.2	1\\
0.3	1\\
0.4	1\\
0.5	0\\
0.6	0\\
0.7	0\\
0.8	0\\
0.9	0\\
};
\addlegendentry{Sensitivity - Alg. 1 - H}

\addplot[area legend, draw=none, fill=mycolor2, fill opacity=0.2, forget plot]
table[row sep=crcr] {%
x	y\\
0	1\\
0.1	0.982449582268423\\
0.2	1\\
0.3	1\\
0.4	1\\
0.5	0\\
0.6	0\\
0.7	0\\
0.8	0\\
0.9	0\\
0.9	0\\
0.8	0\\
0.7	0\\
0.6	0\\
0.5	0\\
0.4	1\\
0.3	1\\
0.2	1\\
0.1	1.00273560291676\\
0	1\\
}--cycle;
\addplot [color=white!55!mycolor2, forget plot]
  table[row sep=crcr]{%
0	1\\
0.1	0.982449582268423\\
0.2	1\\
0.3	1\\
0.4	1\\
0.5	0\\
0.6	0\\
0.7	0\\
0.8	0\\
0.9	0\\
};
\addplot [color=white!55!mycolor2, forget plot]
  table[row sep=crcr]{%
0	1\\
0.1	1.00273560291676\\
0.2	1\\
0.3	1\\
0.4	1\\
0.5	0\\
0.6	0\\
0.7	0\\
0.8	0\\
0.9	0\\
};
\addplot [color=mycolor2, line width=1.0pt, mark size=3.3pt, mark=triangle, mark options={solid, rotate=270, mycolor2}]
  table[row sep=crcr]{%
0	1\\
0.1	0.992592592592593\\
0.2	1\\
0.3	1\\
0.4	1\\
0.5	0\\
0.6	0\\
0.7	0\\
0.8	0\\
0.9	0\\
};
\addlegendentry{Specificity - Alg. 1 - H}

\addplot[area legend, draw=none, fill=mycolor3, fill opacity=0.2, forget plot]
table[row sep=crcr] {%
x	y\\
0	0\\
0.1	1\\
0.2	1\\
0.3	1\\
0.4	1\\
0.5	1\\
0.6	1\\
0.7	1\\
0.8	1\\
0.9	1\\
0.9	1\\
0.8	1\\
0.7	1\\
0.6	1\\
0.5	1\\
0.4	1\\
0.3	1\\
0.2	1\\
0.1	1\\
0	0\\
}--cycle;
\addplot [color=white!55!mycolor3, forget plot]
  table[row sep=crcr]{%
0	0\\
0.1	1\\
0.2	1\\
0.3	1\\
0.4	1\\
0.5	1\\
0.6	1\\
0.7	1\\
0.8	1\\
0.9	1\\
};
\addplot [color=white!55!mycolor3, forget plot]
  table[row sep=crcr]{%
0	0\\
0.1	1\\
0.2	1\\
0.3	1\\
0.4	1\\
0.5	1\\
0.6	1\\
0.7	1\\
0.8	1\\
0.9	1\\
};
\addplot [color=mycolor3, dashed, line width=1.0pt, mark size=5.0pt, mark=pentagon, mark options={solid, mycolor3}]
  table[row sep=crcr]{%
0	0\\
0.1	1\\
0.2	1\\
0.3	1\\
0.4	1\\
0.5	1\\
0.6	1\\
0.7	1\\
0.8	1\\
0.9	1\\
};
\addlegendentry{Sensitivity - Alg. 1 - TA}

\addplot[area legend, draw=none, fill=mycolor4, fill opacity=0.2, forget plot]
table[row sep=crcr] {%
x	y\\
0	1\\
0.1	0.982449582268423\\
0.2	1\\
0.3	1\\
0.4	1\\
0.5	1\\
0.6	1\\
0.7	1\\
0.8	1\\
0.9	1\\
0.9	1\\
0.8	1\\
0.7	1\\
0.6	1\\
0.5	1\\
0.4	1\\
0.3	1\\
0.2	1\\
0.1	1.00273560291676\\
0	1\\
}--cycle;
\addplot [color=white!55!mycolor4, forget plot]
  table[row sep=crcr]{%
0	1\\
0.1	0.982449582268423\\
0.2	1\\
0.3	1\\
0.4	1\\
0.5	1\\
0.6	1\\
0.7	1\\
0.8	1\\
0.9	1\\
};
\addplot [color=white!55!mycolor4, forget plot]
  table[row sep=crcr]{%
0	1\\
0.1	1.00273560291676\\
0.2	1\\
0.3	1\\
0.4	1\\
0.5	1\\
0.6	1\\
0.7	1\\
0.8	1\\
0.9	1\\
};
\addplot [color=mycolor4, dashed, line width=1.0pt, mark size=3.3pt, mark=triangle, mark options={solid, rotate=90, mycolor4}]
  table[row sep=crcr]{%
0	1\\
0.1	0.992592592592593\\
0.2	1\\
0.3	1\\
0.4	1\\
0.5	1\\
0.6	1\\
0.7	1\\
0.8	1\\
0.9	1\\
};
\addlegendentry{Specificity - Alg. 1 - TA}

\addplot[area legend, draw=none, fill=mycolor5, fill opacity=0.2, forget plot]
table[row sep=crcr] {%
x	y\\
0	1\\
0.1	0.984204624041581\\
0.2	1\\
0.3	1\\
0.4	1\\
0.5	1\\
0.6	1\\
0.7	1\\
0.8	1\\
0.9	1\\
0.9	1\\
0.8	1\\
0.7	1\\
0.6	1\\
0.5	1\\
0.4	1\\
0.3	1\\
0.2	1\\
0.1	1.00246204262509\\
0	1\\
}--cycle;
\addplot [color=white!55!mycolor5, forget plot]
  table[row sep=crcr]{%
0	1\\
0.1	0.984204624041581\\
0.2	1\\
0.3	1\\
0.4	1\\
0.5	1\\
0.6	1\\
0.7	1\\
0.8	1\\
0.9	1\\
};
\addplot [color=white!55!mycolor5, forget plot]
  table[row sep=crcr]{%
0	1\\
0.1	1.00246204262509\\
0.2	1\\
0.3	1\\
0.4	1\\
0.5	1\\
0.6	1\\
0.7	1\\
0.8	1\\
0.9	1\\
};
\addplot [color=mycolor5, dashdotted, line width=1.0pt, mark size=3.3pt, mark=triangle, mark options={solid, rotate=180, mycolor5}]
  table[row sep=crcr]{%
0	1\\
0.1	0.993333333333333\\
0.2	1\\
0.3	1\\
0.4	1\\
0.5	1\\
0.6	1\\
0.7	1\\
0.8	1\\
0.9	1\\
};
\addlegendentry{Clustering Accuracy}

\end{axis}
\end{tikzpicture}%

%% file: fig/tikz/synthetic_data/p_adv_synth_data_acc_groundtruth_shaded_acc.tikz
\definecolor{mycolor1}{rgb}{0.00000,0.44700,0.74100}%
\definecolor{mycolor2}{rgb}{0.49400,0.18400,0.55600}%
\definecolor{mycolor3}{rgb}{0.63500,0.07800,0.18400}%
\definecolor{mycolor4}{rgb}{0.92900,0.69400,0.12500}%
\definecolor{mycolor5}{rgb}{0.30100,0.74500,0.93300}%
\definecolor{mycolor6}{rgb}{0.85000,0.32500,0.09800}%
\begin{tikzpicture}

\begin{axis}[%
width=4.464in,
height=3.286in,
at={(0.815in,0.76in)},
scale only axis,
xmin=0.1,
xmax=0.95,
xlabel style={font=\color{white!15!black}},
xlabel={Percentage of corrupted data},
ymin=0,
ymax=1,
ylabel style={font=\color{white!15!black}},
ylabel={Accuracy},
axis background/.style={fill=white},
axis x line*=bottom,
axis y line*=left,
legend style={at={(0.2,0.03)}, legend cell align=left, align=left, anchor = south west, draw=white!15!black}
]

\addplot[area legend, draw=none, fill=mycolor1, fill opacity=0.2, forget plot]
table[row sep=crcr] {%
x	y\\
0.1	0.947286754492703\\
0.2	0.923776084930886\\
0.3	0.906957564630182\\
0.4	0.897375953575627\\
0.5	0.882862534613745\\
0.6	0.879674830775825\\
0.7	0.882104973732768\\
0.8	0.89213033822062\\
0.9	0.898338402555756\\
0.9	0.926999680070659\\
0.8	0.923510473262734\\
0.7	0.905013231387172\\
0.6	0.899907962598406\\
0.5	0.901254567276664\\
0.4	0.907482649343234\\
0.3	0.918370104728685\\
0.2	0.933570080752299\\
0.1	0.955557921345359\\
}--cycle;
\addplot [color=white!55!mycolor1, forget plot]
  table[row sep=crcr]{%
0.1	0.947286754492703\\
0.2	0.923776084930886\\
0.3	0.906957564630182\\
0.4	0.897375953575627\\
0.5	0.882862534613745\\
0.6	0.879674830775825\\
0.7	0.882104973732768\\
0.8	0.89213033822062\\
0.9	0.898338402555756\\
};
\addplot [color=white!55!mycolor1, forget plot]
  table[row sep=crcr]{%
0.1	0.955557921345359\\
0.2	0.933570080752299\\
0.3	0.918370104728685\\
0.4	0.907482649343234\\
0.5	0.901254567276664\\
0.6	0.899907962598406\\
0.7	0.905013231387172\\
0.8	0.923510473262734\\
0.9	0.926999680070659\\
};
\addplot [color=mycolor1, line width=1.0pt, mark size=5.0pt, mark=o, mark options={solid, mycolor1}]
  table[row sep=crcr]{%
0.1	0.951422337919031\\
0.2	0.928673082841593\\
0.3	0.912663834679434\\
0.4	0.90242930145943\\
0.5	0.892058550945204\\
0.6	0.889791396687115\\
0.7	0.89355910255997\\
0.8	0.907820405741677\\
0.9	0.912669041313207\\
};
\addlegendentry{Alg. 1 - TA + DS}

\addplot[area legend, draw=none, fill=mycolor2, fill opacity=0.2, forget plot]
table[row sep=crcr] {%
x	y\\
0.1	0.947286754492703\\
0.2	0.923776084930886\\
0.3	0.906957564630182\\
0.4	0.897375953575627\\
0.5	0.882862534613745\\
0.6	0.879674830775825\\
0.7	0.882104973732768\\
0.8	0.89213033822062\\
0.9	0.898338402555756\\
0.9	0.926999680070659\\
0.8	0.923510473262734\\
0.7	0.905013231387172\\
0.6	0.899907962598406\\
0.5	0.901254567276664\\
0.4	0.907482649343234\\
0.3	0.918370104728685\\
0.2	0.933570080752299\\
0.1	0.955557921345359\\
}--cycle;
\addplot [color=white!55!mycolor2, forget plot]
  table[row sep=crcr]{%
0.1	0.947286754492703\\
0.2	0.923776084930886\\
0.3	0.906957564630182\\
0.4	0.897375953575627\\
0.5	0.882862534613745\\
0.6	0.879674830775825\\
0.7	0.882104973732768\\
0.8	0.89213033822062\\
0.9	0.898338402555756\\
};
\addplot [color=white!55!mycolor2, forget plot]
  table[row sep=crcr]{%
0.1	0.955557921345359\\
0.2	0.933570080752299\\
0.3	0.918370104728685\\
0.4	0.907482649343234\\
0.5	0.901254567276664\\
0.6	0.899907962598406\\
0.7	0.905013231387172\\
0.8	0.923510473262734\\
0.9	0.926999680070659\\
};
\addplot [color=mycolor2, line width=1.0pt, mark size=8.6pt, mark=diamond, mark options={solid, mycolor2}]
  table[row sep=crcr]{%
0.1	0.951422337919031\\
0.2	0.928673082841593\\
0.3	0.912663834679434\\
0.4	0.90242930145943\\
0.5	0.892058550945204\\
0.6	0.889791396687115\\
0.7	0.89355910255997\\
0.8	0.907820405741677\\
0.9	0.912669041313207\\
};
\addlegendentry{Alg. 1 - H + DS}

\addplot[area legend, draw=none, fill=mycolor3, fill opacity=0.2, forget plot]
table[row sep=crcr] {%
x	y\\
0.1	0.897247539941576\\
0.2	0.803088331548328\\
0.3	0.71725856407451\\
0.4	0.641487906489047\\
0.5	0.573032409765167\\
0.6	0.484853920204439\\
0.7	0.528480285900119\\
0.8	0.629887764927276\\
0.9	0.723839863156261\\
0.9	0.759867183365336\\
0.8	0.683679266066515\\
0.7	0.644995693627586\\
0.6	0.576895972864848\\
0.5	0.602602442790148\\
0.4	0.657991590722388\\
0.3	0.724785760167466\\
0.2	0.806229408160001\\
0.1	0.899511453796179\\
}--cycle;
\addplot [color=white!55!mycolor3, forget plot]
  table[row sep=crcr]{%
0.1	0.897247539941576\\
0.2	0.803088331548328\\
0.3	0.71725856407451\\
0.4	0.641487906489047\\
0.5	0.573032409765167\\
0.6	0.484853920204439\\
0.7	0.528480285900119\\
0.8	0.629887764927276\\
0.9	0.723839863156261\\
};
\addplot [color=white!55!mycolor3, forget plot]
  table[row sep=crcr]{%
0.1	0.899511453796179\\
0.2	0.806229408160001\\
0.3	0.724785760167466\\
0.4	0.657991590722388\\
0.5	0.602602442790148\\
0.6	0.576895972864848\\
0.7	0.644995693627586\\
0.8	0.683679266066515\\
0.9	0.759867183365336\\
};
\addplot [color=mycolor3, dashed, line width=1.0pt, mark size=5.0pt, mark=x, mark options={solid, mycolor3}]
  table[row sep=crcr]{%
0.1	0.898379496868877\\
0.2	0.804658869854165\\
0.3	0.721022162120988\\
0.4	0.649739748605717\\
0.5	0.587817426277658\\
0.6	0.530874946534643\\
0.7	0.586737989763852\\
0.8	0.656783515496896\\
0.9	0.741853523260798\\
};
\addlegendentry{MMSR}

\addplot[area legend, draw=none, fill=mycolor4, fill opacity=0.2, forget plot]
table[row sep=crcr] {%
x	y\\
0.1	0.898359132890458\\
0.2	0.799187774458915\\
0.3	0.700342092231063\\
0.4	0.600503663083736\\
0.5	0.499080192514041\\
0.6	0.397055076841599\\
0.7	0.297786707948949\\
0.8	0.197242221883126\\
0.9	0.107834521945002\\
0.9	0.345486157682613\\
0.8	0.199586769052678\\
0.7	0.299889263652041\\
0.6	0.399845654298515\\
0.5	0.50183073437765\\
0.4	0.602835861418794\\
0.3	0.702855674213125\\
0.2	0.801661070467735\\
0.1	0.900922287655903\\
}--cycle;
\addplot [color=white!55!mycolor4, forget plot]
  table[row sep=crcr]{%
0.1	0.898359132890458\\
0.2	0.799187774458915\\
0.3	0.700342092231063\\
0.4	0.600503663083736\\
0.5	0.499080192514041\\
0.6	0.397055076841599\\
0.7	0.297786707948949\\
0.8	0.197242221883126\\
0.9	0.107834521945002\\
};
\addplot [color=white!55!mycolor4, forget plot]
  table[row sep=crcr]{%
0.1	0.900922287655903\\
0.2	0.801661070467735\\
0.3	0.702855674213125\\
0.4	0.602835861418794\\
0.5	0.50183073437765\\
0.6	0.399845654298515\\
0.7	0.299889263652041\\
0.8	0.199586769052678\\
0.9	0.345486157682613\\
};
\addplot [color=mycolor4, dashdotted, line width=1.0pt, mark size=3.5pt, mark=square, mark options={solid, mycolor4}]
  table[row sep=crcr]{%
0.1	0.89964071027318\\
0.2	0.800424422463325\\
0.3	0.701598883222094\\
0.4	0.601669762251265\\
0.5	0.500455463445846\\
0.6	0.398450365570057\\
0.7	0.298837985800495\\
0.8	0.198414495467902\\
0.9	0.226660339813807\\
};
\addlegendentry{DS}

\addplot[area legend, draw=none, fill=mycolor5, fill opacity=0.2, forget plot]
table[row sep=crcr] {%
x	y\\
0.1	0.896149665399658\\
0.2	0.832983517853086\\
0.3	0.768460200556522\\
0.4	0.701533107218889\\
0.5	0.63875755255402\\
0.6	0.575272649790861\\
0.7	0.512087049490818\\
0.8	0.444527596612746\\
0.9	0.38093342600406\\
0.9	0.40450018248525\\
0.8	0.466946883958436\\
0.7	0.526759502429289\\
0.6	0.591170259879741\\
0.5	0.653450625319141\\
0.4	0.715104992210438\\
0.3	0.777230893980142\\
0.2	0.841144959005037\\
0.1	0.905294001742913\\
}--cycle;
\addplot [color=white!55!mycolor5, forget plot]
  table[row sep=crcr]{%
0.1	0.896149665399658\\
0.2	0.832983517853086\\
0.3	0.768460200556522\\
0.4	0.701533107218889\\
0.5	0.63875755255402\\
0.6	0.575272649790861\\
0.7	0.512087049490818\\
0.8	0.444527596612746\\
0.9	0.38093342600406\\
};
\addplot [color=white!55!mycolor5, forget plot]
  table[row sep=crcr]{%
0.1	0.905294001742913\\
0.2	0.841144959005037\\
0.3	0.777230893980142\\
0.4	0.715104992210438\\
0.5	0.653450625319141\\
0.6	0.591170259879741\\
0.7	0.526759502429289\\
0.8	0.466946883958436\\
0.9	0.40450018248525\\
};
\addplot [color=mycolor5, dashdotted, line width=1.0pt, mark size=5.0pt, mark=+, mark options={solid, mycolor5}]
  table[row sep=crcr]{%
0.1	0.900721833571286\\
0.2	0.837064238429061\\
0.3	0.772845547268332\\
0.4	0.708319049714664\\
0.5	0.646104088936581\\
0.6	0.583221454835301\\
0.7	0.519423275960054\\
0.8	0.455737240285591\\
0.9	0.392716804244655\\
};
\addlegendentry{MV}

\addplot[area legend, draw=none, fill=mycolor6, fill opacity=0.2, forget plot]
table[row sep=crcr] {%
x	y\\
0.1	0.897002438669265\\
0.2	0.798067385796057\\
0.3	0.699982802137042\\
0.4	0.601421557555641\\
0.5	0.503388433542021\\
0.6	0.404010836159368\\
0.7	0.306457041921902\\
0.8	0.207623689106083\\
0.9	0.109472953910261\\
0.9	0.111801543282984\\
0.8	0.210926799533493\\
0.7	0.309755552464548\\
0.6	0.407344224736536\\
0.5	0.505590057804361\\
0.4	0.603599620740263\\
0.3	0.702334014906861\\
0.2	0.80089941251645\\
0.1	0.899276062471667\\
}--cycle;
\addplot [color=white!55!mycolor6, forget plot]
  table[row sep=crcr]{%
0.1	0.897002438669265\\
0.2	0.798067385796057\\
0.3	0.699982802137042\\
0.4	0.601421557555641\\
0.5	0.503388433542021\\
0.6	0.404010836159368\\
0.7	0.306457041921902\\
0.8	0.207623689106083\\
0.9	0.109472953910261\\
};
\addplot [color=white!55!mycolor6, forget plot]
  table[row sep=crcr]{%
0.1	0.899276062471667\\
0.2	0.80089941251645\\
0.3	0.702334014906861\\
0.4	0.603599620740263\\
0.5	0.505590057804361\\
0.6	0.407344224736536\\
0.7	0.309755552464548\\
0.8	0.210926799533493\\
0.9	0.111801543282984\\
};
\addplot [color=mycolor6, dashdotted, line width=1.0pt, mark size=3.3pt, mark=triangle, mark options={solid, mycolor6}]
  table[row sep=crcr]{%
0.1	0.898139250570466\\
0.2	0.799483399156253\\
0.3	0.701158408521951\\
0.4	0.602510589147952\\
0.5	0.504489245673191\\
0.6	0.405677530447952\\
0.7	0.308106297193225\\
0.8	0.209275244319788\\
0.9	0.110637248596622\\
};
\addlegendentry{PGD}

\end{axis}
\end{tikzpicture}%

%% file: fig/tikz/synthetic_data/per_adv_synth_data_acc_DS_shaded_acc.tikz
\definecolor{mycolor1}{rgb}{0.00000,0.44700,0.74100}%
\definecolor{mycolor2}{rgb}{0.49400,0.18400,0.55600}%
\definecolor{mycolor3}{rgb}{0.63500,0.07800,0.18400}%
\definecolor{mycolor4}{rgb}{0.92900,0.69400,0.12500}%
\definecolor{mycolor5}{rgb}{0.30100,0.74500,0.93300}%
\definecolor{mycolor6}{rgb}{0.85000,0.32500,0.09800}%
\begin{tikzpicture}

\begin{axis}[%
width=4.464in,
height=3.286in,
at={(0.815in,0.76in)},
scale only axis,
xmin=0,
xmax=0.95,
xlabel style={font=\color{white!15!black}},
xlabel={Percentage of adversaries},
ymin=0,
ymax=1,
ylabel style={font=\color{white!15!black}},
ylabel={Accuracy},
axis background/.style={fill=white},
axis x line*=bottom,
axis y line*=left,
legend style={at={(0.03,0.03)}, anchor=south west, legend cell align=left, align=left, draw=white!15!black}
]

\addplot[area legend, draw=none, fill=mycolor1, fill opacity=0.2, forget plot]
table[row sep=crcr] {%
x	y\\
0	0.895889668075561\\
0.1	0.887355959023496\\
0.2	0.859968024196585\\
0.3	0.411912408214856\\
0.4	0.849912830996197\\
0.5	0.827621261266283\\
0.6	0.776449093390435\\
0.7	0.743374010415199\\
0.8	0.673741139907675\\
0.9	0.219775341225525\\
0.9	0.495385194485723\\
0.8	0.711224273368481\\
0.7	0.753170119661155\\
0.6	0.798391095194348\\
0.5	0.843057762180805\\
0.4	0.868452449009507\\
0.3	0.981376304467825\\
0.2	0.884456033430722\\
0.1	0.901090108141977\\
0	0.901060246687996\\
}--cycle;
\addplot [color=white!55!mycolor1, forget plot]
  table[row sep=crcr]{%
0	0.895889668075561\\
0.1	0.887355959023496\\
0.2	0.859968024196585\\
0.3	0.411912408214856\\
0.4	0.849912830996197\\
0.5	0.827621261266283\\
0.6	0.776449093390435\\
0.7	0.743374010415199\\
0.8	0.673741139907675\\
0.9	0.219775341225525\\
};
\addplot [color=white!55!mycolor1, forget plot]
  table[row sep=crcr]{%
0	0.901060246687996\\
0.1	0.901090108141977\\
0.2	0.884456033430722\\
0.3	0.981376304467825\\
0.4	0.868452449009507\\
0.5	0.843057762180805\\
0.6	0.798391095194348\\
0.7	0.753170119661155\\
0.8	0.711224273368481\\
0.9	0.495385194485723\\
};
\addplot [color=mycolor1, line width=1.0pt, mark size=5.0pt, mark=o, mark options={solid, mycolor1}]
  table[row sep=crcr]{%
0	0.898474957381778\\
0.1	0.894223033582736\\
0.2	0.872212028813654\\
0.3	0.69664435634134\\
0.4	0.859182640002852\\
0.5	0.835339511723544\\
0.6	0.787420094292392\\
0.7	0.748272065038177\\
0.8	0.692482706638078\\
0.9	0.357580267855624\\
};
\addlegendentry{Alg. 1 - TA + DS}

\addplot[area legend, draw=none, fill=mycolor2, fill opacity=0.2, forget plot]
table[row sep=crcr] {%
x	y\\
0	0.895889668075561\\
0.1	0.887355959023496\\
0.2	0.859968024196585\\
0.3	0.411912408214856\\
0.4	0.849912830996197\\
0.5	0.186475993195491\\
0.6	0.151765670488481\\
0.7	0.131699535664106\\
0.8	0.136006315770672\\
0.9	0.194836531533281\\
0.9	0.456378047561204\\
0.8	0.366817004302007\\
0.7	0.274189238099356\\
0.6	0.338748843960299\\
0.5	0.385567018997057\\
0.4	0.868452449009507\\
0.3	0.981376304467825\\
0.2	0.884456033430722\\
0.1	0.901090108141977\\
0	0.901060246687996\\
}--cycle;
\addplot [color=white!55!mycolor2, forget plot]
  table[row sep=crcr]{%
0	0.895889668075561\\
0.1	0.887355959023496\\
0.2	0.859968024196585\\
0.3	0.411912408214856\\
0.4	0.849912830996197\\
0.5	0.186475993195491\\
0.6	0.151765670488481\\
0.7	0.131699535664106\\
0.8	0.136006315770672\\
0.9	0.194836531533281\\
};
\addplot [color=white!55!mycolor2, forget plot]
  table[row sep=crcr]{%
0	0.901060246687996\\
0.1	0.901090108141977\\
0.2	0.884456033430722\\
0.3	0.981376304467825\\
0.4	0.868452449009507\\
0.5	0.385567018997057\\
0.6	0.338748843960299\\
0.7	0.274189238099356\\
0.8	0.366817004302007\\
0.9	0.456378047561204\\
};
\addplot [color=mycolor2, line width=1.0pt, mark size=8.6pt, mark=diamond, mark options={solid, mycolor2}]
  table[row sep=crcr]{%
0	0.898474957381778\\
0.1	0.894223033582736\\
0.2	0.872212028813654\\
0.3	0.69664435634134\\
0.4	0.859182640002852\\
0.5	0.286021506096274\\
0.6	0.24525725722439\\
0.7	0.202944386881731\\
0.8	0.25141166003634\\
0.9	0.325607289547242\\
};
\addlegendentry{Alg. 1 - H + DS}

\addplot[area legend, draw=none, fill=mycolor3, fill opacity=0.2, forget plot]
table[row sep=crcr] {%
x	y\\
0	0.799182049156104\\
0.1	0.778159677373171\\
0.2	0.714531935652835\\
0.3	0.611121381142214\\
0.4	0.489416751410542\\
0.5	0.232874360302017\\
0.6	0.324566790299856\\
0.7	0.34251489067404\\
0.8	0.362415589958828\\
0.9	0.375268645279676\\
0.9	0.400578045617201\\
0.8	0.386478398100225\\
0.7	0.366802481988264\\
0.6	0.349188704426745\\
0.5	0.336265447438688\\
0.4	0.540364760832682\\
0.3	0.671590308107119\\
0.2	0.743754174812843\\
0.1	0.792885565479073\\
0	0.812330457620152\\
}--cycle;
\addplot [color=white!55!mycolor3, forget plot]
  table[row sep=crcr]{%
0	0.799182049156104\\
0.1	0.778159677373171\\
0.2	0.714531935652835\\
0.3	0.611121381142214\\
0.4	0.489416751410542\\
0.5	0.232874360302017\\
0.6	0.324566790299856\\
0.7	0.34251489067404\\
0.8	0.362415589958828\\
0.9	0.375268645279676\\
};
\addplot [color=white!55!mycolor3, forget plot]
  table[row sep=crcr]{%
0	0.812330457620152\\
0.1	0.792885565479073\\
0.2	0.743754174812843\\
0.3	0.671590308107119\\
0.4	0.540364760832682\\
0.5	0.336265447438688\\
0.6	0.349188704426745\\
0.7	0.366802481988264\\
0.8	0.386478398100225\\
0.9	0.400578045617201\\
};
\addplot [color=mycolor3, dashed, line width=1.0pt, mark size=5.0pt, mark=x, mark options={solid, mycolor3}]
  table[row sep=crcr]{%
0	0.805756253388128\\
0.1	0.785522621426122\\
0.2	0.729143055232839\\
0.3	0.641355844624667\\
0.4	0.514890756121612\\
0.5	0.284569903870353\\
0.6	0.336877747363301\\
0.7	0.354658686331152\\
0.8	0.374446994029526\\
0.9	0.387923345448438\\
};
\addlegendentry{MMSR}

\addplot[area legend, draw=none, fill=mycolor4, fill opacity=0.2, forget plot]
table[row sep=crcr] {%
x	y\\
0	0.895889668075561\\
0.1	0.890094527303322\\
0.2	0.860424121760835\\
0.3	0.230531601300086\\
0.4	0.178598735020307\\
0.5	0.154842599797698\\
0.6	0.141883924449337\\
0.7	0.130276961398988\\
0.8	0.134662207806079\\
0.9	0.194711860580839\\
0.9	0.456422229402715\\
0.8	0.365718974028936\\
0.7	0.271568080112871\\
0.6	0.304507802922739\\
0.5	0.261613825385508\\
0.4	0.301814095265474\\
0.3	0.335861038076693\\
0.2	0.881155954292268\\
0.1	0.901155681975041\\
0	0.901060246687996\\
}--cycle;
\addplot [color=white!55!mycolor4, forget plot]
  table[row sep=crcr]{%
0	0.895889668075561\\
0.1	0.890094527303322\\
0.2	0.860424121760835\\
0.3	0.230531601300086\\
0.4	0.178598735020307\\
0.5	0.154842599797698\\
0.6	0.141883924449337\\
0.7	0.130276961398988\\
0.8	0.134662207806079\\
0.9	0.194711860580839\\
};
\addplot [color=white!55!mycolor4, forget plot]
  table[row sep=crcr]{%
0	0.901060246687996\\
0.1	0.901155681975041\\
0.2	0.881155954292268\\
0.3	0.335861038076693\\
0.4	0.301814095265474\\
0.5	0.261613825385508\\
0.6	0.304507802922739\\
0.7	0.271568080112871\\
0.8	0.365718974028936\\
0.9	0.456422229402715\\
};
\addplot [color=mycolor4, dashdotted, line width=1.0pt, mark size=3.5pt, mark=square, mark options={solid, mycolor4}]
  table[row sep=crcr]{%
0	0.898474957381778\\
0.1	0.895625104639181\\
0.2	0.870790038026552\\
0.3	0.28319631968839\\
0.4	0.240206415142891\\
0.5	0.208228212591603\\
0.6	0.223195863686038\\
0.7	0.200922520755929\\
0.8	0.250190590917507\\
0.9	0.325567044991777\\
};
\addlegendentry{DS}

\addplot[area legend, draw=none, fill=mycolor5, fill opacity=0.2, forget plot]
table[row sep=crcr] {%
x	y\\
0	0.783829101155156\\
0.1	0.729069847037975\\
0.2	0.6419475335385\\
0.3	0.556369987903497\\
0.4	0.482955276222855\\
0.5	0.432172519006474\\
0.6	0.399384930150389\\
0.7	0.383509337618996\\
0.8	0.384197928710046\\
0.9	0.379933939386049\\
0.9	0.399316379517421\\
0.8	0.404573562622193\\
0.7	0.400115390997413\\
0.6	0.410590028023296\\
0.5	0.447331013154746\\
0.4	0.499069107756851\\
0.3	0.572553566406561\\
0.2	0.662341832057533\\
0.1	0.738596495569258\\
0	0.793014149209595\\
}--cycle;
\addplot [color=white!55!mycolor5, forget plot]
  table[row sep=crcr]{%
0	0.783829101155156\\
0.1	0.729069847037975\\
0.2	0.6419475335385\\
0.3	0.556369987903497\\
0.4	0.482955276222855\\
0.5	0.432172519006474\\
0.6	0.399384930150389\\
0.7	0.383509337618996\\
0.8	0.384197928710046\\
0.9	0.379933939386049\\
};
\addplot [color=white!55!mycolor5, forget plot]
  table[row sep=crcr]{%
0	0.793014149209595\\
0.1	0.738596495569258\\
0.2	0.662341832057533\\
0.3	0.572553566406561\\
0.4	0.499069107756851\\
0.5	0.447331013154746\\
0.6	0.410590028023296\\
0.7	0.400115390997413\\
0.8	0.404573562622193\\
0.9	0.399316379517421\\
};
\addplot [color=mycolor5, dashdotted, line width=1.0pt, mark size=5.0pt, mark=+, mark options={solid, mycolor5}]
  table[row sep=crcr]{%
0	0.788421625182375\\
0.1	0.733833171303616\\
0.2	0.652144682798017\\
0.3	0.564461777155029\\
0.4	0.491012191989853\\
0.5	0.43975176608061\\
0.6	0.404987479086842\\
0.7	0.391812364308204\\
0.8	0.394385745666119\\
0.9	0.389625159451735\\
};
\addlegendentry{MV}

\addplot[area legend, draw=none, fill=mycolor6, fill opacity=0.2, forget plot]
table[row sep=crcr] {%
x	y\\
0	0.802941150761386\\
0.1	0.722581405584181\\
0.2	0.447803704306304\\
0.3	0.357537260170879\\
0.4	0.351962673997668\\
0.5	0.354934522675487\\
0.6	0.362406208612225\\
0.7	0.366055047729737\\
0.8	0.376986615589958\\
0.9	0.382423152753895\\
0.9	0.405235304057099\\
0.8	0.400013688057631\\
0.7	0.386982450398885\\
0.6	0.38228145729591\\
0.5	0.373017820834024\\
0.4	0.371134305289826\\
0.3	0.376846032290979\\
0.2	0.465410218845732\\
0.1	0.74480273776404\\
0	0.81293516022619\\
}--cycle;
\addplot [color=white!55!mycolor6, forget plot]
  table[row sep=crcr]{%
0	0.802941150761386\\
0.1	0.722581405584181\\
0.2	0.447803704306304\\
0.3	0.357537260170879\\
0.4	0.351962673997668\\
0.5	0.354934522675487\\
0.6	0.362406208612225\\
0.7	0.366055047729737\\
0.8	0.376986615589958\\
0.9	0.382423152753895\\
};
\addplot [color=white!55!mycolor6, forget plot]
  table[row sep=crcr]{%
0	0.81293516022619\\
0.1	0.74480273776404\\
0.2	0.465410218845732\\
0.3	0.376846032290979\\
0.4	0.371134305289826\\
0.5	0.373017820834024\\
0.6	0.38228145729591\\
0.7	0.386982450398885\\
0.8	0.400013688057631\\
0.9	0.405235304057099\\
};
\addplot [color=mycolor6, dashdotted, line width=1.0pt, mark size=3.3pt, mark=triangle, mark options={solid, mycolor6}]
  table[row sep=crcr]{%
0	0.807938155493788\\
0.1	0.73369207167411\\
0.2	0.456606961576018\\
0.3	0.367191646230929\\
0.4	0.361548489643747\\
0.5	0.363976171754755\\
0.6	0.372343832954067\\
0.7	0.376518749064311\\
0.8	0.388500151823795\\
0.9	0.393829228405497\\
};
\addlegendentry{PGD}

\end{axis}
\end{tikzpicture}%

%% file: fig/tikz/synthetic_data/per_adv_synth_data_detect_shaded_DS.tikz
\definecolor{mycolor1}{rgb}{0.00000,0.44700,0.74100}%
\definecolor{mycolor2}{rgb}{0.49400,0.18400,0.55600}%
\definecolor{mycolor3}{rgb}{0.63500,0.07800,0.18400}%
\definecolor{mycolor4}{rgb}{0.92900,0.69400,0.12500}%
\definecolor{mycolor5}{rgb}{0.30100,0.74500,0.93300}%
\begin{tikzpicture}

\begin{axis}[%
width=4.521in,
height=3.286in,
at={(0.758in,0.76in)},
scale only axis,
xmin=0,
xmax=0.95,
xlabel style={font=\color{white!15!black}},
xlabel={Percentage of adversaries},
ymin=0,
ymax=1,
axis background/.style={fill=white},
axis x line*=bottom,
axis y line*=left,
legend style={at={(0.97,0.3)}, anchor=south east, legend cell align=left, align=left, draw=white!15!black}
]

\addplot[area legend, draw=none, fill=mycolor1, fill opacity=0.2, forget plot]
table[row sep=crcr] {%
x	y\\
0	0\\
0.1	1\\
0.2	-0.216227766016838\\
0.3	0.216954108460352\\
0.4	1\\
0.5	0.8\\
0.6	0\\
0.7	0\\
0.8	0\\
0.9	0\\
0.9	0\\
0.8	0\\
0.7	0\\
0.6	0\\
0.5	0.8\\
0.4	1\\
0.3	1.18304589153965\\
0.2	0.416227766016838\\
0.1	1\\
0	0\\
}--cycle;
\addplot [color=white!55!mycolor1, forget plot]
  table[row sep=crcr]{%
0	0\\
0.1	1\\
0.2	-0.216227766016838\\
0.3	0.216954108460352\\
0.4	1\\
0.5	0.8\\
0.6	0\\
0.7	0\\
0.8	0\\
0.9	0\\
};
\addplot [color=white!55!mycolor1, forget plot]
  table[row sep=crcr]{%
0	0\\
0.1	1\\
0.2	0.416227766016838\\
0.3	1.18304589153965\\
0.4	1\\
0.5	0.8\\
0.6	0\\
0.7	0\\
0.8	0\\
0.9	0\\
};
\addplot [color=mycolor1, line width=1.0pt, mark size=5.0pt, mark=star, mark options={solid, mycolor1}]
  table[row sep=crcr]{%
0	0\\
0.1	1\\
0.2	0.1\\
0.3	0.7\\
0.4	1\\
0.5	0.8\\
0.6	0\\
0.7	0\\
0.8	0\\
0.9	0\\
};
\addlegendentry{Sensitivity - Alg. 1 - H}

\addplot[area legend, draw=none, fill=mycolor2, fill opacity=0.2, forget plot]
table[row sep=crcr] {%
x	y\\
0	1\\
0.1	0.976876235221287\\
0.2	1\\
0.3	1\\
0.4	1\\
0.5	1\\
0.6	0\\
0.7	0\\
0.8	0\\
0.9	0\\
0.9	0\\
0.8	0\\
0.7	0\\
0.6	0\\
0.5	1\\
0.4	1\\
0.3	1\\
0.2	1\\
0.1	1.0120126536676\\
0	1\\
}--cycle;
\addplot [color=white!55!mycolor2, forget plot]
  table[row sep=crcr]{%
0	1\\
0.1	0.976876235221287\\
0.2	1\\
0.3	1\\
0.4	1\\
0.5	1\\
0.6	0\\
0.7	0\\
0.8	0\\
0.9	0\\
};
\addplot [color=white!55!mycolor2, forget plot]
  table[row sep=crcr]{%
0	1\\
0.1	1.0120126536676\\
0.2	1\\
0.3	1\\
0.4	1\\
0.5	1\\
0.6	0\\
0.7	0\\
0.8	0\\
0.9	0\\
};
\addplot [color=mycolor2, line width=1.0pt, mark size=3.3pt, mark=triangle, mark options={solid, rotate=270, mycolor2}]
  table[row sep=crcr]{%
0	1\\
0.1	0.994444444444444\\
0.2	1\\
0.3	1\\
0.4	1\\
0.5	1\\
0.6	0\\
0.7	0\\
0.8	0\\
0.9	0\\
};
\addlegendentry{Specificity - Alg. 1 - H}

\addplot[area legend, draw=none, fill=mycolor3, fill opacity=0.2, forget plot]
table[row sep=crcr] {%
x	y\\
0	0\\
0.1	1\\
0.2	-0.216227766016838\\
0.3	0.216954108460352\\
0.4	1\\
0.5	1\\
0.6	1\\
0.7	1\\
0.8	1\\
0.9	1\\
0.9	1\\
0.8	1\\
0.7	1\\
0.6	1\\
0.5	1\\
0.4	1\\
0.3	1.18304589153965\\
0.2	0.416227766016838\\
0.1	1\\
0	0\\
}--cycle;
\addplot [color=white!55!mycolor3, forget plot]
  table[row sep=crcr]{%
0	0\\
0.1	1\\
0.2	-0.216227766016838\\
0.3	0.216954108460352\\
0.4	1\\
0.5	1\\
0.6	1\\
0.7	1\\
0.8	1\\
0.9	1\\
};
\addplot [color=white!55!mycolor3, forget plot]
  table[row sep=crcr]{%
0	0\\
0.1	1\\
0.2	0.416227766016838\\
0.3	1.18304589153965\\
0.4	1\\
0.5	1\\
0.6	1\\
0.7	1\\
0.8	1\\
0.9	1\\
};
\addplot [color=mycolor3, dashed, line width=1.0pt, mark size=5.0pt, mark=pentagon, mark options={solid, mycolor3}]
  table[row sep=crcr]{%
0	0\\
0.1	1\\
0.2	0.1\\
0.3	0.7\\
0.4	1\\
0.5	1\\
0.6	1\\
0.7	1\\
0.8	1\\
0.9	1\\
};
\addlegendentry{Sensitivity - Alg. 1 - TA}

\addplot[area legend, draw=none, fill=mycolor4, fill opacity=0.2, forget plot]
table[row sep=crcr] {%
x	y\\
0	1\\
0.1	0.976876235221287\\
0.2	1\\
0.3	1\\
0.4	1\\
0.5	1\\
0.6	1\\
0.7	1\\
0.8	1\\
0.9	1\\
0.9	1\\
0.8	1\\
0.7	1\\
0.6	1\\
0.5	1\\
0.4	1\\
0.3	1\\
0.2	1\\
0.1	1.0120126536676\\
0	1\\
}--cycle;
\addplot [color=white!55!mycolor4, forget plot]
  table[row sep=crcr]{%
0	1\\
0.1	0.976876235221287\\
0.2	1\\
0.3	1\\
0.4	1\\
0.5	1\\
0.6	1\\
0.7	1\\
0.8	1\\
0.9	1\\
};
\addplot [color=white!55!mycolor4, forget plot]
  table[row sep=crcr]{%
0	1\\
0.1	1.0120126536676\\
0.2	1\\
0.3	1\\
0.4	1\\
0.5	1\\
0.6	1\\
0.7	1\\
0.8	1\\
0.9	1\\
};
\addplot [color=mycolor4, dashed, line width=1.0pt, mark size=3.3pt, mark=triangle, mark options={solid, rotate=90, mycolor4}]
  table[row sep=crcr]{%
0	1\\
0.1	0.994444444444444\\
0.2	1\\
0.3	1\\
0.4	1\\
0.5	1\\
0.6	1\\
0.7	1\\
0.8	1\\
0.9	1\\
};
\addlegendentry{Specificity - Alg. 1 - TA}

\addplot[area legend, draw=none, fill=mycolor5, fill opacity=0.2, forget plot]
table[row sep=crcr] {%
x	y\\
0	1\\
0.1	0.979188611699158\\
0.2	0.756754446796632\\
0.3	0.765086232538105\\
0.4	1\\
0.5	1\\
0.6	1\\
0.7	1\\
0.8	1\\
0.9	1\\
0.9	1\\
0.8	1\\
0.7	1\\
0.6	1\\
0.5	1\\
0.4	1\\
0.3	1.05491376746189\\
0.2	0.883245553203368\\
0.1	1.01081138830084\\
0	1\\
}--cycle;
\addplot [color=white!55!mycolor5, forget plot]
  table[row sep=crcr]{%
0	1\\
0.1	0.979188611699158\\
0.2	0.756754446796632\\
0.3	0.765086232538105\\
0.4	1\\
0.5	1\\
0.6	1\\
0.7	1\\
0.8	1\\
0.9	1\\
};
\addplot [color=white!55!mycolor5, forget plot]
  table[row sep=crcr]{%
0	1\\
0.1	1.01081138830084\\
0.2	0.883245553203368\\
0.3	1.05491376746189\\
0.4	1\\
0.5	1\\
0.6	1\\
0.7	1\\
0.8	1\\
0.9	1\\
};
\addplot [color=mycolor5, dashdotted, line width=1.0pt, mark size=3.3pt, mark=triangle, mark options={solid, rotate=180, mycolor5}]
  table[row sep=crcr]{%
0	1\\
0.1	0.995\\
0.2	0.82\\
0.3	0.91\\
0.4	1\\
0.5	1\\
0.6	1\\
0.7	1\\
0.8	1\\
0.9	1\\
};
\addlegendentry{Clustering Accuracy}

\end{axis}
\end{tikzpicture}%

%% file: fig/tikz/synthetic_data/p_adv_synth_data_acc_DS_shaded_acc.tikz
\definecolor{mycolor1}{rgb}{0.00000,0.44700,0.74100}%
\definecolor{mycolor2}{rgb}{0.49400,0.18400,0.55600}%
\definecolor{mycolor3}{rgb}{0.63500,0.07800,0.18400}%
\definecolor{mycolor4}{rgb}{0.92900,0.69400,0.12500}%
\definecolor{mycolor5}{rgb}{0.30100,0.74500,0.93300}%
\definecolor{mycolor6}{rgb}{0.85000,0.32500,0.09800}%
\begin{tikzpicture}

\begin{axis}[%
width=4.464in,
height=3.286in,
at={(0.815in,0.76in)},
scale only axis,
xmin=0,
xmax=0.95,
xlabel style={font=\color{white!15!black}},
xlabel={Percentage of corrupted data},
ymin=0,
ymax=1,
ylabel style={font=\color{white!15!black}},
ylabel={Accuracy},
axis background/.style={fill=white},
axis x line*=bottom,
axis y line*=left,
legend style={at={(0.03,0.03)}, anchor=south west, legend cell align=left, align=left, draw=white!15!black}
]

\addplot[area legend, draw=none, fill=mycolor1, fill opacity=0.2, forget plot]
table[row sep=crcr] {%
x	y\\
0.1	0.862079181283053\\
0.2	0.78275119998231\\
0.3	0.224170991360747\\
0.4	0.154587203188578\\
0.5	0.862878973503041\\
0.6	0.877720981504678\\
0.7	0.880783219441504\\
0.8	0.892026241828537\\
0.9	0.903251366208034\\
0.9	0.948083106697528\\
0.8	0.921928853494825\\
0.7	0.898712450778871\\
0.6	0.894782535465425\\
0.5	0.877196847707369\\
0.4	0.668738323619568\\
0.3	0.524995513962743\\
0.2	0.907836562760281\\
0.1	0.875796537626989\\
}--cycle;
\addplot [color=white!55!mycolor1, forget plot]
  table[row sep=crcr]{%
0.1	0.862079181283053\\
0.2	0.78275119998231\\
0.3	0.224170991360747\\
0.4	0.154587203188578\\
0.5	0.862878973503041\\
0.6	0.877720981504678\\
0.7	0.880783219441504\\
0.8	0.892026241828537\\
0.9	0.903251366208034\\
};
\addplot [color=white!55!mycolor1, forget plot]
  table[row sep=crcr]{%
0.1	0.875796537626989\\
0.2	0.907836562760281\\
0.3	0.524995513962743\\
0.4	0.668738323619568\\
0.5	0.877196847707369\\
0.6	0.894782535465425\\
0.7	0.898712450778871\\
0.8	0.921928853494825\\
0.9	0.948083106697528\\
};
\addplot [color=mycolor1, line width=1.0pt, mark size=5.0pt, mark=o, mark options={solid, mycolor1}]
  table[row sep=crcr]{%
0.1	0.868937859455021\\
0.2	0.845293881371296\\
0.3	0.374583252661745\\
0.4	0.411662763404073\\
0.5	0.870037910605205\\
0.6	0.886251758485052\\
0.7	0.889747835110187\\
0.8	0.906977547661681\\
0.9	0.925667236452781\\
};
\addlegendentry{Alg. 1 - TA + DS}

\addplot[area legend, draw=none, fill=mycolor2, fill opacity=0.2, forget plot]
table[row sep=crcr] {%
x	y\\
0.1	0.862079181283053\\
0.2	0.78275119998231\\
0.3	0.224170991360747\\
0.4	0.154587203188578\\
0.5	0.862878973503041\\
0.6	0.877720981504678\\
0.7	0.880783219441504\\
0.8	0.892026241828537\\
0.9	0.903251366208034\\
0.9	0.948083106697528\\
0.8	0.921928853494825\\
0.7	0.898712450778871\\
0.6	0.894782535465425\\
0.5	0.877196847707369\\
0.4	0.668738323619568\\
0.3	0.524995513962743\\
0.2	0.907836562760281\\
0.1	0.875796537626989\\
}--cycle;
\addplot [color=white!55!mycolor2, forget plot]
  table[row sep=crcr]{%
0.1	0.862079181283053\\
0.2	0.78275119998231\\
0.3	0.224170991360747\\
0.4	0.154587203188578\\
0.5	0.862878973503041\\
0.6	0.877720981504678\\
0.7	0.880783219441504\\
0.8	0.892026241828537\\
0.9	0.903251366208034\\
};
\addplot [color=white!55!mycolor2, forget plot]
  table[row sep=crcr]{%
0.1	0.875796537626989\\
0.2	0.907836562760281\\
0.3	0.524995513962743\\
0.4	0.668738323619568\\
0.5	0.877196847707369\\
0.6	0.894782535465425\\
0.7	0.898712450778871\\
0.8	0.921928853494825\\
0.9	0.948083106697528\\
};
\addplot [color=mycolor2, line width=1.0pt, mark size=8.6pt, mark=diamond, mark options={solid, mycolor2}]
  table[row sep=crcr]{%
0.1	0.868937859455021\\
0.2	0.845293881371296\\
0.3	0.374583252661745\\
0.4	0.411662763404073\\
0.5	0.870037910605205\\
0.6	0.886251758485052\\
0.7	0.889747835110187\\
0.8	0.906977547661681\\
0.9	0.925667236452781\\
};
\addlegendentry{Alg. 1 - H + DS}

\addplot[area legend, draw=none, fill=mycolor3, fill opacity=0.2, forget plot]
table[row sep=crcr] {%
x	y\\
0.1	0.751957944482727\\
0.2	0.70736129706015\\
0.3	0.706189557409107\\
0.4	0.623453079225102\\
0.5	0.619663882383207\\
0.6	0.615217117570328\\
0.7	0.643741416790863\\
0.8	0.714009442432422\\
0.9	0.750999711584698\\
0.9	0.771817890171924\\
0.8	0.732254688295475\\
0.7	0.677997180294583\\
0.6	0.68162738545519\\
0.5	0.635574969107571\\
0.4	0.658705188329823\\
0.3	0.720641392595307\\
0.2	0.739797586063865\\
0.1	0.773839601723934\\
}--cycle;
\addplot [color=white!55!mycolor3, forget plot]
  table[row sep=crcr]{%
0.1	0.751957944482727\\
0.2	0.70736129706015\\
0.3	0.706189557409107\\
0.4	0.623453079225102\\
0.5	0.619663882383207\\
0.6	0.615217117570328\\
0.7	0.643741416790863\\
0.8	0.714009442432422\\
0.9	0.750999711584698\\
};
\addplot [color=white!55!mycolor3, forget plot]
  table[row sep=crcr]{%
0.1	0.773839601723934\\
0.2	0.739797586063865\\
0.3	0.720641392595307\\
0.4	0.658705188329823\\
0.5	0.635574969107571\\
0.6	0.68162738545519\\
0.7	0.677997180294583\\
0.8	0.732254688295475\\
0.9	0.771817890171924\\
};
\addplot [color=mycolor3, dashed, line width=1.0pt, mark size=5.0pt, mark=x, mark options={solid, mycolor3}]
  table[row sep=crcr]{%
0.1	0.762898773103331\\
0.2	0.723579441562007\\
0.3	0.713415475002207\\
0.4	0.641079133777463\\
0.5	0.627619425745389\\
0.6	0.648422251512759\\
0.7	0.660869298542723\\
0.8	0.723132065363948\\
0.9	0.761408800878311\\
};
\addlegendentry{MMSR}

\addplot[area legend, draw=none, fill=mycolor4, fill opacity=0.2, forget plot]
table[row sep=crcr] {%
x	y\\
0.1	0.862079181283053\\
0.2	0.78275119998231\\
0.3	0.224170991360747\\
0.4	0.244915508464108\\
0.5	0.193701897285362\\
0.6	0.143436009355271\\
0.7	0.114656101059046\\
0.8	0.170047826071254\\
0.9	0.0618216614594372\\
0.9	0.352585778978701\\
0.8	0.398486820363855\\
0.7	0.130360403864933\\
0.6	0.19193140632457\\
0.5	0.319580267419611\\
0.4	0.377448864959978\\
0.3	0.524995513962743\\
0.2	0.907836562760281\\
0.1	0.875796537626989\\
}--cycle;
\addplot [color=white!55!mycolor4, forget plot]
  table[row sep=crcr]{%
0.1	0.862079181283053\\
0.2	0.78275119998231\\
0.3	0.224170991360747\\
0.4	0.244915508464108\\
0.5	0.193701897285362\\
0.6	0.143436009355271\\
0.7	0.114656101059046\\
0.8	0.170047826071254\\
0.9	0.0618216614594372\\
};
\addplot [color=white!55!mycolor4, forget plot]
  table[row sep=crcr]{%
0.1	0.875796537626989\\
0.2	0.907836562760281\\
0.3	0.524995513962743\\
0.4	0.377448864959978\\
0.5	0.319580267419611\\
0.6	0.19193140632457\\
0.7	0.130360403864933\\
0.8	0.398486820363855\\
0.9	0.352585778978701\\
};
\addplot [color=mycolor4, dashdotted, line width=1.0pt, mark size=3.5pt, mark=square, mark options={solid, mycolor4}]
  table[row sep=crcr]{%
0.1	0.868937859455021\\
0.2	0.845293881371296\\
0.3	0.374583252661745\\
0.4	0.311182186712043\\
0.5	0.256641082352487\\
0.6	0.167683707839921\\
0.7	0.12250825246199\\
0.8	0.284267323217555\\
0.9	0.207203720219069\\
};
\addlegendentry{DS}

\addplot[area legend, draw=none, fill=mycolor5, fill opacity=0.2, forget plot]
table[row sep=crcr] {%
x	y\\
0.1	0.739250211719198\\
0.2	0.70011726193352\\
0.3	0.642255114803453\\
0.4	0.601831679058978\\
0.5	0.54702796309063\\
0.6	0.508783143042023\\
0.7	0.456432240769181\\
0.8	0.423805500448804\\
0.9	0.360159069603373\\
0.9	0.3939662299488\\
0.8	0.447665979472838\\
0.7	0.48421096039201\\
0.6	0.526850794242786\\
0.5	0.570515785922164\\
0.4	0.616252185560673\\
0.3	0.661711014008925\\
0.2	0.710930014504447\\
0.1	0.762761522142901\\
}--cycle;
\addplot [color=white!55!mycolor5, forget plot]
  table[row sep=crcr]{%
0.1	0.739250211719198\\
0.2	0.70011726193352\\
0.3	0.642255114803453\\
0.4	0.601831679058978\\
0.5	0.54702796309063\\
0.6	0.508783143042023\\
0.7	0.456432240769181\\
0.8	0.423805500448804\\
0.9	0.360159069603373\\
};
\addplot [color=white!55!mycolor5, forget plot]
  table[row sep=crcr]{%
0.1	0.762761522142901\\
0.2	0.710930014504447\\
0.3	0.661711014008925\\
0.4	0.616252185560673\\
0.5	0.570515785922164\\
0.6	0.526850794242786\\
0.7	0.48421096039201\\
0.8	0.447665979472838\\
0.9	0.3939662299488\\
};
\addplot [color=mycolor5, dashdotted, line width=1.0pt, mark size=5.0pt, mark=+, mark options={solid, mycolor5}]
  table[row sep=crcr]{%
0.1	0.75100586693105\\
0.2	0.705523638218983\\
0.3	0.651983064406189\\
0.4	0.609041932309826\\
0.5	0.558771874506397\\
0.6	0.517816968642404\\
0.7	0.470321600580595\\
0.8	0.435735739960821\\
0.9	0.377062649776086\\
};
\addlegendentry{MV}

\addplot[area legend, draw=none, fill=mycolor6, fill opacity=0.2, forget plot]
table[row sep=crcr] {%
x	y\\
0.1	0.749373490023618\\
0.2	0.658562470484064\\
0.3	0.5183208401634\\
0.4	0.433722585491927\\
0.5	0.359890163125411\\
0.6	0.282090016184488\\
0.7	0.222224095118877\\
0.8	0.14563889188135\\
0.9	0.0786376262404846\\
0.9	0.087097315784373\\
0.8	0.159287383597732\\
0.7	0.228798216195034\\
0.6	0.305595517527646\\
0.5	0.36661052718121\\
0.4	0.462239253029355\\
0.3	0.549201537420877\\
0.2	0.682246621181505\\
0.1	0.770178240647837\\
}--cycle;
\addplot [color=white!55!mycolor6, forget plot]
  table[row sep=crcr]{%
0.1	0.749373490023618\\
0.2	0.658562470484064\\
0.3	0.5183208401634\\
0.4	0.433722585491927\\
0.5	0.359890163125411\\
0.6	0.282090016184488\\
0.7	0.222224095118877\\
0.8	0.14563889188135\\
0.9	0.0786376262404846\\
};
\addplot [color=white!55!mycolor6, forget plot]
  table[row sep=crcr]{%
0.1	0.770178240647837\\
0.2	0.682246621181505\\
0.3	0.549201537420877\\
0.4	0.462239253029355\\
0.5	0.36661052718121\\
0.6	0.305595517527646\\
0.7	0.228798216195034\\
0.8	0.159287383597732\\
0.9	0.087097315784373\\
};
\addplot [color=mycolor6, dashdotted, line width=1.0pt, mark size=3.3pt, mark=triangle, mark options={solid, mycolor6}]
  table[row sep=crcr]{%
0.1	0.759775865335727\\
0.2	0.670404545832785\\
0.3	0.533761188792139\\
0.4	0.447980919260641\\
0.5	0.363250345153311\\
0.6	0.293842766856067\\
0.7	0.225511155656955\\
0.8	0.152463137739541\\
0.9	0.0828674710124288\\
};
\addlegendentry{PGD}

\end{axis}
\end{tikzpicture}%

%% file: fig/tikz/synthetic_data/p_adv_synth_data_detect_shaded_DS.tikz
\definecolor{mycolor1}{rgb}{0.00000,0.44700,0.74100}%
\definecolor{mycolor2}{rgb}{0.49400,0.18400,0.55600}%
\definecolor{mycolor3}{rgb}{0.63500,0.07800,0.18400}%
\definecolor{mycolor4}{rgb}{0.92900,0.69400,0.12500}%
\definecolor{mycolor5}{rgb}{0.30100,0.74500,0.93300}%
\begin{tikzpicture}

\begin{axis}[%
width=4.521in,
height=3.286in,
at={(0.758in,0.76in)},
scale only axis,
xmin=0.1,
xmax=0.95,
xlabel style={font=\color{white!15!black}},
xlabel={Percentage of corrupted data},
ymin=0,
ymax=1,
axis background/.style={fill=white},
axis x line*=bottom,
axis y line*=left,
legend style={at={(0.97,0.03)}, anchor=south east, legend cell align=left, align=left, draw=white!15!black}
]

\addplot[area legend, draw=none, fill=mycolor1, fill opacity=0.2, forget plot]
table[row sep=crcr] {%
x	y\\
0.1	0\\
0.2	0\\
0.3	0\\
0.4	-0.247213595499958\\
0.5	1\\
0.6	1\\
0.7	1\\
0.8	1\\
0.9	1\\
0.9	1\\
0.8	1\\
0.7	1\\
0.6	1\\
0.5	1\\
0.4	0.647213595499958\\
0.3	0\\
0.2	0\\
0.1	0\\
}--cycle;
\addplot [color=white!55!mycolor1, forget plot]
  table[row sep=crcr]{%
0.1	0\\
0.2	0\\
0.3	0\\
0.4	-0.247213595499958\\
0.5	1\\
0.6	1\\
0.7	1\\
0.8	1\\
0.9	1\\
};
\addplot [color=white!55!mycolor1, forget plot]
  table[row sep=crcr]{%
0.1	0\\
0.2	0\\
0.3	0\\
0.4	0.647213595499958\\
0.5	1\\
0.6	1\\
0.7	1\\
0.8	1\\
0.9	1\\
};
\addplot [color=mycolor1, line width=1.0pt, mark size=5.0pt, mark=star, mark options={solid, mycolor1}]
  table[row sep=crcr]{%
0.1	0\\
0.2	0\\
0.3	0\\
0.4	0.2\\
0.5	1\\
0.6	1\\
0.7	1\\
0.8	1\\
0.9	1\\
};
\addlegendentry{Sensitivity - Alg. 1 - H}

\addplot[area legend, draw=none, fill=mycolor2, fill opacity=0.2, forget plot]
table[row sep=crcr] {%
x	y\\
0.1	1\\
0.2	1\\
0.3	1\\
0.4	1\\
0.5	1\\
0.6	1\\
0.7	1\\
0.8	1\\
0.9	1\\
0.9	1\\
0.8	1\\
0.7	1\\
0.6	1\\
0.5	1\\
0.4	1\\
0.3	1\\
0.2	1\\
0.1	1\\
}--cycle;
\addplot [color=white!55!mycolor2, forget plot]
  table[row sep=crcr]{%
0.1	1\\
0.2	1\\
0.3	1\\
0.4	1\\
0.5	1\\
0.6	1\\
0.7	1\\
0.8	1\\
0.9	1\\
};
\addplot [color=white!55!mycolor2, forget plot]
  table[row sep=crcr]{%
0.1	1\\
0.2	1\\
0.3	1\\
0.4	1\\
0.5	1\\
0.6	1\\
0.7	1\\
0.8	1\\
0.9	1\\
};
\addplot [color=mycolor2, line width=1.0pt, mark size=3.3pt, mark=triangle, mark options={solid, rotate=270, mycolor2}]
  table[row sep=crcr]{%
0.1	1\\
0.2	1\\
0.3	1\\
0.4	1\\
0.5	1\\
0.6	1\\
0.7	1\\
0.8	1\\
0.9	1\\
};
\addlegendentry{Specificity - Alg. 1 - H}

\addplot[area legend, draw=none, fill=mycolor3, fill opacity=0.2, forget plot]
table[row sep=crcr] {%
x	y\\
0.1	0\\
0.2	0\\
0.3	0\\
0.4	-0.247213595499958\\
0.5	1\\
0.6	1\\
0.7	1\\
0.8	1\\
0.9	1\\
0.9	1\\
0.8	1\\
0.7	1\\
0.6	1\\
0.5	1\\
0.4	0.647213595499958\\
0.3	0\\
0.2	0\\
0.1	0\\
}--cycle;
\addplot [color=white!55!mycolor3, forget plot]
  table[row sep=crcr]{%
0.1	0\\
0.2	0\\
0.3	0\\
0.4	-0.247213595499958\\
0.5	1\\
0.6	1\\
0.7	1\\
0.8	1\\
0.9	1\\
};
\addplot [color=white!55!mycolor3, forget plot]
  table[row sep=crcr]{%
0.1	0\\
0.2	0\\
0.3	0\\
0.4	0.647213595499958\\
0.5	1\\
0.6	1\\
0.7	1\\
0.8	1\\
0.9	1\\
};
\addplot [color=mycolor3, dashed, line width=1.0pt, mark size=5.0pt, mark=star, mark options={solid, mycolor3}]
  table[row sep=crcr]{%
0.1	0\\
0.2	0\\
0.3	0\\
0.4	0.2\\
0.5	1\\
0.6	1\\
0.7	1\\
0.8	1\\
0.9	1\\
};
\addlegendentry{Sensitivity - Alg. 1 - TA}

\addplot[area legend, draw=none, fill=mycolor4, fill opacity=0.2, forget plot]
table[row sep=crcr] {%
x	y\\
0.1	1\\
0.2	1\\
0.3	1\\
0.4	1\\
0.5	1\\
0.6	1\\
0.7	1\\
0.8	1\\
0.9	1\\
0.9	1\\
0.8	1\\
0.7	1\\
0.6	1\\
0.5	1\\
0.4	1\\
0.3	1\\
0.2	1\\
0.1	1\\
}--cycle;
\addplot [color=white!55!mycolor4, forget plot]
  table[row sep=crcr]{%
0.1	1\\
0.2	1\\
0.3	1\\
0.4	1\\
0.5	1\\
0.6	1\\
0.7	1\\
0.8	1\\
0.9	1\\
};
\addplot [color=white!55!mycolor4, forget plot]
  table[row sep=crcr]{%
0.1	1\\
0.2	1\\
0.3	1\\
0.4	1\\
0.5	1\\
0.6	1\\
0.7	1\\
0.8	1\\
0.9	1\\
};
\addplot [color=mycolor4, dashed, line width=1.0pt, mark size=3.3pt, mark=triangle, mark options={solid, rotate=90, mycolor4}]
  table[row sep=crcr]{%
0.1	1\\
0.2	1\\
0.3	1\\
0.4	1\\
0.5	1\\
0.6	1\\
0.7	1\\
0.8	1\\
0.9	1\\
};
\addlegendentry{Specificity - Alg. 1 - TA}

\addplot[area legend, draw=none, fill=mycolor5, fill opacity=0.2, forget plot]
table[row sep=crcr] {%
x	y\\
0.1	0.7\\
0.2	0.7\\
0.3	0.7\\
0.4	0.625835921350013\\
0.5	1\\
0.6	1\\
0.7	1\\
0.8	1\\
0.9	1\\
0.9	1\\
0.8	1\\
0.7	1\\
0.6	1\\
0.5	1\\
0.4	0.894164078649987\\
0.3	0.7\\
0.2	0.7\\
0.1	0.7\\
}--cycle;
\addplot [color=white!55!mycolor5, forget plot]
  table[row sep=crcr]{%
0.1	0.7\\
0.2	0.7\\
0.3	0.7\\
0.4	0.625835921350013\\
0.5	1\\
0.6	1\\
0.7	1\\
0.8	1\\
0.9	1\\
};
\addplot [color=white!55!mycolor5, forget plot]
  table[row sep=crcr]{%
0.1	0.7\\
0.2	0.7\\
0.3	0.7\\
0.4	0.894164078649987\\
0.5	1\\
0.6	1\\
0.7	1\\
0.8	1\\
0.9	1\\
};
\addplot [color=mycolor5, dashdotted, line width=1.0pt, mark size=3.3pt, mark=triangle, mark options={solid, rotate=180, mycolor5}]
  table[row sep=crcr]{%
0.1	0.7\\
0.2	0.7\\
0.3	0.7\\
0.4	0.76\\
0.5	1\\
0.6	1\\
0.7	1\\
0.8	1\\
0.9	1\\
};
\addlegendentry{Clustering Accuracy}

\end{axis}
\end{tikzpicture}%

%% file: fig/tikz/legend_acc.tikz
% \definecolor{mycolor1}{rgb}{0.00000,0.44700,0.74100}%
% \definecolor{mycolor2}{rgb}{0.85000,0.32500,0.09800}%
% \definecolor{mycolor3}{rgb}{0.92900,0.69400,0.12500}%
% \definecolor{mycolor4}{rgb}{0.49400,0.18400,0.55600}%
% \definecolor{mycolor5}{rgb}{0.46600,0.67400,0.18800}%
% \definecolor{mycolor6}{rgb}{0.30100,0.74500,0.93300}%
%

\definecolor{mycolor1}{rgb}{0.00000,0.44700,0.74100}%
\definecolor{mycolor2}{rgb}{0.49400,0.18400,0.55600}%
\definecolor{mycolor3}{rgb}{0.63500,0.07800,0.18400}%
\definecolor{mycolor4}{rgb}{0.92900,0.69400,0.12500}%
\definecolor{mycolor5}{rgb}{0.30100,0.74500,0.93300}%
\definecolor{mycolor6}{rgb}{0.85000,0.32500,0.09800}%
\begin{tikzpicture}

\begin{axis}[%
width=6.313in,
height=0.002in,
at={(1.059in,0.139in)},
hide axis,
scale only axis,
xmin=0,
xmax=10,
ymin=-10.9802415033966,
ymax=6.72700694801328,
axis background/.style={fill=white},
axis x line*=bottom,
axis y line*=left,
legend style={at={(0,1)}, anchor=south west, legend columns=6, legend cell align=center, align=center, draw=white!15!black}
]
\addlegendimage{color=mycolor1, line width=1.0pt, mark size=5.0pt, mark=o, mark options={solid, mycolor1}}
\addlegendentry{Alg. 1 - TA + DS}
\addlegendimage{color=mycolor2, line width=1.0pt, mark size=8.6pt, mark=diamond, mark options={solid, mycolor2}}
\addlegendentry{Alg. 1 - H + DS}
\addlegendimage{color=mycolor3, dashed, line width=1.0pt, mark size=5.0pt, mark=x, mark options={solid, mycolor3}}
\addlegendentry{MMSR}
\addlegendimage{color=mycolor4, dashdotted, line width=1.0pt, mark size=3.5pt, mark=square, mark options={solid, mycolor4}}
\addlegendentry{DS}
\addlegendimage{color=mycolor5, dashdotted, line width=1.0pt, mark size=5.0pt, mark=+, mark options={solid, mycolor5}}
\addlegendentry{MV}
\addlegendimage{color=mycolor6, dashdotted, line width=1.0pt, mark size=3.3pt, mark=triangle, mark options={solid, mycolor6}}
\addlegendentry{PGD}

\end{axis}
\end{tikzpicture}%

%% file: fig/tikz/real_data/per_adv_groundtruth_bluebird_shaded_acc.tikz
\definecolor{mycolor1}{rgb}{0.00000,0.44700,0.74100}%
\definecolor{mycolor2}{rgb}{0.49400,0.18400,0.55600}%
\definecolor{mycolor3}{rgb}{0.63500,0.07800,0.18400}%
\definecolor{mycolor4}{rgb}{0.92900,0.69400,0.12500}%
\definecolor{mycolor5}{rgb}{0.30100,0.74500,0.93300}%
\definecolor{mycolor6}{rgb}{0.85000,0.32500,0.09800}%
\begin{tikzpicture}

\begin{axis}[%
width=4.464in,
height=3.286in,
at={(0.815in,0.76in)},
scale only axis,
xmin=0,
xmax=0.95,
xlabel style={font=\color{white!15!black}},
xlabel={Percentage of adversaries},
ymin=0,
ymax=1,
ylabel style={font=\color{white!15!black}},
ylabel={Accuracy},
axis background/.style={fill=white},
axis x line*=bottom,
axis y line*=left,
]

\addplot[area legend, draw=none, fill=mycolor1, fill opacity=0.2, forget plot]
table[row sep=crcr] {%
x	y\\
0	0.537396483863875\\
0.1	0.874628005323156\\
0.2	0.872306571944253\\
0.3	0.878485755498452\\
0.4	0.877098888397517\\
0.5	0.868289399338967\\
0.6	0.867634767453608\\
0.7	0.701090297233303\\
0.8	0.654010847424058\\
0.9	0.61492121247556\\
0.9	0.888782491228143\\
0.8	0.971915078501868\\
0.7	0.987798591655586\\
0.6	0.887920788101947\\
0.5	0.890969859920292\\
0.4	0.908086296787668\\
0.3	0.91410683709414\\
0.2	0.912878613240932\\
0.1	0.895742365047214\\
0	0.932973886506495\\
}--cycle;
\addplot [color=white!55!mycolor1, forget plot]
  table[row sep=crcr]{%
0	0.537396483863875\\
0.1	0.874628005323156\\
0.2	0.872306571944253\\
0.3	0.878485755498452\\
0.4	0.877098888397517\\
0.5	0.868289399338967\\
0.6	0.867634767453608\\
0.7	0.701090297233303\\
0.8	0.654010847424058\\
0.9	0.61492121247556\\
};
\addplot [color=white!55!mycolor1, forget plot]
  table[row sep=crcr]{%
0	0.932973886506495\\
0.1	0.895742365047214\\
0.2	0.912878613240932\\
0.3	0.91410683709414\\
0.4	0.908086296787668\\
0.5	0.890969859920292\\
0.6	0.887920788101947\\
0.7	0.987798591655586\\
0.8	0.971915078501868\\
0.9	0.888782491228143\\
};
\addplot [color=mycolor1, line width=1.0pt, mark size=5.0pt, mark=o, mark options={solid, mycolor1}]
  table[row sep=crcr]{%
0	0.735185185185185\\
0.1	0.885185185185185\\
0.2	0.892592592592593\\
0.3	0.896296296296296\\
0.4	0.892592592592593\\
0.5	0.87962962962963\\
0.6	0.877777777777778\\
0.7	0.844444444444444\\
0.8	0.812962962962963\\
0.9	0.751851851851852\\
};
%\addlegendentry{Alg. 1 - TA + DS}

\addplot[area legend, draw=none, fill=mycolor2, fill opacity=0.2, forget plot]
table[row sep=crcr] {%
x	y\\
0	0.879629629629629\\
0.1	0.874628005323156\\
0.2	0.872306571944253\\
0.3	0.878485755498452\\
0.4	0.877098888397517\\
0.5	0.117574598183613\\
0.6	0.119623955824685\\
0.7	0.11886673725039\\
0.8	0.0186530610513871\\
0.9	0.118071669343615\\
0.9	0.663409812137866\\
0.8	0.566532124133798\\
0.7	0.670022151638499\\
0.6	0.687783451582722\\
0.5	0.641684661075646\\
0.4	0.908086296787668\\
0.3	0.91410683709414\\
0.2	0.912878613240932\\
0.1	0.895742365047214\\
0	0.87962962962963\\
}--cycle;
\addplot [color=white!55!mycolor2, forget plot]
  table[row sep=crcr]{%
0	0.879629629629629\\
0.1	0.874628005323156\\
0.2	0.872306571944253\\
0.3	0.878485755498452\\
0.4	0.877098888397517\\
0.5	0.117574598183613\\
0.6	0.119623955824685\\
0.7	0.11886673725039\\
0.8	0.0186530610513871\\
0.9	0.118071669343615\\
};
\addplot [color=white!55!mycolor2, forget plot]
  table[row sep=crcr]{%
0	0.87962962962963\\
0.1	0.895742365047214\\
0.2	0.912878613240932\\
0.3	0.91410683709414\\
0.4	0.908086296787668\\
0.5	0.641684661075646\\
0.6	0.687783451582722\\
0.7	0.670022151638499\\
0.8	0.566532124133798\\
0.9	0.663409812137866\\
};
\addplot [color=mycolor2, line width=1.0pt, mark size=8.6pt, mark=diamond, mark options={solid, mycolor2}]
  table[row sep=crcr]{%
0	0.87962962962963\\
0.1	0.885185185185185\\
0.2	0.892592592592593\\
0.3	0.896296296296296\\
0.4	0.892592592592593\\
0.5	0.37962962962963\\
0.6	0.403703703703704\\
0.7	0.394444444444444\\
0.8	0.292592592592593\\
0.9	0.390740740740741\\
};
%\addlegendentry{Alg. 1 - H + DS}

\addplot[area legend, draw=none, fill=mycolor3, fill opacity=0.2, forget plot]
table[row sep=crcr] {%
x	y\\
0	0.712962962962963\\
0.1	0.751958927582324\\
0.2	0.591516514279343\\
0.3	0.555369465559162\\
0.4	0.102836400622661\\
0.5	0.117574598183613\\
0.6	0.119623955824685\\
0.7	0.11886673725039\\
0.8	0.0186530610513871\\
0.9	0.118071669343615\\
0.9	0.663409812137866\\
0.8	0.566532124133798\\
0.7	0.670022151638499\\
0.6	0.687783451582722\\
0.5	0.641684661075646\\
0.4	0.819385821599562\\
0.3	0.89648238629269\\
0.2	0.87144644868362\\
0.1	0.799892924269527\\
0	0.712962962962963\\
}--cycle;
\addplot [color=white!55!mycolor3, forget plot]
  table[row sep=crcr]{%
0	0.712962962962963\\
0.1	0.751958927582324\\
0.2	0.591516514279343\\
0.3	0.555369465559162\\
0.4	0.102836400622661\\
0.5	0.117574598183613\\
0.6	0.119623955824685\\
0.7	0.11886673725039\\
0.8	0.0186530610513871\\
0.9	0.118071669343615\\
};
\addplot [color=white!55!mycolor3, forget plot]
  table[row sep=crcr]{%
0	0.712962962962963\\
0.1	0.799892924269527\\
0.2	0.87144644868362\\
0.3	0.89648238629269\\
0.4	0.819385821599562\\
0.5	0.641684661075646\\
0.6	0.687783451582722\\
0.7	0.670022151638499\\
0.8	0.566532124133798\\
0.9	0.663409812137866\\
};
\addplot [color=mycolor3, dashed, line width=1.0pt, mark size=5.0pt, mark=x, mark options={solid, mycolor3}]
  table[row sep=crcr]{%
0	0.712962962962963\\
0.1	0.775925925925926\\
0.2	0.731481481481482\\
0.3	0.725925925925926\\
0.4	0.461111111111111\\
0.5	0.37962962962963\\
0.6	0.403703703703704\\
0.7	0.394444444444444\\
0.8	0.292592592592593\\
0.9	0.390740740740741\\
};
%\addlegendentry{MMSR}

\addplot[area legend, draw=none, fill=mycolor4, fill opacity=0.2, forget plot]
table[row sep=crcr] {%
x	y\\
0	0.879629629629629\\
0.1	0.930183560878712\\
0.2	0.881948368141199\\
0.3	0.109301349217117\\
0.4	0.016508329134002\\
0.5	0.117574598183613\\
0.6	0.119623955824685\\
0.7	0.11886673725039\\
0.8	0.0186530610513871\\
0.9	0.118071669343615\\
0.9	0.663409812137866\\
0.8	0.566532124133798\\
0.7	0.670022151638499\\
0.6	0.687783451582722\\
0.5	0.641684661075646\\
0.4	0.579787967162294\\
0.3	0.842550502634735\\
0.2	0.93657015037732\\
0.1	0.95129792060277\\
0	0.87962962962963\\
}--cycle;
\addplot [color=white!55!mycolor4, forget plot]
  table[row sep=crcr]{%
0	0.879629629629629\\
0.1	0.930183560878712\\
0.2	0.881948368141199\\
0.3	0.109301349217117\\
0.4	0.016508329134002\\
0.5	0.117574598183613\\
0.6	0.119623955824685\\
0.7	0.11886673725039\\
0.8	0.0186530610513871\\
0.9	0.118071669343615\\
};
\addplot [color=white!55!mycolor4, forget plot]
  table[row sep=crcr]{%
0	0.87962962962963\\
0.1	0.95129792060277\\
0.2	0.93657015037732\\
0.3	0.842550502634735\\
0.4	0.579787967162294\\
0.5	0.641684661075646\\
0.6	0.687783451582722\\
0.7	0.670022151638499\\
0.8	0.566532124133798\\
0.9	0.663409812137866\\
};
\addplot [color=mycolor4, dashdotted, line width=1.0pt, mark size=3.5pt, mark=square, mark options={solid, mycolor4}]
  table[row sep=crcr]{%
0	0.87962962962963\\
0.1	0.940740740740741\\
0.2	0.909259259259259\\
0.3	0.475925925925926\\
0.4	0.298148148148148\\
0.5	0.37962962962963\\
0.6	0.403703703703704\\
0.7	0.394444444444444\\
0.8	0.292592592592593\\
0.9	0.390740740740741\\
};
%\addlegendentry{DS}

\addplot[area legend, draw=none, fill=mycolor5, fill opacity=0.2, forget plot]
table[row sep=crcr] {%
x	y\\
0	0.759259259259259\\
0.1	0.705098239322297\\
0.2	0.653637622281551\\
0.3	0.373413300161744\\
0.4	0.171626655242975\\
0.5	0.117574598183613\\
0.6	0.119623955824685\\
0.7	0.11886673725039\\
0.8	0.0186530610513871\\
0.9	0.118071669343615\\
0.9	0.663409812137866\\
0.8	0.566532124133798\\
0.7	0.670022151638499\\
0.6	0.687783451582722\\
0.5	0.641684661075646\\
0.4	0.583928900312581\\
0.3	0.693253366504923\\
0.2	0.731547562903634\\
0.1	0.743049908825851\\
0	0.759259259259259\\
}--cycle;
\addplot [color=white!55!mycolor5, forget plot]
  table[row sep=crcr]{%
0	0.759259259259259\\
0.1	0.705098239322297\\
0.2	0.653637622281551\\
0.3	0.373413300161744\\
0.4	0.171626655242975\\
0.5	0.117574598183613\\
0.6	0.119623955824685\\
0.7	0.11886673725039\\
0.8	0.0186530610513871\\
0.9	0.118071669343615\\
};
\addplot [color=white!55!mycolor5, forget plot]
  table[row sep=crcr]{%
0	0.759259259259259\\
0.1	0.743049908825851\\
0.2	0.731547562903634\\
0.3	0.693253366504923\\
0.4	0.583928900312581\\
0.5	0.641684661075646\\
0.6	0.687783451582722\\
0.7	0.670022151638499\\
0.8	0.566532124133798\\
0.9	0.663409812137866\\
};
\addplot [color=mycolor5, dashdotted, line width=1.0pt, mark size=5.0pt, mark=+, mark options={solid, mycolor5}]
  table[row sep=crcr]{%
0	0.759259259259259\\
0.1	0.724074074074074\\
0.2	0.692592592592593\\
0.3	0.533333333333333\\
0.4	0.377777777777778\\
0.5	0.37962962962963\\
0.6	0.403703703703704\\
0.7	0.394444444444444\\
0.8	0.292592592592593\\
0.9	0.390740740740741\\
};
%\addlegendentry{MV}

\addplot[area legend, draw=none, fill=mycolor6, fill opacity=0.2, forget plot]
table[row sep=crcr] {%
x	y\\
0	0.759259259259259\\
0.1	0.710957498032792\\
0.2	0.512686013555706\\
0.3	0.118984936450346\\
0.4	0.016508329134002\\
0.5	0.117574598183613\\
0.6	0.119623955824685\\
0.7	0.11886673725039\\
0.8	0.0186530610513871\\
0.9	0.118071669343615\\
0.9	0.663409812137866\\
0.8	0.566532124133798\\
0.7	0.670022151638499\\
0.6	0.687783451582722\\
0.5	0.641684661075646\\
0.4	0.579787967162294\\
0.3	0.677311359845951\\
0.2	0.605832504962812\\
0.1	0.744598057522764\\
0	0.759259259259259\\
}--cycle;
\addplot [color=white!55!mycolor6, forget plot]
  table[row sep=crcr]{%
0	0.759259259259259\\
0.1	0.710957498032792\\
0.2	0.512686013555706\\
0.3	0.118984936450346\\
0.4	0.016508329134002\\
0.5	0.117574598183613\\
0.6	0.119623955824685\\
0.7	0.11886673725039\\
0.8	0.0186530610513871\\
0.9	0.118071669343615\\
};
\addplot [color=white!55!mycolor6, forget plot]
  table[row sep=crcr]{%
0	0.759259259259259\\
0.1	0.744598057522764\\
0.2	0.605832504962812\\
0.3	0.677311359845951\\
0.4	0.579787967162294\\
0.5	0.641684661075646\\
0.6	0.687783451582722\\
0.7	0.670022151638499\\
0.8	0.566532124133798\\
0.9	0.663409812137866\\
};
\addplot [color=mycolor6, dashdotted, line width=1.0pt, mark size=3.3pt, mark=triangle, mark options={solid, mycolor6}]
  table[row sep=crcr]{%
0	0.759259259259259\\
0.1	0.727777777777778\\
0.2	0.559259259259259\\
0.3	0.398148148148148\\
0.4	0.298148148148148\\
0.5	0.37962962962963\\
0.6	0.403703703703704\\
0.7	0.394444444444444\\
0.8	0.292592592592593\\
0.9	0.390740740740741\\
};
%\addlegendentry{PGD}

\end{axis}

\begin{axis}[%
width=5.833in,
height=4.375in,
at={(0in,0in)},
scale only axis,
xmin=0,
xmax=1,
ymin=0,
ymax=1,
axis line style={draw=none},
ticks=none,
axis x line*=bottom,
axis y line*=left,
legend style={legend cell align=left, align=left, draw=white!15!black}
]
\end{axis}
\end{tikzpicture}%

%% file: fig/tikz/real_data/per_adv_groundtruth_rte_shaded_acc.tikz
\definecolor{mycolor1}{rgb}{0.00000,0.44700,0.74100}%
\definecolor{mycolor2}{rgb}{0.49400,0.18400,0.55600}%
\definecolor{mycolor3}{rgb}{0.63500,0.07800,0.18400}%
\definecolor{mycolor4}{rgb}{0.92900,0.69400,0.12500}%
\definecolor{mycolor5}{rgb}{0.30100,0.74500,0.93300}%
\definecolor{mycolor6}{rgb}{0.85000,0.32500,0.09800}%
\begin{tikzpicture}

\begin{axis}[%
width=4.464in,
height=3.286in,
at={(0.815in,0.76in)},
scale only axis,
xmin=0,
xmax=0.95,
xlabel style={font=\color{white!15!black}},
xlabel={Percentage of adversaries},
ymin=0,
ymax=1,
ylabel style={font=\color{white!15!black}},
ylabel={Accuracy},
axis background/.style={fill=white},
axis x line*=bottom,
axis y line*=left,
]

\addplot[area legend, draw=none, fill=mycolor1, fill opacity=0.2, forget plot]
table[row sep=crcr] {%
x	y\\
0	0.916315346803119\\
0.1	0.89884644002085\\
0.2	0.567665111335107\\
0.3	0.887649244290377\\
0.4	0.678286555316343\\
0.5	0.870151177304516\\
0.6	0.908625449646106\\
0.7	0.812578022446259\\
0.8	0.686884293808739\\
0.9	0.660255633611352\\
0.9	0.954244366388648\\
0.8	0.90711570619126\\
0.7	0.905421977553741\\
0.6	0.943374550353894\\
0.5	0.942348822695484\\
0.4	1.01421344468366\\
0.3	0.948850755709623\\
0.2	0.991834888664893\\
0.1	0.93115355997915\\
0	0.917684653196882\\
}--cycle;
\addplot [color=white!55!mycolor1, forget plot]
  table[row sep=crcr]{%
0	0.916315346803119\\
0.1	0.89884644002085\\
0.2	0.567665111335107\\
0.3	0.887649244290377\\
0.4	0.678286555316343\\
0.5	0.870151177304516\\
0.6	0.908625449646106\\
0.7	0.812578022446259\\
0.8	0.686884293808739\\
0.9	0.660255633611352\\
};
\addplot [color=white!55!mycolor1, forget plot]
  table[row sep=crcr]{%
0	0.917684653196882\\
0.1	0.93115355997915\\
0.2	0.991834888664893\\
0.3	0.948850755709623\\
0.4	1.01421344468366\\
0.5	0.942348822695484\\
0.6	0.943374550353894\\
0.7	0.905421977553741\\
0.8	0.90711570619126\\
0.9	0.954244366388648\\
};
\addplot [color=mycolor1, line width=1.0pt, mark size=5.0pt, mark=o, mark options={solid, mycolor1}]
  table[row sep=crcr]{%
0	0.917\\
0.1	0.915\\
0.2	0.77975\\
0.3	0.91825\\
0.4	0.84625\\
0.5	0.90625\\
0.6	0.926\\
0.7	0.859\\
0.8	0.797\\
0.9	0.80725\\
};
%\addlegendentry{Alg. 1 - TA + DS}

\addplot[area legend, draw=none, fill=mycolor2, fill opacity=0.2, forget plot]
table[row sep=crcr] {%
x	y\\
0	0.9175\\
0.1	0.89884644002085\\
0.2	0.91522078101061\\
0.3	0.887649244290377\\
0.4	0.905585759374297\\
0.5	0.0329878208170923\\
0.6	-0.0115690669918563\\
0.7	0.122489007780561\\
0.8	0.122123438301762\\
0.9	0.0317813463899217\\
0.9	0.528218653610078\\
0.8	0.612376561698238\\
0.7	0.620010992219439\\
0.6	0.387569066991856\\
0.5	0.520512179182908\\
0.4	0.929914240625703\\
0.3	0.948850755709623\\
0.2	0.93877921898939\\
0.1	0.93115355997915\\
0	0.9175\\
}--cycle;
\addplot [color=white!55!mycolor2, forget plot]
  table[row sep=crcr]{%
0	0.9175\\
0.1	0.89884644002085\\
0.2	0.91522078101061\\
0.3	0.887649244290377\\
0.4	0.905585759374297\\
0.5	0.0329878208170923\\
0.6	-0.0115690669918563\\
0.7	0.122489007780561\\
0.8	0.122123438301762\\
0.9	0.0317813463899217\\
};
\addplot [color=white!55!mycolor2, forget plot]
  table[row sep=crcr]{%
0	0.9175\\
0.1	0.93115355997915\\
0.2	0.93877921898939\\
0.3	0.948850755709623\\
0.4	0.929914240625703\\
0.5	0.520512179182908\\
0.6	0.387569066991856\\
0.7	0.620010992219439\\
0.8	0.612376561698238\\
0.9	0.528218653610078\\
};
\addplot [color=mycolor2, line width=1.0pt, mark size=8.6pt, mark=diamond, mark options={solid, mycolor2}]
  table[row sep=crcr]{%
0	0.9175\\
0.1	0.915\\
0.2	0.927\\
0.3	0.91825\\
0.4	0.91775\\
0.5	0.27675\\
0.6	0.188\\
0.7	0.37125\\
0.8	0.36725\\
0.9	0.28\\
};
%\addlegendentry{Alg. 1 - H + DS}

\addplot[area legend, draw=none, fill=mycolor3, fill opacity=0.2, forget plot]
table[row sep=crcr] {%
x	y\\
0	0.8975\\
0.1	0.505457485434388\\
0.2	0.496582947365027\\
0.3	0.573099004748195\\
0.4	0.496671728954292\\
0.5	0.0329878208170923\\
0.6	-0.0115690669918563\\
0.7	0.122489007780561\\
0.8	0.122123438301762\\
0.9	0.0317813463899217\\
0.9	0.528218653610078\\
0.8	0.612376561698238\\
0.7	0.620010992219439\\
0.6	0.387569066991856\\
0.5	0.520512179182908\\
0.4	0.882828271045708\\
0.3	0.952400995251805\\
0.2	0.883417052634973\\
0.1	0.810542514565612\\
0	0.8975\\
}--cycle;
\addplot [color=white!55!mycolor3, forget plot]
  table[row sep=crcr]{%
0	0.8975\\
0.1	0.505457485434388\\
0.2	0.496582947365027\\
0.3	0.573099004748195\\
0.4	0.496671728954292\\
0.5	0.0329878208170923\\
0.6	-0.0115690669918563\\
0.7	0.122489007780561\\
0.8	0.122123438301762\\
0.9	0.0317813463899217\\
};
\addplot [color=white!55!mycolor3, forget plot]
  table[row sep=crcr]{%
0	0.8975\\
0.1	0.810542514565612\\
0.2	0.883417052634973\\
0.3	0.952400995251805\\
0.4	0.882828271045708\\
0.5	0.520512179182908\\
0.6	0.387569066991856\\
0.7	0.620010992219439\\
0.8	0.612376561698238\\
0.9	0.528218653610078\\
};
\addplot [color=mycolor3, dashed, line width=1.0pt, mark size=5.0pt, mark=x, mark options={solid, mycolor3}]
  table[row sep=crcr]{%
0	0.8975\\
0.1	0.658\\
0.2	0.69\\
0.3	0.76275\\
0.4	0.68975\\
0.5	0.27675\\
0.6	0.188\\
0.7	0.37125\\
0.8	0.36725\\
0.9	0.28\\
};
%\addlegendentry{MMSR}

\addplot[area legend, draw=none, fill=mycolor4, fill opacity=0.2, forget plot]
table[row sep=crcr] {%
x	y\\
0	0.92875\\
0.1	0.395746231828412\\
0.2	0.122229167548762\\
0.3	0.031350020008004\\
0.4	0.122270094632443\\
0.5	0.0329878208170923\\
0.6	-0.0115690669918563\\
0.7	0.122489007780561\\
0.8	0.122123438301762\\
0.9	0.0317813463899217\\
0.9	0.528218653610078\\
0.8	0.612376561698238\\
0.7	0.620010992219439\\
0.6	0.387569066991856\\
0.5	0.520512179182908\\
0.4	0.615229905367557\\
0.3	0.531149979991996\\
0.2	0.614270832451238\\
0.1	1.11725376817159\\
0	0.92875\\
}--cycle;
\addplot [color=white!55!mycolor4, forget plot]
  table[row sep=crcr]{%
0	0.92875\\
0.1	0.395746231828412\\
0.2	0.122229167548762\\
0.3	0.031350020008004\\
0.4	0.122270094632443\\
0.5	0.0329878208170923\\
0.6	-0.0115690669918563\\
0.7	0.122489007780561\\
0.8	0.122123438301762\\
0.9	0.0317813463899217\\
};
\addplot [color=white!55!mycolor4, forget plot]
  table[row sep=crcr]{%
0	0.92875\\
0.1	1.11725376817159\\
0.2	0.614270832451238\\
0.3	0.531149979991996\\
0.4	0.615229905367557\\
0.5	0.520512179182908\\
0.6	0.387569066991856\\
0.7	0.620010992219439\\
0.8	0.612376561698238\\
0.9	0.528218653610078\\
};
\addplot [color=mycolor4, dashdotted, line width=1.0pt, mark size=3.5pt, mark=square, mark options={solid, mycolor4}]
  table[row sep=crcr]{%
0	0.92875\\
0.1	0.7565\\
0.2	0.36825\\
0.3	0.28125\\
0.4	0.36875\\
0.5	0.27675\\
0.6	0.188\\
0.7	0.37125\\
0.8	0.36725\\
0.9	0.28\\
};
%\addlegendentry{DS}

\addplot[area legend, draw=none, fill=mycolor5, fill opacity=0.2, forget plot]
table[row sep=crcr] {%
x	y\\
0	0.91875\\
0.1	0.583843900064502\\
0.2	0.171515158076426\\
0.3	0.0312557443216728\\
0.4	0.122270094632443\\
0.5	0.0329878208170923\\
0.6	-0.0115690669918563\\
0.7	0.122489007780561\\
0.8	0.122123438301762\\
0.9	0.0317813463899217\\
0.9	0.528218653610078\\
0.8	0.612376561698238\\
0.7	0.620010992219439\\
0.6	0.387569066991856\\
0.5	0.520512179182908\\
0.4	0.615229905367557\\
0.3	0.531744255678327\\
0.2	0.639484841923574\\
0.1	0.850656099935498\\
0	0.91875\\
}--cycle;
\addplot [color=white!55!mycolor5, forget plot]
  table[row sep=crcr]{%
0	0.91875\\
0.1	0.583843900064502\\
0.2	0.171515158076426\\
0.3	0.0312557443216728\\
0.4	0.122270094632443\\
0.5	0.0329878208170923\\
0.6	-0.0115690669918563\\
0.7	0.122489007780561\\
0.8	0.122123438301762\\
0.9	0.0317813463899217\\
};
\addplot [color=white!55!mycolor5, forget plot]
  table[row sep=crcr]{%
0	0.91875\\
0.1	0.850656099935498\\
0.2	0.639484841923574\\
0.3	0.531744255678327\\
0.4	0.615229905367557\\
0.5	0.520512179182908\\
0.6	0.387569066991856\\
0.7	0.620010992219439\\
0.8	0.612376561698238\\
0.9	0.528218653610078\\
};
\addplot [color=mycolor5, dashdotted, line width=1.0pt, mark size=5.0pt, mark=+, mark options={solid, mycolor5}]
  table[row sep=crcr]{%
0	0.91875\\
0.1	0.71725\\
0.2	0.4055\\
0.3	0.2815\\
0.4	0.36875\\
0.5	0.27675\\
0.6	0.188\\
0.7	0.37125\\
0.8	0.36725\\
0.9	0.28\\
};
%\addlegendentry{MV}

\addplot[area legend, draw=none, fill=mycolor6, fill opacity=0.2, forget plot]
table[row sep=crcr] {%
x	y\\
0	0.84125\\
0.1	0.269487002288768\\
0.2	0.122249365483051\\
0.3	0.031350020008004\\
0.4	0.122270094632443\\
0.5	0.0329878208170923\\
0.6	-0.0115690669918563\\
0.7	0.122489007780561\\
0.8	0.122123438301762\\
0.9	0.0317813463899217\\
0.9	0.528218653610078\\
0.8	0.612376561698238\\
0.7	0.620010992219439\\
0.6	0.387569066991856\\
0.5	0.520512179182908\\
0.4	0.615229905367557\\
0.3	0.531149979991996\\
0.2	0.614750634516949\\
0.1	0.671012997711232\\
0	0.84125\\
}--cycle;
\addplot [color=white!55!mycolor6, forget plot]
  table[row sep=crcr]{%
0	0.84125\\
0.1	0.269487002288768\\
0.2	0.122249365483051\\
0.3	0.031350020008004\\
0.4	0.122270094632443\\
0.5	0.0329878208170923\\
0.6	-0.0115690669918563\\
0.7	0.122489007780561\\
0.8	0.122123438301762\\
0.9	0.0317813463899217\\
};
\addplot [color=white!55!mycolor6, forget plot]
  table[row sep=crcr]{%
0	0.84125\\
0.1	0.671012997711232\\
0.2	0.614750634516949\\
0.3	0.531149979991996\\
0.4	0.615229905367557\\
0.5	0.520512179182908\\
0.6	0.387569066991856\\
0.7	0.620010992219439\\
0.8	0.612376561698238\\
0.9	0.528218653610078\\
};
\addplot [color=mycolor6, dashdotted, line width=1.0pt, mark size=3.3pt, mark=triangle, mark options={solid, mycolor6}]
  table[row sep=crcr]{%
0	0.84125\\
0.1	0.47025\\
0.2	0.3685\\
0.3	0.28125\\
0.4	0.36875\\
0.5	0.27675\\
0.6	0.188\\
0.7	0.37125\\
0.8	0.36725\\
0.9	0.28\\
};
%\addlegendentry{PGD}

\end{axis}
\end{tikzpicture}%

%% file: fig/tikz/real_data/per_adv_groundtruth_sen_polarity_shaded_acc.tikz
\definecolor{mycolor1}{rgb}{0.00000,0.44700,0.74100}%
\definecolor{mycolor2}{rgb}{0.49400,0.18400,0.55600}%
\definecolor{mycolor3}{rgb}{0.63500,0.07800,0.18400}%
\definecolor{mycolor4}{rgb}{0.92900,0.69400,0.12500}%
\definecolor{mycolor5}{rgb}{0.30100,0.74500,0.93300}%
\definecolor{mycolor6}{rgb}{0.85000,0.32500,0.09800}%
\begin{tikzpicture}

\begin{axis}[%
width=4.464in,
height=3.286in,
at={(0.815in,0.76in)},
scale only axis,
xmin=0,
xmax=0.95,
xlabel style={font=\color{white!15!black}},
xlabel={Percentage of adversaries},
ymin=0,
ymax=1,
ylabel style={font=\color{white!15!black}},
ylabel={Accuracy},
axis background/.style={fill=white},
axis x line*=bottom,
axis y line*=left,
]

\addplot[area legend, draw=none, fill=mycolor1, fill opacity=0.2, forget plot]
table[row sep=crcr] {%
x	y\\
0	0.909916558943756\\
0.1	0.62841065236413\\
0.2	0.868284508261622\\
0.3	0.888628740391965\\
0.4	0.887889022989775\\
0.5	0.890261663947132\\
0.6	0.860434428788636\\
0.7	0.796124139772249\\
0.8	0.692562111862493\\
0.9	0.296395793774007\\
0.9	0.805024490282805\\
0.8	0.898076015763032\\
0.7	0.936942473550415\\
0.6	0.921762010499221\\
0.5	0.91617962431052\\
0.4	0.926713897594342\\
0.3	0.924213828121737\\
0.2	0.921033355311092\\
0.1	0.963587747315806\\
0	0.916928810130059\\
}--cycle;
\addplot [color=white!55!mycolor1, forget plot]
  table[row sep=crcr]{%
0	0.909916558943756\\
0.1	0.62841065236413\\
0.2	0.868284508261622\\
0.3	0.888628740391965\\
0.4	0.887889022989775\\
0.5	0.890261663947132\\
0.6	0.860434428788636\\
0.7	0.796124139772249\\
0.8	0.692562111862493\\
0.9	0.296395793774007\\
};
\addplot [color=white!55!mycolor1, forget plot]
  table[row sep=crcr]{%
0	0.916928810130059\\
0.1	0.963587747315806\\
0.2	0.921033355311092\\
0.3	0.924213828121737\\
0.4	0.926713897594342\\
0.5	0.91617962431052\\
0.6	0.921762010499221\\
0.7	0.936942473550415\\
0.8	0.898076015763032\\
0.9	0.805024490282805\\
};
\addplot [color=mycolor1, line width=1.0pt, mark size=5.0pt, mark=o, mark options={solid, mycolor1}]
  table[row sep=crcr]{%
0	0.913422684536907\\
0.1	0.795999199839968\\
0.2	0.894658931786357\\
0.3	0.906421284256851\\
0.4	0.907301460292058\\
0.5	0.903220644128826\\
0.6	0.891098219643929\\
0.7	0.866533306661332\\
0.8	0.795319063812763\\
0.9	0.550710142028406\\
};
%\addlegendentry{Alg. 1 - TA + DS}

\addplot[area legend, draw=none, fill=mycolor2, fill opacity=0.2, forget plot]
table[row sep=crcr] {%
x	y\\
0	0.915983196639328\\
0.1	0.678065454483497\\
0.2	0.868284508261622\\
0.3	0.888628740391965\\
0.4	0.887889022989775\\
0.5	0.0335514233478727\\
0.6	0.0331391847901884\\
0.7	0.0333741254086949\\
0.8	0.123314711888319\\
0.9	0.25898752373223\\
0.9	0.662556785129542\\
0.8	0.615793109675994\\
0.7	0.526097768970181\\
0.6	0.527612965639898\\
0.5	0.524960278992595\\
0.4	0.926713897594342\\
0.3	0.924213828121737\\
0.2	0.921033355311092\\
0.1	0.949860130633527\\
0	0.915983196639328\\
}--cycle;
\addplot [color=white!55!mycolor2, forget plot]
  table[row sep=crcr]{%
0	0.915983196639328\\
0.1	0.678065454483497\\
0.2	0.868284508261622\\
0.3	0.888628740391965\\
0.4	0.887889022989775\\
0.5	0.0335514233478727\\
0.6	0.0331391847901884\\
0.7	0.0333741254086949\\
0.8	0.123314711888319\\
0.9	0.25898752373223\\
};
\addplot [color=white!55!mycolor2, forget plot]
  table[row sep=crcr]{%
0	0.915983196639328\\
0.1	0.949860130633527\\
0.2	0.921033355311092\\
0.3	0.924213828121737\\
0.4	0.926713897594342\\
0.5	0.524960278992595\\
0.6	0.527612965639898\\
0.7	0.526097768970181\\
0.8	0.615793109675994\\
0.9	0.662556785129542\\
};
\addplot [color=mycolor2, line width=1.0pt, mark size=8.6pt, mark=diamond, mark options={solid, mycolor2}]
  table[row sep=crcr]{%
0	0.915983196639328\\
0.1	0.813962792558512\\
0.2	0.894658931786357\\
0.3	0.906421284256851\\
0.4	0.907301460292058\\
0.5	0.279255851170234\\
0.6	0.280376075215043\\
0.7	0.279735947189438\\
0.8	0.369553910782156\\
0.9	0.460772154430886\\
};
%\addlegendentry{Alg. 1 - H + DS}

\addplot[area legend, draw=none, fill=mycolor3, fill opacity=0.2, forget plot]
table[row sep=crcr] {%
x	y\\
0	0.901180236047209\\
0.1	0.7122817093967\\
0.2	0.547201742777314\\
0.3	0.497860013761961\\
0.4	0.567556612283997\\
0.5	0.0335514233478727\\
0.6	0.0331391847901884\\
0.7	0.0333741254086949\\
0.8	0.123314711888319\\
0.9	0.25898752373223\\
0.9	0.662556785129542\\
0.8	0.615793109675994\\
0.7	0.526097768970181\\
0.6	0.527612965639898\\
0.5	0.524960278992595\\
0.4	0.95202730449936\\
0.3	0.881855929426677\\
0.2	0.553498397250692\\
0.1	0.905641875320243\\
0	0.901180236047209\\
}--cycle;
\addplot [color=white!55!mycolor3, forget plot]
  table[row sep=crcr]{%
0	0.901180236047209\\
0.1	0.7122817093967\\
0.2	0.547201742777314\\
0.3	0.497860013761961\\
0.4	0.567556612283997\\
0.5	0.0335514233478727\\
0.6	0.0331391847901884\\
0.7	0.0333741254086949\\
0.8	0.123314711888319\\
0.9	0.25898752373223\\
};
\addplot [color=white!55!mycolor3, forget plot]
  table[row sep=crcr]{%
0	0.901180236047209\\
0.1	0.905641875320243\\
0.2	0.553498397250692\\
0.3	0.881855929426677\\
0.4	0.95202730449936\\
0.5	0.524960278992595\\
0.6	0.527612965639898\\
0.7	0.526097768970181\\
0.8	0.615793109675994\\
0.9	0.662556785129542\\
};
\addplot [color=mycolor3, dashed, line width=1.0pt, mark size=5.0pt, mark=x, mark options={solid, mycolor3}]
  table[row sep=crcr]{%
0	0.901180236047209\\
0.1	0.808961792358472\\
0.2	0.550350070014003\\
0.3	0.689857971594319\\
0.4	0.759791958391678\\
0.5	0.279255851170234\\
0.6	0.280376075215043\\
0.7	0.279735947189438\\
0.8	0.369553910782156\\
0.9	0.460772154430886\\
};
%\addlegendentry{MMSR}

\addplot[area legend, draw=none, fill=mycolor4, fill opacity=0.2, forget plot]
table[row sep=crcr] {%
x	y\\
0	0.914782956591318\\
0.1	0.123179534251087\\
0.2	0.546934485323787\\
0.3	0.123334749968722\\
0.4	0.0334341188742328\\
0.5	0.0335514233478727\\
0.6	0.0331391847901884\\
0.7	0.0333741254086949\\
0.8	0.123314711888319\\
0.9	0.25898752373223\\
0.9	0.662556785129542\\
0.8	0.615793109675994\\
0.7	0.526097768970181\\
0.6	0.527612965639898\\
0.5	0.524960278992595\\
0.4	0.525717711491841\\
0.3	0.616093135608393\\
0.2	0.553285558685015\\
0.1	0.614888079271617\\
0	0.914782956591318\\
}--cycle;
\addplot [color=white!55!mycolor4, forget plot]
  table[row sep=crcr]{%
0	0.914782956591318\\
0.1	0.123179534251087\\
0.2	0.546934485323787\\
0.3	0.123334749968722\\
0.4	0.0334341188742328\\
0.5	0.0335514233478727\\
0.6	0.0331391847901884\\
0.7	0.0333741254086949\\
0.8	0.123314711888319\\
0.9	0.25898752373223\\
};
\addplot [color=white!55!mycolor4, forget plot]
  table[row sep=crcr]{%
0	0.914782956591318\\
0.1	0.614888079271617\\
0.2	0.553285558685015\\
0.3	0.616093135608393\\
0.4	0.525717711491841\\
0.5	0.524960278992595\\
0.6	0.527612965639898\\
0.7	0.526097768970181\\
0.8	0.615793109675994\\
0.9	0.662556785129542\\
};
\addplot [color=mycolor4, dashdotted, line width=1.0pt, mark size=3.5pt, mark=square, mark options={solid, mycolor4}]
  table[row sep=crcr]{%
0	0.914782956591318\\
0.1	0.369033806761352\\
0.2	0.550110022004401\\
0.3	0.369713942788558\\
0.4	0.279575915183037\\
0.5	0.279255851170234\\
0.6	0.280376075215043\\
0.7	0.279735947189438\\
0.8	0.369553910782156\\
0.9	0.460772154430886\\
};
%\addlegendentry{DS}

\addplot[area legend, draw=none, fill=mycolor5, fill opacity=0.2, forget plot]
table[row sep=crcr] {%
x	y\\
0	0.889577915583117\\
0.1	0.242997391077585\\
0.2	0.547195830861016\\
0.3	0.123334749968722\\
0.4	0.0334341188742328\\
0.5	0.0335514233478727\\
0.6	0.0331391847901884\\
0.7	0.0333741254086949\\
0.8	0.123314711888319\\
0.9	0.25898752373223\\
0.9	0.662556785129542\\
0.8	0.615793109675994\\
0.7	0.526097768970181\\
0.6	0.527612965639898\\
0.5	0.524960278992595\\
0.4	0.525717711491841\\
0.3	0.616093135608393\\
0.2	0.553824373179792\\
0.1	0.654782164833598\\
0	0.889577915583117\\
}--cycle;
\addplot [color=white!55!mycolor5, forget plot]
  table[row sep=crcr]{%
0	0.889577915583117\\
0.1	0.242997391077585\\
0.2	0.547195830861016\\
0.3	0.123334749968722\\
0.4	0.0334341188742328\\
0.5	0.0335514233478727\\
0.6	0.0331391847901884\\
0.7	0.0333741254086949\\
0.8	0.123314711888319\\
0.9	0.25898752373223\\
};
\addplot [color=white!55!mycolor5, forget plot]
  table[row sep=crcr]{%
0	0.889577915583117\\
0.1	0.654782164833598\\
0.2	0.553824373179792\\
0.3	0.616093135608393\\
0.4	0.525717711491841\\
0.5	0.524960278992595\\
0.6	0.527612965639898\\
0.7	0.526097768970181\\
0.8	0.615793109675994\\
0.9	0.662556785129542\\
};
\addplot [color=mycolor5, dashdotted, line width=1.0pt, mark size=5.0pt, mark=+, mark options={solid, mycolor5}]
  table[row sep=crcr]{%
0	0.889577915583117\\
0.1	0.448889777955591\\
0.2	0.550510102020404\\
0.3	0.369713942788558\\
0.4	0.279575915183037\\
0.5	0.279255851170234\\
0.6	0.280376075215043\\
0.7	0.279735947189438\\
0.8	0.369553910782156\\
0.9	0.460772154430886\\
};
%\addlegendentry{MV}

\addplot[area legend, draw=none, fill=mycolor6, fill opacity=0.2, forget plot]
table[row sep=crcr] {%
x	y\\
0	0.888377675535107\\
0.1	0.128130222698089\\
0.2	0.546934485323787\\
0.3	0.123334749968722\\
0.4	0.0334341188742328\\
0.5	0.0335514233478727\\
0.6	0.0331391847901884\\
0.7	0.0333741254086949\\
0.8	0.123314711888319\\
0.9	0.25898752373223\\
0.9	0.662556785129542\\
0.8	0.615793109675994\\
0.7	0.526097768970181\\
0.6	0.527612965639898\\
0.5	0.524960278992595\\
0.4	0.525717711491841\\
0.3	0.616093135608393\\
0.2	0.553285558685015\\
0.1	0.621539711288708\\
0	0.888377675535107\\
}--cycle;
\addplot [color=white!55!mycolor6, forget plot]
  table[row sep=crcr]{%
0	0.888377675535107\\
0.1	0.128130222698089\\
0.2	0.546934485323787\\
0.3	0.123334749968722\\
0.4	0.0334341188742328\\
0.5	0.0335514233478727\\
0.6	0.0331391847901884\\
0.7	0.0333741254086949\\
0.8	0.123314711888319\\
0.9	0.25898752373223\\
};
\addplot [color=white!55!mycolor6, forget plot]
  table[row sep=crcr]{%
0	0.888377675535107\\
0.1	0.621539711288708\\
0.2	0.553285558685015\\
0.3	0.616093135608393\\
0.4	0.525717711491841\\
0.5	0.524960278992595\\
0.6	0.527612965639898\\
0.7	0.526097768970181\\
0.8	0.615793109675994\\
0.9	0.662556785129542\\
};
\addplot [color=mycolor6, dashdotted, line width=1.0pt, mark size=3.3pt, mark=triangle, mark options={solid, mycolor6}]
  table[row sep=crcr]{%
0	0.888377675535107\\
0.1	0.374834966993399\\
0.2	0.550110022004401\\
0.3	0.369713942788558\\
0.4	0.279575915183037\\
0.5	0.279255851170234\\
0.6	0.280376075215043\\
0.7	0.279735947189438\\
0.8	0.369553910782156\\
0.9	0.460772154430886\\
};
%\addlegendentry{PGD}

\end{axis}
\end{tikzpicture}%

%% file: fig/tikz/real_data/per_adv_groundtruth_dog_shaded_acc.tikz
\definecolor{mycolor1}{rgb}{0.00000,0.44700,0.74100}%
\definecolor{mycolor2}{rgb}{0.49400,0.18400,0.55600}%
\definecolor{mycolor3}{rgb}{0.63500,0.07800,0.18400}%
\definecolor{mycolor4}{rgb}{0.92900,0.69400,0.12500}%
\definecolor{mycolor5}{rgb}{0.30100,0.74500,0.93300}%
\definecolor{mycolor6}{rgb}{0.85000,0.32500,0.09800}%
\begin{tikzpicture}

\begin{axis}[%
width=4.464in,
height=3.286in,
at={(0.815in,0.76in)},
scale only axis,
xmin=0,
xmax=0.95,
xlabel style={font=\color{white!15!black}},
xlabel={Percentage of adversaries},
ymin=0,
ymax=1,
ylabel style={font=\color{white!15!black}},
ylabel={Accuracy},
axis background/.style={fill=white},
axis x line*=bottom,
axis y line*=left,
]

\addplot[area legend, draw=none, fill=mycolor1, fill opacity=0.2, forget plot]
table[row sep=crcr] {%
x	y\\
0	0.837670384138786\\
0.1	0.826553182488425\\
0.2	0.838273669398736\\
0.3	0.837845574710651\\
0.4	0.827891797769327\\
0.5	0.821583974237867\\
0.6	0.805175762292787\\
0.7	0.765183410356267\\
0.8	0.809912100918175\\
0.9	0.192341381996129\\
0.9	0.878786251213288\\
0.8	0.862950352613423\\
0.7	0.868026007239768\\
0.6	0.866199702391228\\
0.5	0.852765468141315\\
0.4	0.85785789244133\\
0.3	0.866243644620204\\
0.2	0.861850246338563\\
0.1	0.935528601898192\\
0	0.837670384138786\\
}--cycle;
\addplot [color=white!55!mycolor1, forget plot]
  table[row sep=crcr]{%
0	0.837670384138786\\
0.1	0.826553182488425\\
0.2	0.838273669398736\\
0.3	0.837845574710651\\
0.4	0.827891797769327\\
0.5	0.821583974237867\\
0.6	0.805175762292787\\
0.7	0.765183410356267\\
0.8	0.809912100918175\\
0.9	0.192341381996129\\
};
\addplot [color=white!55!mycolor1, forget plot]
  table[row sep=crcr]{%
0	0.837670384138786\\
0.1	0.935528601898192\\
0.2	0.861850246338563\\
0.3	0.866243644620204\\
0.4	0.85785789244133\\
0.5	0.852765468141315\\
0.6	0.866199702391228\\
0.7	0.868026007239768\\
0.8	0.862950352613423\\
0.9	0.878786251213288\\
};
\addplot [color=mycolor1, line width=1.0pt, mark size=5.0pt, mark=o, mark options={solid, mycolor1}]
  table[row sep=crcr]{%
0	0.837670384138786\\
0.1	0.881040892193309\\
0.2	0.850061957868649\\
0.3	0.852044609665427\\
0.4	0.842874845105328\\
0.5	0.837174721189591\\
0.6	0.835687732342008\\
0.7	0.816604708798017\\
0.8	0.836431226765799\\
0.9	0.535563816604709\\
};
%\addlegendentry{Alg. 1 - TA + DS}

\addplot[area legend, draw=none, fill=mycolor2, fill opacity=0.2, forget plot]
table[row sep=crcr] {%
x	y\\
0	0.837670384138786\\
0.1	0.833489083693467\\
0.2	0.838273669398736\\
0.3	0.837845574710651\\
0.4	0.827891797769327\\
0.5	0.0991325898389095\\
0.6	0.0991325898389095\\
0.7	0.0991325898389095\\
0.8	0.0991325898389095\\
0.9	0.0991325898389095\\
0.9	0.0991325898389095\\
0.8	0.0991325898389095\\
0.7	0.0991325898389095\\
0.6	0.0991325898389095\\
0.5	0.0991325898389095\\
0.4	0.85785789244133\\
0.3	0.866243644620204\\
0.2	0.861850246338563\\
0.1	0.888444001808392\\
0	0.837670384138786\\
}--cycle;
\addplot [color=white!55!mycolor2, forget plot]
  table[row sep=crcr]{%
0	0.837670384138786\\
0.1	0.833489083693467\\
0.2	0.838273669398736\\
0.3	0.837845574710651\\
0.4	0.827891797769327\\
0.5	0.0991325898389095\\
0.6	0.0991325898389095\\
0.7	0.0991325898389095\\
0.8	0.0991325898389095\\
0.9	0.0991325898389095\\
};
\addplot [color=white!55!mycolor2, forget plot]
  table[row sep=crcr]{%
0	0.837670384138786\\
0.1	0.888444001808392\\
0.2	0.861850246338563\\
0.3	0.866243644620204\\
0.4	0.85785789244133\\
0.5	0.0991325898389095\\
0.6	0.0991325898389095\\
0.7	0.0991325898389095\\
0.8	0.0991325898389095\\
0.9	0.0991325898389095\\
};
\addplot [color=mycolor2, line width=1.0pt, mark size=8.6pt, mark=diamond, mark options={solid, mycolor2}]
  table[row sep=crcr]{%
0	0.837670384138786\\
0.1	0.860966542750929\\
0.2	0.850061957868649\\
0.3	0.852044609665427\\
0.4	0.842874845105328\\
0.5	0.0991325898389095\\
0.6	0.0991325898389095\\
0.7	0.0991325898389095\\
0.8	0.0991325898389095\\
0.9	0.0991325898389095\\
};
%\addlegendentry{Alg. 1 - H + DS}

\addplot[area legend, draw=none, fill=mycolor3, fill opacity=0.2, forget plot]
table[row sep=crcr] {%
x	y\\
0	0.827757125154895\\
0.1	0.800905765300913\\
0.2	0.766732943716433\\
0.3	0.75943623382747\\
0.4	0.747769964259965\\
0.5	0.0991325898389095\\
0.6	0.0991325898389095\\
0.7	0.0991325898389095\\
0.8	0.0991325898389095\\
0.9	0.0991325898389095\\
0.9	0.0991325898389095\\
0.8	0.0991325898389095\\
0.7	0.0991325898389095\\
0.6	0.0991325898389095\\
0.5	0.0991325898389095\\
0.4	0.824968573534335\\
0.3	0.827180866544277\\
0.2	0.838719348724707\\
0.1	0.833294978193511\\
0	0.827757125154895\\
}--cycle;
\addplot [color=white!55!mycolor3, forget plot]
  table[row sep=crcr]{%
0	0.827757125154895\\
0.1	0.800905765300913\\
0.2	0.766732943716433\\
0.3	0.75943623382747\\
0.4	0.747769964259965\\
0.5	0.0991325898389095\\
0.6	0.0991325898389095\\
0.7	0.0991325898389095\\
0.8	0.0991325898389095\\
0.9	0.0991325898389095\\
};
\addplot [color=white!55!mycolor3, forget plot]
  table[row sep=crcr]{%
0	0.827757125154895\\
0.1	0.833294978193511\\
0.2	0.838719348724707\\
0.3	0.827180866544277\\
0.4	0.824968573534335\\
0.5	0.0991325898389095\\
0.6	0.0991325898389095\\
0.7	0.0991325898389095\\
0.8	0.0991325898389095\\
0.9	0.0991325898389095\\
};
\addplot [color=mycolor3, dashed, line width=1.0pt, mark size=5.0pt, mark=x, mark options={solid, mycolor3}]
  table[row sep=crcr]{%
0	0.827757125154895\\
0.1	0.817100371747212\\
0.2	0.80272614622057\\
0.3	0.793308550185874\\
0.4	0.78636926889715\\
0.5	0.0991325898389095\\
0.6	0.0991325898389095\\
0.7	0.0991325898389095\\
0.8	0.0991325898389095\\
0.9	0.0991325898389095\\
};
%\addlegendentry{MMSR}

\addplot[area legend, draw=none, fill=mycolor4, fill opacity=0.2, forget plot]
table[row sep=crcr] {%
x	y\\
0	0.833952912019826\\
0.1	0.859507862353285\\
0.2	0.242738993116968\\
0.3	0.0972807593667907\\
0.4	0.0988262532893433\\
0.5	0.0991325898389095\\
0.6	0.0991325898389095\\
0.7	0.0991325898389095\\
0.8	0.0991325898389095\\
0.9	0.0991325898389095\\
0.9	0.0991325898389095\\
0.8	0.0991325898389095\\
0.7	0.0991325898389095\\
0.6	0.0991325898389095\\
0.5	0.0991325898389095\\
0.4	0.0999345893376703\\
0.3	0.0994974314634448\\
0.2	0.485885542199017\\
0.1	0.922400440001114\\
0	0.833952912019826\\
}--cycle;
\addplot [color=white!55!mycolor4, forget plot]
  table[row sep=crcr]{%
0	0.833952912019826\\
0.1	0.859507862353285\\
0.2	0.242738993116968\\
0.3	0.0972807593667907\\
0.4	0.0988262532893433\\
0.5	0.0991325898389095\\
0.6	0.0991325898389095\\
0.7	0.0991325898389095\\
0.8	0.0991325898389095\\
0.9	0.0991325898389095\\
};
\addplot [color=white!55!mycolor4, forget plot]
  table[row sep=crcr]{%
0	0.833952912019826\\
0.1	0.922400440001114\\
0.2	0.485885542199017\\
0.3	0.0994974314634448\\
0.4	0.0999345893376703\\
0.5	0.0991325898389095\\
0.6	0.0991325898389095\\
0.7	0.0991325898389095\\
0.8	0.0991325898389095\\
0.9	0.0991325898389095\\
};
\addplot [color=mycolor4, dashdotted, line width=1.0pt, mark size=3.5pt, mark=square, mark options={solid, mycolor4}]
  table[row sep=crcr]{%
0	0.833952912019826\\
0.1	0.890954151177199\\
0.2	0.364312267657993\\
0.3	0.0983890954151177\\
0.4	0.0993804213135068\\
0.5	0.0991325898389095\\
0.6	0.0991325898389095\\
0.7	0.0991325898389095\\
0.8	0.0991325898389095\\
0.9	0.0991325898389095\\
};
%\addlegendentry{DS}

\addplot[area legend, draw=none, fill=mycolor5, fill opacity=0.2, forget plot]
table[row sep=crcr] {%
x	y\\
0	0.81908302354399\\
0.1	0.753277163285738\\
0.2	0.411872710515053\\
0.3	0.130435991399085\\
0.4	0.103467966413055\\
0.5	0.099087162296736\\
0.6	0.0991325898389095\\
0.7	0.0991325898389095\\
0.8	0.0991325898389095\\
0.9	0.0991325898389095\\
0.9	0.0991325898389095\\
0.8	0.0991325898389095\\
0.7	0.0991325898389095\\
0.6	0.0991325898389095\\
0.5	0.101160669177861\\
0.4	0.112641079435767\\
0.3	0.189266610831398\\
0.2	0.427780325420511\\
0.1	0.802113171286753\\
0	0.81908302354399\\
}--cycle;
\addplot [color=white!55!mycolor5, forget plot]
  table[row sep=crcr]{%
0	0.81908302354399\\
0.1	0.753277163285738\\
0.2	0.411872710515053\\
0.3	0.130435991399085\\
0.4	0.103467966413055\\
0.5	0.099087162296736\\
0.6	0.0991325898389095\\
0.7	0.0991325898389095\\
0.8	0.0991325898389095\\
0.9	0.0991325898389095\\
};
\addplot [color=white!55!mycolor5, forget plot]
  table[row sep=crcr]{%
0	0.81908302354399\\
0.1	0.802113171286753\\
0.2	0.427780325420511\\
0.3	0.189266610831398\\
0.4	0.112641079435767\\
0.5	0.101160669177861\\
0.6	0.0991325898389095\\
0.7	0.0991325898389095\\
0.8	0.0991325898389095\\
0.9	0.0991325898389095\\
};
\addplot [color=mycolor5, dashdotted, line width=1.0pt, mark size=5.0pt, mark=+, mark options={solid, mycolor5}]
  table[row sep=crcr]{%
0	0.81908302354399\\
0.1	0.777695167286245\\
0.2	0.419826517967782\\
0.3	0.159851301115242\\
0.4	0.108054522924411\\
0.5	0.100123915737299\\
0.6	0.0991325898389095\\
0.7	0.0991325898389095\\
0.8	0.0991325898389095\\
0.9	0.0991325898389095\\
};
%\addlegendentry{MV}

\addplot[area legend, draw=none, fill=mycolor6, fill opacity=0.2, forget plot]
table[row sep=crcr] {%
x	y\\
0	0.830235439900867\\
0.1	0.814023161220133\\
0.2	0.607492466084449\\
0.3	0.104590516754478\\
0.4	0.0988262532893433\\
0.5	0.0991325898389095\\
0.6	0.0991325898389095\\
0.7	0.0991325898389095\\
0.8	0.0991325898389095\\
0.9	0.0991325898389095\\
0.9	0.0991325898389095\\
0.8	0.0991325898389095\\
0.7	0.0991325898389095\\
0.6	0.0991325898389095\\
0.5	0.0991325898389095\\
0.4	0.0999345893376703\\
0.3	0.120936125129041\\
0.2	0.710475315823854\\
0.1	0.828108189461404\\
0	0.830235439900867\\
}--cycle;
\addplot [color=white!55!mycolor6, forget plot]
  table[row sep=crcr]{%
0	0.830235439900867\\
0.1	0.814023161220133\\
0.2	0.607492466084449\\
0.3	0.104590516754478\\
0.4	0.0988262532893433\\
0.5	0.0991325898389095\\
0.6	0.0991325898389095\\
0.7	0.0991325898389095\\
0.8	0.0991325898389095\\
0.9	0.0991325898389095\\
};
\addplot [color=white!55!mycolor6, forget plot]
  table[row sep=crcr]{%
0	0.830235439900867\\
0.1	0.828108189461404\\
0.2	0.710475315823854\\
0.3	0.120936125129041\\
0.4	0.0999345893376703\\
0.5	0.0991325898389095\\
0.6	0.0991325898389095\\
0.7	0.0991325898389095\\
0.8	0.0991325898389095\\
0.9	0.0991325898389095\\
};
\addplot [color=mycolor6, dashdotted, line width=1.0pt, mark size=3.3pt, mark=triangle, mark options={solid, mycolor6}]
  table[row sep=crcr]{%
0	0.830235439900867\\
0.1	0.821065675340768\\
0.2	0.658983890954151\\
0.3	0.11276332094176\\
0.4	0.0993804213135068\\
0.5	0.0991325898389095\\
0.6	0.0991325898389095\\
0.7	0.0991325898389095\\
0.8	0.0991325898389095\\
0.9	0.0991325898389095\\
};
%\addlegendentry{PGD}

\end{axis}
\end{tikzpicture}%

%% file: fig/tikz/real_data/per_adv_groundtruth_web_shaded_acc.tikz
\definecolor{mycolor1}{rgb}{0.00000,0.44700,0.74100}%
\definecolor{mycolor2}{rgb}{0.49400,0.18400,0.55600}%
\definecolor{mycolor3}{rgb}{0.63500,0.07800,0.18400}%
\definecolor{mycolor4}{rgb}{0.92900,0.69400,0.12500}%
\definecolor{mycolor5}{rgb}{0.30100,0.74500,0.93300}%
\definecolor{mycolor6}{rgb}{0.85000,0.32500,0.09800}%
\begin{tikzpicture}

\begin{axis}[%
width=4.464in,
height=3.286in,
at={(0.815in,0.76in)},
scale only axis,
xmin=0,
xmax=0.95,
xlabel style={font=\color{white!15!black}},
xlabel={Percentage of adversaries},
ymin=0,
ymax=1,
ylabel style={font=\color{white!15!black}},
ylabel={Accuracy},
axis background/.style={fill=white},
axis x line*=bottom,
axis y line*=left,
]

\addplot[area legend, draw=none, fill=mycolor1, fill opacity=0.2, forget plot]
table[row sep=crcr] {%
x	y\\
0	0.822345890410959\\
0.1	0.000159336627106543\\
0.2	0.840271337636814\\
0.3	0.82341310943461\\
0.4	0.896128298765028\\
0.5	0.853172800882732\\
0.6	0.820544499661089\\
0.7	0.609670342794354\\
0.8	0.37209890745974\\
0.9	0.0948785388401101\\
0.9	0.43478599187983\\
0.8	0.935100489449419\\
0.7	0.982640248988533\\
0.6	0.877608534639702\\
0.5	0.887799683097668\\
0.4	0.9553605817476\\
0.3	0.946960053022986\\
0.2	0.941331841918566\\
0.1	0.657875488742498\\
0	0.822345890410959\\
}--cycle;
\addplot [color=white!55!mycolor1, forget plot]
  table[row sep=crcr]{%
0	0.822345890410959\\
0.1	0.000159336627106543\\
0.2	0.840271337636814\\
0.3	0.82341310943461\\
0.4	0.896128298765028\\
0.5	0.853172800882732\\
0.6	0.820544499661089\\
0.7	0.609670342794354\\
0.8	0.37209890745974\\
0.9	0.0948785388401101\\
};
\addplot [color=white!55!mycolor1, forget plot]
  table[row sep=crcr]{%
0	0.822345890410959\\
0.1	0.657875488742498\\
0.2	0.941331841918566\\
0.3	0.946960053022986\\
0.4	0.9553605817476\\
0.5	0.887799683097668\\
0.6	0.877608534639702\\
0.7	0.982640248988533\\
0.8	0.935100489449419\\
0.9	0.43478599187983\\
};
\addplot [color=mycolor1, line width=1.0pt, mark size=5.0pt, mark=o, mark options={solid, mycolor1}]
  table[row sep=crcr]{%
0	0.822345890410959\\
0.1	0.329017412684802\\
0.2	0.89080158977769\\
0.3	0.885186581228798\\
0.4	0.925744440256314\\
0.5	0.8704862419902\\
0.6	0.849076517150396\\
0.7	0.796155295891444\\
0.8	0.65359969845458\\
0.9	0.26483226535997\\
};
%\addlegendentry{Alg. 1 - TA + DS}

\addplot[area legend, draw=none, fill=mycolor2, fill opacity=0.2, forget plot]
table[row sep=crcr] {%
x	y\\
0	0.822345890410959\\
0.1	0.00851864363306176\\
0.2	0.840271337636814\\
0.3	0.82341310943461\\
0.4	0.896128298765028\\
0.5	0.0997005551485111\\
0.6	0.100263852242744\\
0.7	0.0994252752086726\\
0.8	0.0997005551485111\\
0.9	0.0994961703631609\\
0.9	0.10012689786149\\
0.8	0.100374831206559\\
0.7	0.100348565726118\\
0.6	0.100263852242744\\
0.5	0.100374831206559\\
0.4	0.9553605817476\\
0.3	0.946960053022986\\
0.2	0.941331841918566\\
0.1	0.658586476521123\\
0	0.822345890410959\\
}--cycle;
\addplot [color=white!55!mycolor2, forget plot]
  table[row sep=crcr]{%
0	0.822345890410959\\
0.1	0.00851864363306176\\
0.2	0.840271337636814\\
0.3	0.82341310943461\\
0.4	0.896128298765028\\
0.5	0.0997005551485111\\
0.6	0.100263852242744\\
0.7	0.0994252752086726\\
0.8	0.0997005551485111\\
0.9	0.0994961703631609\\
};
\addplot [color=white!55!mycolor2, forget plot]
  table[row sep=crcr]{%
0	0.822345890410959\\
0.1	0.658586476521123\\
0.2	0.941331841918566\\
0.3	0.946960053022986\\
0.4	0.9553605817476\\
0.5	0.100374831206559\\
0.6	0.100263852242744\\
0.7	0.100348565726118\\
0.8	0.100374831206559\\
0.9	0.10012689786149\\
};
\addplot [color=mycolor2, line width=1.0pt, mark size=8.6pt, mark=diamond, mark options={solid, mycolor2}]
  table[row sep=crcr]{%
0	0.822345890410959\\
0.1	0.333552560077092\\
0.2	0.89080158977769\\
0.3	0.885186581228798\\
0.4	0.925744440256314\\
0.5	0.100037693177535\\
0.6	0.100263852242744\\
0.7	0.0998869204673954\\
0.8	0.100037693177535\\
0.9	0.0998115341123257\\
};
%\addlegendentry{Alg. 1 - H + DS}

\addplot[area legend, draw=none, fill=mycolor3, fill opacity=0.2, forget plot]
table[row sep=crcr] {%
x	y\\
0	0.690924657534247\\
0.1	0.668140384932082\\
0.2	0.325120928489682\\
0.3	0.657362311888172\\
0.4	0.128993953485822\\
0.5	0.0997005551485111\\
0.6	0.100263852242744\\
0.7	0.0994252752086726\\
0.8	0.0997005551485111\\
0.9	0.0994961703631609\\
0.9	0.10012689786149\\
0.8	0.100374831206559\\
0.7	0.100348565726118\\
0.6	0.100263852242744\\
0.5	0.100374831206559\\
0.4	0.802329077045652\\
0.3	0.758242663611263\\
0.2	0.902297250216613\\
0.1	0.760871703408888\\
0	0.690924657534247\\
}--cycle;
\addplot [color=white!55!mycolor3, forget plot]
  table[row sep=crcr]{%
0	0.690924657534247\\
0.1	0.668140384932082\\
0.2	0.325120928489682\\
0.3	0.657362311888172\\
0.4	0.128993953485822\\
0.5	0.0997005551485111\\
0.6	0.100263852242744\\
0.7	0.0994252752086726\\
0.8	0.0997005551485111\\
0.9	0.0994961703631609\\
};
\addplot [color=white!55!mycolor3, forget plot]
  table[row sep=crcr]{%
0	0.690924657534247\\
0.1	0.760871703408888\\
0.2	0.902297250216613\\
0.3	0.758242663611263\\
0.4	0.802329077045652\\
0.5	0.100374831206559\\
0.6	0.100263852242744\\
0.7	0.100348565726118\\
0.8	0.100374831206559\\
0.9	0.10012689786149\\
};
\addplot [color=mycolor3, dashed, line width=1.0pt, mark size=5.0pt, mark=x, mark options={solid, mycolor3}]
  table[row sep=crcr]{%
0	0.690924657534247\\
0.1	0.714506044170485\\
0.2	0.613709089353147\\
0.3	0.707802487749717\\
0.4	0.465661515265737\\
0.5	0.100037693177535\\
0.6	0.100263852242744\\
0.7	0.0998869204673954\\
0.8	0.100037693177535\\
0.9	0.0998115341123257\\
};
%\addlegendentry{MMSR}

\addplot[area legend, draw=none, fill=mycolor4, fill opacity=0.2, forget plot]
table[row sep=crcr] {%
x	y\\
0	0.835188356164384\\
0.1	0.0810938836440514\\
0.2	0.099203466713146\\
0.3	0.0994961703631609\\
0.4	0.0993063794100644\\
0.5	0.0997005551485111\\
0.6	0.100263852242744\\
0.7	0.0994252752086726\\
0.8	0.0997005551485111\\
0.9	0.0994961703631609\\
0.9	0.10012689786149\\
0.8	0.100374831206559\\
0.7	0.100348565726118\\
0.6	0.100263852242744\\
0.5	0.100374831206559\\
0.4	0.100165916104447\\
0.3	0.10012689786149\\
0.2	0.0998308139802773\\
0.1	0.254093889432374\\
0	0.835188356164384\\
}--cycle;
\addplot [color=white!55!mycolor4, forget plot]
  table[row sep=crcr]{%
0	0.835188356164384\\
0.1	0.0810938836440514\\
0.2	0.099203466713146\\
0.3	0.0994961703631609\\
0.4	0.0993063794100644\\
0.5	0.0997005551485111\\
0.6	0.100263852242744\\
0.7	0.0994252752086726\\
0.8	0.0997005551485111\\
0.9	0.0994961703631609\\
};
\addplot [color=white!55!mycolor4, forget plot]
  table[row sep=crcr]{%
0	0.835188356164384\\
0.1	0.254093889432374\\
0.2	0.0998308139802773\\
0.3	0.10012689786149\\
0.4	0.100165916104447\\
0.5	0.100374831206559\\
0.6	0.100263852242744\\
0.7	0.100348565726118\\
0.8	0.100374831206559\\
0.9	0.10012689786149\\
};
\addplot [color=mycolor4, dashdotted, line width=1.0pt, mark size=3.5pt, mark=square, mark options={solid, mycolor4}]
  table[row sep=crcr]{%
0	0.835188356164384\\
0.1	0.167593886538213\\
0.2	0.0995171403467116\\
0.3	0.0998115341123257\\
0.4	0.0997361477572559\\
0.5	0.100037693177535\\
0.6	0.100263852242744\\
0.7	0.0998869204673954\\
0.8	0.100037693177535\\
0.9	0.0998115341123257\\
};
%\addlegendentry{DS}

\addplot[area legend, draw=none, fill=mycolor5, fill opacity=0.2, forget plot]
table[row sep=crcr] {%
x	y\\
0	0.787671232876712\\
0.1	0.206833807602169\\
0.2	0.104819071970848\\
0.3	0.0995493987504673\\
0.4	0.0993063794100644\\
0.5	0.0997005551485111\\
0.6	0.100263852242744\\
0.7	0.0994252752086726\\
0.8	0.0997005551485111\\
0.9	0.0994961703631609\\
0.9	0.10012689786149\\
0.8	0.100374831206559\\
0.7	0.100348565726118\\
0.6	0.100263852242744\\
0.5	0.100374831206559\\
0.4	0.100165916104447\\
0.3	0.100375214894463\\
0.2	0.108098020692133\\
0.1	0.279360731947196\\
0	0.787671232876712\\
}--cycle;
\addplot [color=white!55!mycolor5, forget plot]
  table[row sep=crcr]{%
0	0.787671232876712\\
0.1	0.206833807602169\\
0.2	0.104819071970848\\
0.3	0.0995493987504673\\
0.4	0.0993063794100644\\
0.5	0.0997005551485111\\
0.6	0.100263852242744\\
0.7	0.0994252752086726\\
0.8	0.0997005551485111\\
0.9	0.0994961703631609\\
};
\addplot [color=white!55!mycolor5, forget plot]
  table[row sep=crcr]{%
0	0.787671232876712\\
0.1	0.279360731947196\\
0.2	0.108098020692133\\
0.3	0.100375214894463\\
0.4	0.100165916104447\\
0.5	0.100374831206559\\
0.6	0.100263852242744\\
0.7	0.100348565726118\\
0.8	0.100374831206559\\
0.9	0.10012689786149\\
};
\addplot [color=mycolor5, dashdotted, line width=1.0pt, mark size=5.0pt, mark=+, mark options={solid, mycolor5}]
  table[row sep=crcr]{%
0	0.787671232876712\\
0.1	0.243097269774683\\
0.2	0.106458546331491\\
0.3	0.0999623068224651\\
0.4	0.0997361477572559\\
0.5	0.100037693177535\\
0.6	0.100263852242744\\
0.7	0.0998869204673954\\
0.8	0.100037693177535\\
0.9	0.0998115341123257\\
};
%\addlegendentry{MV}

\addplot[area legend, draw=none, fill=mycolor6, fill opacity=0.2, forget plot]
table[row sep=crcr] {%
x	y\\
0	0.899400684931507\\
0.1	0.103099949093193\\
0.2	0.099203466713146\\
0.3	0.0994961703631609\\
0.4	0.0993063794100644\\
0.5	0.0997005551485111\\
0.6	0.100263852242744\\
0.7	0.0994252752086726\\
0.8	0.0997005551485111\\
0.9	0.0994961703631609\\
0.9	0.10012689786149\\
0.8	0.100374831206559\\
0.7	0.100348565726118\\
0.6	0.100263852242744\\
0.5	0.100374831206559\\
0.4	0.100165916104447\\
0.3	0.10012689786149\\
0.2	0.0998308139802773\\
0.1	0.113594310242758\\
0	0.899400684931507\\
}--cycle;
\addplot [color=white!55!mycolor6, forget plot]
  table[row sep=crcr]{%
0	0.899400684931507\\
0.1	0.103099949093193\\
0.2	0.099203466713146\\
0.3	0.0994961703631609\\
0.4	0.0993063794100644\\
0.5	0.0997005551485111\\
0.6	0.100263852242744\\
0.7	0.0994252752086726\\
0.8	0.0997005551485111\\
0.9	0.0994961703631609\\
};
\addplot [color=white!55!mycolor6, forget plot]
  table[row sep=crcr]{%
0	0.899400684931507\\
0.1	0.113594310242758\\
0.2	0.0998308139802773\\
0.3	0.10012689786149\\
0.4	0.100165916104447\\
0.5	0.100374831206559\\
0.6	0.100263852242744\\
0.7	0.100348565726118\\
0.8	0.100374831206559\\
0.9	0.10012689786149\\
};
\addplot [color=mycolor6, dashdotted, line width=1.0pt, mark size=3.3pt, mark=triangle, mark options={solid, mycolor6}]
  table[row sep=crcr]{%
0	0.899400684931507\\
0.1	0.108347129667976\\
0.2	0.0995171403467116\\
0.3	0.0998115341123257\\
0.4	0.0997361477572559\\
0.5	0.100037693177535\\
0.6	0.100263852242744\\
0.7	0.0998869204673954\\
0.8	0.100037693177535\\
0.9	0.0998115341123257\\
};
%\addlegendentry{PGD}

\end{axis}
\end{tikzpicture}%

%% file: fig/tikz/real_data/per_adv_groundtruth_Adult2_shaded_acc.tikz
\definecolor{mycolor1}{rgb}{0.00000,0.44700,0.74100}%
\definecolor{mycolor2}{rgb}{0.49400,0.18400,0.55600}%
\definecolor{mycolor3}{rgb}{0.63500,0.07800,0.18400}%
\definecolor{mycolor4}{rgb}{0.92900,0.69400,0.12500}%
\definecolor{mycolor5}{rgb}{0.30100,0.74500,0.93300}%
\definecolor{mycolor6}{rgb}{0.85000,0.32500,0.09800}%
\begin{tikzpicture}

\begin{axis}[%
width=4.464in,
height=3.286in,
at={(0.815in,0.76in)},
scale only axis,
xmin=0,
xmax=0.95,
xlabel style={font=\color{white!15!black}},
xlabel={Percentage of adversaries},
ymin=0,
ymax=0.9,
ylabel style={font=\color{white!15!black}},
ylabel={Accuracy},
axis background/.style={fill=white},
axis x line*=bottom,
axis y line*=left,
]

\addplot[area legend, draw=none, fill=mycolor1, fill opacity=0.2, forget plot]
table[row sep=crcr] {%
x	y\\
0	0.774509803921569\\
0.1	0.762918279275655\\
0.2	0.738725677953644\\
0.3	0.738741841503983\\
0.4	0.733639456104344\\
0.5	0.721939637338209\\
0.6	0.693629149446991\\
0.7	0.733576441117248\\
0.8	0.698063219875031\\
0.9	0.552900689561069\\
0.9	0.771947795287415\\
0.8	0.787997386185575\\
0.7	0.765817498276691\\
0.6	0.7572799414621\\
0.5	0.782302786904215\\
0.4	0.763330240865353\\
0.3	0.766712703950563\\
0.2	0.759456140228174\\
0.1	0.788735118930341\\
0	0.774509803921569\\
}--cycle;
\addplot [color=white!55!mycolor1, forget plot]
  table[row sep=crcr]{%
0	0.774509803921569\\
0.1	0.762918279275655\\
0.2	0.738725677953644\\
0.3	0.738741841503983\\
0.4	0.733639456104344\\
0.5	0.721939637338209\\
0.6	0.693629149446991\\
0.7	0.733576441117248\\
0.8	0.698063219875031\\
0.9	0.552900689561069\\
};
\addplot [color=white!55!mycolor1, forget plot]
  table[row sep=crcr]{%
0	0.774509803921569\\
0.1	0.788735118930341\\
0.2	0.759456140228174\\
0.3	0.766712703950563\\
0.4	0.763330240865353\\
0.5	0.782302786904215\\
0.6	0.7572799414621\\
0.7	0.765817498276691\\
0.8	0.787997386185575\\
0.9	0.771947795287415\\
};
\addplot [color=mycolor1, line width=1.0pt, mark size=5.0pt, mark=o, mark options={solid, mycolor1}]
  table[row sep=crcr]{%
0	0.774509803921569\\
0.1	0.775826699102998\\
0.2	0.749090909090909\\
0.3	0.752727272727273\\
0.4	0.748484848484848\\
0.5	0.752121212121212\\
0.6	0.725454545454545\\
0.7	0.74969696969697\\
0.8	0.743030303030303\\
0.9	0.662424242424242\\
};
%\addlegendentry{Alg. 1 - TA + DS}

\addplot[area legend, draw=none, fill=mycolor2, fill opacity=0.2, forget plot]
table[row sep=crcr] {%
x	y\\
0	0.774509803921569\\
0.1	0.762918279275655\\
0.2	0.738725677953644\\
0.3	0.738741841503983\\
0.4	0.733639456104344\\
0.5	0.096969696969697\\
0.6	0.096969696969697\\
0.7	0.096969696969697\\
0.8	0.096969696969697\\
0.9	0.096969696969697\\
0.9	0.096969696969697\\
0.8	0.096969696969697\\
0.7	0.096969696969697\\
0.6	0.096969696969697\\
0.5	0.096969696969697\\
0.4	0.763330240865353\\
0.3	0.766712703950563\\
0.2	0.759456140228174\\
0.1	0.788735118930341\\
0	0.774509803921569\\
}--cycle;
\addplot [color=white!55!mycolor2, forget plot]
  table[row sep=crcr]{%
0	0.774509803921569\\
0.1	0.762918279275655\\
0.2	0.738725677953644\\
0.3	0.738741841503983\\
0.4	0.733639456104344\\
0.5	0.096969696969697\\
0.6	0.096969696969697\\
0.7	0.096969696969697\\
0.8	0.096969696969697\\
0.9	0.096969696969697\\
};
\addplot [color=white!55!mycolor2, forget plot]
  table[row sep=crcr]{%
0	0.774509803921569\\
0.1	0.788735118930341\\
0.2	0.759456140228174\\
0.3	0.766712703950563\\
0.4	0.763330240865353\\
0.5	0.096969696969697\\
0.6	0.096969696969697\\
0.7	0.096969696969697\\
0.8	0.096969696969697\\
0.9	0.096969696969697\\
};
\addplot [color=mycolor2, line width=1.0pt, mark size=8.6pt, mark=diamond, mark options={solid, mycolor2}]
  table[row sep=crcr]{%
0	0.774509803921569\\
0.1	0.775826699102998\\
0.2	0.749090909090909\\
0.3	0.752727272727273\\
0.4	0.748484848484848\\
0.5	0.096969696969697\\
0.6	0.096969696969697\\
0.7	0.096969696969697\\
0.8	0.096969696969697\\
0.9	0.096969696969697\\
};
%\addlegendentry{Alg. 1 - H + DS}

\addplot[area legend, draw=none, fill=mycolor3, fill opacity=0.2, forget plot]
table[row sep=crcr] {%
x	y\\
0	0.627990395544308\\
0.1	0.75089711114993\\
0.2	0.697640792336524\\
0.3	0.672994007445722\\
0.4	0.662931263391673\\
0.5	0.096969696969697\\
0.6	0.096969696969697\\
0.7	0.096969696969697\\
0.8	0.096969696969697\\
0.9	0.096969696969697\\
0.9	0.096969696969697\\
0.8	0.096969696969697\\
0.7	0.096969696969697\\
0.6	0.096969696969697\\
0.5	0.096969696969697\\
0.4	0.761311160850751\\
0.3	0.746399931948218\\
0.2	0.739934965239234\\
0.1	0.771187319202638\\
0	0.693578231906672\\
}--cycle;
\addplot [color=white!55!mycolor3, forget plot]
  table[row sep=crcr]{%
0	0.627990395544308\\
0.1	0.75089711114993\\
0.2	0.697640792336524\\
0.3	0.672994007445722\\
0.4	0.662931263391673\\
0.5	0.096969696969697\\
0.6	0.096969696969697\\
0.7	0.096969696969697\\
0.8	0.096969696969697\\
0.9	0.096969696969697\\
};
\addplot [color=white!55!mycolor3, forget plot]
  table[row sep=crcr]{%
0	0.693578231906672\\
0.1	0.771187319202638\\
0.2	0.739934965239234\\
0.3	0.746399931948218\\
0.4	0.761311160850751\\
0.5	0.096969696969697\\
0.6	0.096969696969697\\
0.7	0.096969696969697\\
0.8	0.096969696969697\\
0.9	0.096969696969697\\
};
\addplot [color=mycolor3, dashed, line width=1.0pt, mark size=5.0pt, mark=x, mark options={solid, mycolor3}]
  table[row sep=crcr]{%
0	0.66078431372549\\
0.1	0.761042215176284\\
0.2	0.718787878787879\\
0.3	0.70969696969697\\
0.4	0.712121212121212\\
0.5	0.096969696969697\\
0.6	0.096969696969697\\
0.7	0.096969696969697\\
0.8	0.096969696969697\\
0.9	0.096969696969697\\
};
%\addlegendentry{MMSR}

\addplot[area legend, draw=none, fill=mycolor4, fill opacity=0.2, forget plot]
table[row sep=crcr] {%
x	y\\
0	0.774509803921569\\
0.1	0.634918861514998\\
0.2	0.0858410569575005\\
0.3	0.0968586059801998\\
0.4	0.0950084436500001\\
0.5	0.096969696969697\\
0.6	0.096969696969697\\
0.7	0.096969696969697\\
0.8	0.096969696969697\\
0.9	0.096969696969697\\
0.9	0.096969696969697\\
0.8	0.096969696969697\\
0.7	0.096969696969697\\
0.6	0.096969696969697\\
0.5	0.096969696969697\\
0.4	0.0977188290772726\\
0.3	0.101929272807679\\
0.2	0.163855912739469\\
0.1	0.689491840250725\\
0	0.774509803921569\\
}--cycle;
\addplot [color=white!55!mycolor4, forget plot]
  table[row sep=crcr]{%
0	0.774509803921569\\
0.1	0.634918861514998\\
0.2	0.0858410569575005\\
0.3	0.0968586059801998\\
0.4	0.0950084436500001\\
0.5	0.096969696969697\\
0.6	0.096969696969697\\
0.7	0.096969696969697\\
0.8	0.096969696969697\\
0.9	0.096969696969697\\
};
\addplot [color=white!55!mycolor4, forget plot]
  table[row sep=crcr]{%
0	0.774509803921569\\
0.1	0.689491840250725\\
0.2	0.163855912739469\\
0.3	0.101929272807679\\
0.4	0.0977188290772726\\
0.5	0.096969696969697\\
0.6	0.096969696969697\\
0.7	0.096969696969697\\
0.8	0.096969696969697\\
0.9	0.096969696969697\\
};
\addplot [color=mycolor4, dashdotted, line width=1.0pt, mark size=3.5pt, mark=square, mark options={solid, mycolor4}]
  table[row sep=crcr]{%
0	0.774509803921569\\
0.1	0.662205350882862\\
0.2	0.124848484848485\\
0.3	0.0993939393939394\\
0.4	0.0963636363636364\\
0.5	0.096969696969697\\
0.6	0.096969696969697\\
0.7	0.096969696969697\\
0.8	0.096969696969697\\
0.9	0.096969696969697\\
};
%\addlegendentry{DS}

\addplot[area legend, draw=none, fill=mycolor5, fill opacity=0.2, forget plot]
table[row sep=crcr] {%
x	y\\
0	0.76797385620915\\
0.1	0.592740274347126\\
0.2	0.276038983080757\\
0.3	0.112474904324243\\
0.4	0.0962205648621213\\
0.5	0.096969696969697\\
0.6	0.096969696969697\\
0.7	0.096969696969697\\
0.8	0.096969696969697\\
0.9	0.096969696969697\\
0.9	0.096969696969697\\
0.8	0.096969696969697\\
0.7	0.096969696969697\\
0.6	0.096969696969697\\
0.5	0.096969696969697\\
0.4	0.0989309502893938\\
0.3	0.128737216887878\\
0.2	0.297294350252576\\
0.1	0.627234821665943\\
0	0.76797385620915\\
}--cycle;
\addplot [color=white!55!mycolor5, forget plot]
  table[row sep=crcr]{%
0	0.76797385620915\\
0.1	0.592740274347126\\
0.2	0.276038983080757\\
0.3	0.112474904324243\\
0.4	0.0962205648621213\\
0.5	0.096969696969697\\
0.6	0.096969696969697\\
0.7	0.096969696969697\\
0.8	0.096969696969697\\
0.9	0.096969696969697\\
};
\addplot [color=white!55!mycolor5, forget plot]
  table[row sep=crcr]{%
0	0.76797385620915\\
0.1	0.627234821665943\\
0.2	0.297294350252576\\
0.3	0.128737216887878\\
0.4	0.0989309502893938\\
0.5	0.096969696969697\\
0.6	0.096969696969697\\
0.7	0.096969696969697\\
0.8	0.096969696969697\\
0.9	0.096969696969697\\
};
\addplot [color=mycolor5, dashdotted, line width=1.0pt, mark size=5.0pt, mark=+, mark options={solid, mycolor5}]
  table[row sep=crcr]{%
0	0.76797385620915\\
0.1	0.609987548006534\\
0.2	0.286666666666667\\
0.3	0.120606060606061\\
0.4	0.0975757575757576\\
0.5	0.096969696969697\\
0.6	0.096969696969697\\
0.7	0.096969696969697\\
0.8	0.096969696969697\\
0.9	0.096969696969697\\
};
%\addlegendentry{MV}

\addplot[area legend, draw=none, fill=mycolor6, fill opacity=0.2, forget plot]
table[row sep=crcr] {%
x	y\\
0	0.764705882352941\\
0.1	0.65477443815649\\
0.2	0.160606060606061\\
0.3	0.096969696969697\\
0.4	0.096969696969697\\
0.5	0.096969696969697\\
0.6	0.096969696969697\\
0.7	0.096969696969697\\
0.8	0.096969696969697\\
0.9	0.096969696969697\\
0.9	0.096969696969697\\
0.8	0.096969696969697\\
0.7	0.096969696969697\\
0.6	0.096969696969697\\
0.5	0.096969696969697\\
0.4	0.096969696969697\\
0.3	0.096969696969697\\
0.2	0.196969696969697\\
0.1	0.670791616819895\\
0	0.764705882352941\\
}--cycle;
\addplot [color=white!55!mycolor6, forget plot]
  table[row sep=crcr]{%
0	0.764705882352941\\
0.1	0.65477443815649\\
0.2	0.160606060606061\\
0.3	0.096969696969697\\
0.4	0.096969696969697\\
0.5	0.096969696969697\\
0.6	0.096969696969697\\
0.7	0.096969696969697\\
0.8	0.096969696969697\\
0.9	0.096969696969697\\
};
\addplot [color=white!55!mycolor6, forget plot]
  table[row sep=crcr]{%
0	0.764705882352941\\
0.1	0.670791616819895\\
0.2	0.196969696969697\\
0.3	0.096969696969697\\
0.4	0.096969696969697\\
0.5	0.096969696969697\\
0.6	0.096969696969697\\
0.7	0.096969696969697\\
0.8	0.096969696969697\\
0.9	0.096969696969697\\
};
\addplot [color=mycolor6, dashdotted, line width=1.0pt, mark size=3.3pt, mark=triangle, mark options={solid, mycolor6}]
  table[row sep=crcr]{%
0	0.764705882352941\\
0.1	0.662783027488193\\
0.2	0.178787878787879\\
0.3	0.096969696969697\\
0.4	0.096969696969697\\
0.5	0.096969696969697\\
0.6	0.096969696969697\\
0.7	0.096969696969697\\
0.8	0.096969696969697\\
0.9	0.096969696969697\\
};
%\addlegendentry{PGD}

\end{axis}
\end{tikzpicture}%

%% file: fig/tikz/real_data/per_adv_DS_bluebird_shaded_acc.tikz
\definecolor{mycolor1}{rgb}{0.00000,0.44700,0.74100}%
\definecolor{mycolor2}{rgb}{0.49400,0.18400,0.55600}%
\definecolor{mycolor3}{rgb}{0.63500,0.07800,0.18400}%
\definecolor{mycolor4}{rgb}{0.92900,0.69400,0.12500}%
\definecolor{mycolor5}{rgb}{0.30100,0.74500,0.93300}%
\definecolor{mycolor6}{rgb}{0.85000,0.32500,0.09800}%
\begin{tikzpicture}

\begin{axis}[%
width=4.464in,
height=3.286in,
at={(0.815in,0.76in)},
scale only axis,
xmin=0,
xmax=0.95,
xlabel style={font=\color{white!15!black}},
xlabel={Percentage of adversaries},
ymin=0.2,
ymax=1,
ylabel style={font=\color{white!15!black}},
ylabel={Accuracy},
axis background/.style={fill=white},
axis x line*=bottom,
axis y line*=left,
]

\addplot[area legend, draw=none, fill=mycolor1, fill opacity=0.2, forget plot]
table[row sep=crcr] {%
x	y\\
0	0.729325158567994\\
0.1	0.754549105165592\\
0.2	0.695298804916924\\
0.3	0.393783766902744\\
0.4	0.811678174773686\\
0.5	0.718203864189302\\
0.6	0.709234494146583\\
0.7	0.762923636066109\\
0.8	0.738703565299346\\
0.9	0.338349320252203\\
0.9	0.778317346414463\\
0.8	0.903889027293246\\
0.7	0.920409697267224\\
0.6	0.959284024371936\\
0.5	0.955870209884772\\
0.4	0.917951454855944\\
0.3	0.922882899763923\\
0.2	0.960256750638631\\
0.1	0.951006450389964\\
0	0.957711878469043\\
}--cycle;
\addplot [color=white!55!mycolor1, forget plot]
  table[row sep=crcr]{%
0	0.729325158567994\\
0.1	0.754549105165592\\
0.2	0.695298804916924\\
0.3	0.393783766902744\\
0.4	0.811678174773686\\
0.5	0.718203864189302\\
0.6	0.709234494146583\\
0.7	0.762923636066109\\
0.8	0.738703565299346\\
0.9	0.338349320252203\\
};
\addplot [color=white!55!mycolor1, forget plot]
  table[row sep=crcr]{%
0	0.957711878469043\\
0.1	0.951006450389964\\
0.2	0.960256750638631\\
0.3	0.922882899763923\\
0.4	0.917951454855944\\
0.5	0.955870209884772\\
0.6	0.959284024371936\\
0.7	0.920409697267224\\
0.8	0.903889027293246\\
0.9	0.778317346414463\\
};
\addplot [color=mycolor1, line width=1.0pt, mark size=5.0pt, mark=o, mark options={solid, mycolor1}]
  table[row sep=crcr]{%
0	0.843518518518519\\
0.1	0.852777777777778\\
0.2	0.827777777777778\\
0.3	0.658333333333333\\
0.4	0.864814814814815\\
0.5	0.837037037037037\\
0.6	0.834259259259259\\
0.7	0.841666666666667\\
0.8	0.821296296296296\\
0.9	0.558333333333333\\
};
%\addlegendentry{Alg. 1 - TA + DS}

\addplot[area legend, draw=none, fill=mycolor2, fill opacity=0.2, forget plot]
table[row sep=crcr] {%
x	y\\
0	0.879629629629629\\
0.1	0.873623302795829\\
0.2	0.815023696870508\\
0.3	0.392131916914051\\
0.4	0.601395229991594\\
0.5	0.425050846852369\\
0.6	0.440462155727014\\
0.7	0.484786782263369\\
0.8	0.456702347036258\\
0.9	0.492348057030713\\
0.9	0.798392683710027\\
0.8	0.743297652963742\\
0.7	0.730028032551446\\
0.6	0.722500807235948\\
0.5	0.741615819814297\\
0.4	0.935641807045443\\
0.3	0.902312527530394\\
0.2	0.920161488314677\\
0.1	0.891191512018986\\
0	0.87962962962963\\
}--cycle;
\addplot [color=white!55!mycolor2, forget plot]
  table[row sep=crcr]{%
0	0.879629629629629\\
0.1	0.873623302795829\\
0.2	0.815023696870508\\
0.3	0.392131916914051\\
0.4	0.601395229991594\\
0.5	0.425050846852369\\
0.6	0.440462155727014\\
0.7	0.484786782263369\\
0.8	0.456702347036258\\
0.9	0.492348057030713\\
};
\addplot [color=white!55!mycolor2, forget plot]
  table[row sep=crcr]{%
0	0.87962962962963\\
0.1	0.891191512018986\\
0.2	0.920161488314677\\
0.3	0.902312527530394\\
0.4	0.935641807045443\\
0.5	0.741615819814297\\
0.6	0.722500807235948\\
0.7	0.730028032551446\\
0.8	0.743297652963742\\
0.9	0.798392683710027\\
};
\addplot [color=mycolor2, line width=1.0pt, mark size=8.6pt, mark=diamond, mark options={solid, mycolor2}]
  table[row sep=crcr]{%
0	0.87962962962963\\
0.1	0.882407407407407\\
0.2	0.867592592592593\\
0.3	0.647222222222222\\
0.4	0.768518518518518\\
0.5	0.583333333333333\\
0.6	0.581481481481481\\
0.7	0.607407407407407\\
0.8	0.6\\
0.9	0.64537037037037\\
};
%\addlegendentry{Alg. 1 - H + DS}

\addplot[area legend, draw=none, fill=mycolor3, fill opacity=0.2, forget plot]
table[row sep=crcr] {%
x	y\\
0	0.712962962962963\\
0.1	0.663065962972672\\
0.2	0.612250768377659\\
0.3	0.650654144937021\\
0.4	0.319339680539581\\
0.5	0.446603400195038\\
0.6	0.434494226841197\\
0.7	0.510689640609806\\
0.8	0.486918172571643\\
0.9	0.520836500112852\\
0.9	0.829163499887148\\
0.8	0.772341086687616\\
0.7	0.809680729760564\\
0.6	0.791431699084729\\
0.5	0.818211414619777\\
0.4	0.819549208349308\\
0.3	0.816012521729646\\
0.2	0.772934416807526\\
0.1	0.751748851842143\\
0	0.712962962962963\\
}--cycle;
\addplot [color=white!55!mycolor3, forget plot]
  table[row sep=crcr]{%
0	0.712962962962963\\
0.1	0.663065962972672\\
0.2	0.612250768377659\\
0.3	0.650654144937021\\
0.4	0.319339680539581\\
0.5	0.446603400195038\\
0.6	0.434494226841197\\
0.7	0.510689640609806\\
0.8	0.486918172571643\\
0.9	0.520836500112852\\
};
\addplot [color=white!55!mycolor3, forget plot]
  table[row sep=crcr]{%
0	0.712962962962963\\
0.1	0.751748851842143\\
0.2	0.772934416807526\\
0.3	0.816012521729646\\
0.4	0.819549208349308\\
0.5	0.818211414619777\\
0.6	0.791431699084729\\
0.7	0.809680729760564\\
0.8	0.772341086687616\\
0.9	0.829163499887148\\
};
\addplot [color=mycolor3, dashed, line width=1.0pt, mark size=5.0pt, mark=x, mark options={solid, mycolor3}]
  table[row sep=crcr]{%
0	0.712962962962963\\
0.1	0.707407407407407\\
0.2	0.692592592592593\\
0.3	0.733333333333333\\
0.4	0.569444444444444\\
0.5	0.632407407407407\\
0.6	0.612962962962963\\
0.7	0.660185185185185\\
0.8	0.62962962962963\\
0.9	0.675\\
};
%\addlegendentry{MMSR}

\addplot[area legend, draw=none, fill=mycolor4, fill opacity=0.2, forget plot]
table[row sep=crcr] {%
x	y\\
0	0.879629629629629\\
0.1	0.876144552352501\\
0.2	0.871318354663516\\
0.3	0.413142006404829\\
0.4	0.474412444747309\\
0.5	0.529155644155534\\
0.6	0.462265479625747\\
0.7	0.492855165832658\\
0.8	0.456196203417552\\
0.9	0.488942788095494\\
0.9	0.799946100793395\\
0.8	0.7456556484343\\
0.7	0.753441130463638\\
0.6	0.758104890744623\\
0.5	0.820844355844466\\
0.4	0.775587555252691\\
0.3	0.953524660261838\\
0.2	0.910163126817966\\
0.1	0.909040632832684\\
0	0.87962962962963\\
}--cycle;
\addplot [color=white!55!mycolor4, forget plot]
  table[row sep=crcr]{%
0	0.879629629629629\\
0.1	0.876144552352501\\
0.2	0.871318354663516\\
0.3	0.413142006404829\\
0.4	0.474412444747309\\
0.5	0.529155644155534\\
0.6	0.462265479625747\\
0.7	0.492855165832658\\
0.8	0.456196203417552\\
0.9	0.488942788095494\\
};
\addplot [color=white!55!mycolor4, forget plot]
  table[row sep=crcr]{%
0	0.87962962962963\\
0.1	0.909040632832684\\
0.2	0.910163126817966\\
0.3	0.953524660261838\\
0.4	0.775587555252691\\
0.5	0.820844355844466\\
0.6	0.758104890744623\\
0.7	0.753441130463638\\
0.8	0.7456556484343\\
0.9	0.799946100793395\\
};
\addplot [color=mycolor4, dashdotted, line width=1.0pt, mark size=3.5pt, mark=square, mark options={solid, mycolor4}]
  table[row sep=crcr]{%
0	0.87962962962963\\
0.1	0.892592592592593\\
0.2	0.890740740740741\\
0.3	0.683333333333333\\
0.4	0.625\\
0.5	0.675\\
0.6	0.610185185185185\\
0.7	0.623148148148148\\
0.8	0.600925925925926\\
0.9	0.644444444444444\\
};
%\addlegendentry{DS}

\addplot[area legend, draw=none, fill=mycolor5, fill opacity=0.2, forget plot]
table[row sep=crcr] {%
x	y\\
0	0.759259259259259\\
0.1	0.736429711870581\\
0.2	0.704025328500936\\
0.3	0.584031189738983\\
0.4	0.46517178107851\\
0.5	0.507187252116292\\
0.6	0.462986729577163\\
0.7	0.519776855845784\\
0.8	0.486918172571643\\
0.9	0.518321560076569\\
0.9	0.826122884367876\\
0.8	0.772341086687616\\
0.7	0.80429721822829\\
0.6	0.779605863015429\\
0.5	0.802072007142967\\
0.4	0.729272663365935\\
0.3	0.697450291742498\\
0.2	0.770048745573138\\
0.1	0.774681399240531\\
0	0.75925925925926\\
}--cycle;
\addplot [color=white!55!mycolor5, forget plot]
  table[row sep=crcr]{%
0	0.759259259259259\\
0.1	0.736429711870581\\
0.2	0.704025328500936\\
0.3	0.584031189738983\\
0.4	0.46517178107851\\
0.5	0.507187252116292\\
0.6	0.462986729577163\\
0.7	0.519776855845784\\
0.8	0.486918172571643\\
0.9	0.518321560076569\\
};
\addplot [color=white!55!mycolor5, forget plot]
  table[row sep=crcr]{%
0	0.75925925925926\\
0.1	0.774681399240531\\
0.2	0.770048745573138\\
0.3	0.697450291742498\\
0.4	0.729272663365935\\
0.5	0.802072007142967\\
0.6	0.779605863015429\\
0.7	0.80429721822829\\
0.8	0.772341086687616\\
0.9	0.826122884367876\\
};
\addplot [color=mycolor5, dashdotted, line width=1.0pt, mark size=5.0pt, mark=+, mark options={solid, mycolor5}]
  table[row sep=crcr]{%
0	0.759259259259259\\
0.1	0.755555555555556\\
0.2	0.737037037037037\\
0.3	0.640740740740741\\
0.4	0.597222222222222\\
0.5	0.65462962962963\\
0.6	0.621296296296296\\
0.7	0.662037037037037\\
0.8	0.62962962962963\\
0.9	0.672222222222222\\
};
%\addlegendentry{MV}

\addplot[area legend, draw=none, fill=mycolor6, fill opacity=0.2, forget plot]
table[row sep=crcr] {%
x	y\\
0	0.759259259259259\\
0.1	0.75157554901881\\
0.2	0.708930494163845\\
0.3	0.667976433438783\\
0.4	0.45758755809283\\
0.5	0.53307979246123\\
0.6	0.481989365267152\\
0.7	0.52193543852845\\
0.8	0.488885863019485\\
0.9	0.520836500112852\\
0.9	0.829163499887148\\
0.8	0.772225248091626\\
0.7	0.809546042953031\\
0.6	0.792084708806922\\
0.5	0.811364651983215\\
0.4	0.753523553018281\\
0.3	0.713505048042698\\
0.2	0.779958394725043\\
0.1	0.778054080610819\\
0	0.75925925925926\\
}--cycle;
\addplot [color=white!55!mycolor6, forget plot]
  table[row sep=crcr]{%
0	0.759259259259259\\
0.1	0.75157554901881\\
0.2	0.708930494163845\\
0.3	0.667976433438783\\
0.4	0.45758755809283\\
0.5	0.53307979246123\\
0.6	0.481989365267152\\
0.7	0.52193543852845\\
0.8	0.488885863019485\\
0.9	0.520836500112852\\
};
\addplot [color=white!55!mycolor6, forget plot]
  table[row sep=crcr]{%
0	0.75925925925926\\
0.1	0.778054080610819\\
0.2	0.779958394725043\\
0.3	0.713505048042698\\
0.4	0.753523553018281\\
0.5	0.811364651983215\\
0.6	0.792084708806922\\
0.7	0.809546042953031\\
0.8	0.772225248091626\\
0.9	0.829163499887148\\
};
\addplot [color=mycolor6, dashdotted, line width=1.0pt, mark size=3.3pt, mark=triangle, mark options={solid, mycolor6}]
  table[row sep=crcr]{%
0	0.759259259259259\\
0.1	0.764814814814815\\
0.2	0.744444444444444\\
0.3	0.690740740740741\\
0.4	0.605555555555556\\
0.5	0.672222222222222\\
0.6	0.637037037037037\\
0.7	0.665740740740741\\
0.8	0.630555555555556\\
0.9	0.675\\
};
%\addlegendentry{PGD}

\end{axis}
\end{tikzpicture}%

%% file: fig/tikz/real_data/per_adv_DS_rte_shaded_acc.tikz
\definecolor{mycolor1}{rgb}{0.00000,0.44700,0.74100}%
\definecolor{mycolor2}{rgb}{0.49400,0.18400,0.55600}%
\definecolor{mycolor3}{rgb}{0.63500,0.07800,0.18400}%
\definecolor{mycolor4}{rgb}{0.92900,0.69400,0.12500}%
\definecolor{mycolor5}{rgb}{0.30100,0.74500,0.93300}%
\definecolor{mycolor6}{rgb}{0.85000,0.32500,0.09800}%
\begin{tikzpicture}

\begin{axis}[%
width=4.464in,
height=3.286in,
at={(0.815in,0.76in)},
scale only axis,
xmin=0,
xmax=0.95,
xlabel style={font=\color{white!15!black}},
xlabel={Percentage of adversaries},
ymin=0.369226535512713,
ymax=1,
ylabel style={font=\color{white!15!black}},
ylabel={Accuracy},
axis background/.style={fill=white},
axis x line*=bottom,
axis y line*=left,
]

\addplot[area legend, draw=none, fill=mycolor1, fill opacity=0.2, forget plot]
table[row sep=crcr] {%
x	y\\
0	0.916104502775632\\
0.1	0.877085481156726\\
0.2	0.369384242851758\\
0.3	0.683865031217537\\
0.4	0.720539367172324\\
0.5	0.799658511257126\\
0.6	0.795940673709729\\
0.7	0.798024844751892\\
0.8	0.562725763559492\\
0.9	0.539662007488657\\
0.9	0.689087992511343\\
0.8	0.835024236440508\\
0.7	0.881225155248108\\
0.6	0.854059326290271\\
0.5	0.919091488742874\\
0.4	0.972710632827676\\
0.3	0.965134968782464\\
0.2	0.716865757148242\\
0.1	0.957914518843274\\
0	0.917395497224368\\
}--cycle;
\addplot [color=white!55!mycolor1, forget plot]
  table[row sep=crcr]{%
0	0.916104502775632\\
0.1	0.877085481156726\\
0.2	0.369384242851758\\
0.3	0.683865031217537\\
0.4	0.720539367172324\\
0.5	0.799658511257126\\
0.6	0.795940673709729\\
0.7	0.798024844751892\\
0.8	0.562725763559492\\
0.9	0.539662007488657\\
};
\addplot [color=white!55!mycolor1, forget plot]
  table[row sep=crcr]{%
0	0.917395497224368\\
0.1	0.957914518843274\\
0.2	0.716865757148242\\
0.3	0.965134968782464\\
0.4	0.972710632827676\\
0.5	0.919091488742874\\
0.6	0.854059326290271\\
0.7	0.881225155248108\\
0.8	0.835024236440508\\
0.9	0.689087992511343\\
};
\addplot [color=mycolor1, line width=1.0pt, mark size=5.0pt, mark=o, mark options={solid, mycolor1}]
  table[row sep=crcr]{%
0	0.91675\\
0.1	0.9175\\
0.2	0.543125\\
0.3	0.8245\\
0.4	0.846625\\
0.5	0.859375\\
0.6	0.825\\
0.7	0.839625\\
0.8	0.698875\\
0.9	0.614375\\
};
%\addlegendentry{Alg. 1 - TA + DS}

\addplot[area legend, draw=none, fill=mycolor2, fill opacity=0.2, forget plot]
table[row sep=crcr] {%
x	y\\
0	0.9175\\
0.1	0.919379961724244\\
0.2	0.369384242851758\\
0.3	0.683865031217537\\
0.4	0.720539367172324\\
0.5	0.449370877302719\\
0.6	0.456419189130991\\
0.7	0.441245883617954\\
0.8	0.489986118818599\\
0.9	0.439336270247892\\
0.9	0.590663729752108\\
0.8	0.515013881181401\\
0.7	0.593504116382046\\
0.6	0.605080810869009\\
0.5	0.670879122697281\\
0.4	0.972710632827676\\
0.3	0.965134968782464\\
0.2	0.716865757148242\\
0.1	0.936870038275756\\
0	0.9175\\
}--cycle;
\addplot [color=white!55!mycolor2, forget plot]
  table[row sep=crcr]{%
0	0.9175\\
0.1	0.919379961724244\\
0.2	0.369384242851758\\
0.3	0.683865031217537\\
0.4	0.720539367172324\\
0.5	0.449370877302719\\
0.6	0.456419189130991\\
0.7	0.441245883617954\\
0.8	0.489986118818599\\
0.9	0.439336270247892\\
};
\addplot [color=white!55!mycolor2, forget plot]
  table[row sep=crcr]{%
0	0.9175\\
0.1	0.936870038275756\\
0.2	0.716865757148242\\
0.3	0.965134968782464\\
0.4	0.972710632827676\\
0.5	0.670879122697281\\
0.6	0.605080810869009\\
0.7	0.593504116382046\\
0.8	0.515013881181401\\
0.9	0.590663729752108\\
};
\addplot [color=mycolor2, line width=1.0pt, mark size=8.6pt, mark=diamond, mark options={solid, mycolor2}]
  table[row sep=crcr]{%
0	0.9175\\
0.1	0.928125\\
0.2	0.543125\\
0.3	0.8245\\
0.4	0.846625\\
0.5	0.560125\\
0.6	0.53075\\
0.7	0.517375\\
0.8	0.5025\\
0.9	0.515\\
};
%\addlegendentry{Alg. 1 - H + DS}

\addplot[area legend, draw=none, fill=mycolor3, fill opacity=0.2, forget plot]
table[row sep=crcr] {%
x	y\\
0	0.8975\\
0.1	0.730485198419553\\
0.2	0.709407271630605\\
0.3	0.567664175775218\\
0.4	0.539274111034302\\
0.5	0.502640792249705\\
0.6	0.55049427779834\\
0.7	0.488615780976975\\
0.8	0.46856244852278\\
0.9	0.470145117027724\\
0.9	0.721604882972276\\
0.8	0.72993755147722\\
0.7	0.749634219023026\\
0.6	0.79200572220166\\
0.5	0.763609207750295\\
0.4	0.745975888965698\\
0.3	0.690335824224782\\
0.2	0.761342728369395\\
0.1	0.852014801580447\\
0	0.8975\\
}--cycle;
\addplot [color=white!55!mycolor3, forget plot]
  table[row sep=crcr]{%
0	0.8975\\
0.1	0.730485198419553\\
0.2	0.709407271630605\\
0.3	0.567664175775218\\
0.4	0.539274111034302\\
0.5	0.502640792249705\\
0.6	0.55049427779834\\
0.7	0.488615780976975\\
0.8	0.46856244852278\\
0.9	0.470145117027724\\
};
\addplot [color=white!55!mycolor3, forget plot]
  table[row sep=crcr]{%
0	0.8975\\
0.1	0.852014801580447\\
0.2	0.761342728369395\\
0.3	0.690335824224782\\
0.4	0.745975888965698\\
0.5	0.763609207750295\\
0.6	0.79200572220166\\
0.7	0.749634219023026\\
0.8	0.72993755147722\\
0.9	0.721604882972276\\
};
\addplot [color=mycolor3, dashed, line width=1.0pt, mark size=5.0pt, mark=x, mark options={solid, mycolor3}]
  table[row sep=crcr]{%
0	0.8975\\
0.1	0.79125\\
0.2	0.735375\\
0.3	0.629\\
0.4	0.642625\\
0.5	0.633125\\
0.6	0.67125\\
0.7	0.619125\\
0.8	0.59925\\
0.9	0.595875\\
};
%\addlegendentry{MMSR}

\addplot[area legend, draw=none, fill=mycolor4, fill opacity=0.2, forget plot]
table[row sep=crcr] {%
x	y\\
0	0.92875\\
0.1	0.929332446908819\\
0.2	0.369226535512713\\
0.3	0.435816309836463\\
0.4	0.482536123336263\\
0.5	0.514090760198723\\
0.6	0.484649073589235\\
0.7	0.445138775142053\\
0.8	0.4517760487418\\
0.9	0.434245698000563\\
0.9	0.644504301999437\\
0.8	0.6089739512582\\
0.7	0.641861224857947\\
0.6	0.721600926410764\\
0.5	0.746909239801276\\
0.4	0.731213876663737\\
0.3	0.664683690163537\\
0.2	0.718773464487287\\
0.1	0.937167553091181\\
0	0.92875\\
}--cycle;
\addplot [color=white!55!mycolor4, forget plot]
  table[row sep=crcr]{%
0	0.92875\\
0.1	0.929332446908819\\
0.2	0.369226535512713\\
0.3	0.435816309836463\\
0.4	0.482536123336263\\
0.5	0.514090760198723\\
0.6	0.484649073589235\\
0.7	0.445138775142053\\
0.8	0.4517760487418\\
0.9	0.434245698000563\\
};
\addplot [color=white!55!mycolor4, forget plot]
  table[row sep=crcr]{%
0	0.92875\\
0.1	0.937167553091181\\
0.2	0.718773464487287\\
0.3	0.664683690163537\\
0.4	0.731213876663737\\
0.5	0.746909239801276\\
0.6	0.721600926410764\\
0.7	0.641861224857947\\
0.8	0.6089739512582\\
0.9	0.644504301999437\\
};
\addplot [color=mycolor4, dashdotted, line width=1.0pt, mark size=3.5pt, mark=square, mark options={solid, mycolor4}]
  table[row sep=crcr]{%
0	0.92875\\
0.1	0.93325\\
0.2	0.544\\
0.3	0.55025\\
0.4	0.606875\\
0.5	0.6305\\
0.6	0.603125\\
0.7	0.5435\\
0.8	0.530375\\
0.9	0.539375\\
};
%\addlegendentry{DS}

\addplot[area legend, draw=none, fill=mycolor5, fill opacity=0.2, forget plot]
table[row sep=crcr] {%
x	y\\
0	0.91875\\
0.1	0.738093165445718\\
0.2	0.470558498610585\\
0.3	0.441672978194726\\
0.4	0.48400103290078\\
0.5	0.514949447197067\\
0.6	0.551062801080698\\
0.7	0.490437434676009\\
0.8	0.46827210146994\\
0.9	0.470031025133302\\
0.9	0.720468974866697\\
0.8	0.72872789853006\\
0.7	0.749562565323991\\
0.6	0.794687198919302\\
0.5	0.762300552802933\\
0.4	0.75124896709922\\
0.3	0.679327021805274\\
0.2	0.726441501389415\\
0.1	0.886406834554282\\
0	0.91875\\
}--cycle;
\addplot [color=white!55!mycolor5, forget plot]
  table[row sep=crcr]{%
0	0.91875\\
0.1	0.738093165445718\\
0.2	0.470558498610585\\
0.3	0.441672978194726\\
0.4	0.48400103290078\\
0.5	0.514949447197067\\
0.6	0.551062801080698\\
0.7	0.490437434676009\\
0.8	0.46827210146994\\
0.9	0.470031025133302\\
};
\addplot [color=white!55!mycolor5, forget plot]
  table[row sep=crcr]{%
0	0.91875\\
0.1	0.886406834554282\\
0.2	0.726441501389415\\
0.3	0.679327021805274\\
0.4	0.75124896709922\\
0.5	0.762300552802933\\
0.6	0.794687198919302\\
0.7	0.749562565323991\\
0.8	0.72872789853006\\
0.9	0.720468974866697\\
};
\addplot [color=mycolor5, dashdotted, line width=1.0pt, mark size=5.0pt, mark=+, mark options={solid, mycolor5}]
  table[row sep=crcr]{%
0	0.91875\\
0.1	0.81225\\
0.2	0.5985\\
0.3	0.5605\\
0.4	0.617625\\
0.5	0.638625\\
0.6	0.672875\\
0.7	0.62\\
0.8	0.5985\\
0.9	0.59525\\
};
%\addlegendentry{MV}

\addplot[area legend, draw=none, fill=mycolor6, fill opacity=0.2, forget plot]
table[row sep=crcr] {%
x	y\\
0	0.84125\\
0.1	0.537905402543867\\
0.2	0.430580760470004\\
0.3	0.437826473235233\\
0.4	0.482191800170453\\
0.5	0.513268065213613\\
0.6	0.552839853318489\\
0.7	0.490946434867637\\
0.8	0.468609129849619\\
0.9	0.470378213142253\\
0.9	0.721621786857747\\
0.8	0.729640870150381\\
0.7	0.749803565132363\\
0.6	0.795410146681511\\
0.5	0.762731934786387\\
0.4	0.749558199829547\\
0.3	0.671673526764767\\
0.2	0.701169239529996\\
0.1	0.772344597456132\\
0	0.84125\\
}--cycle;
\addplot [color=white!55!mycolor6, forget plot]
  table[row sep=crcr]{%
0	0.84125\\
0.1	0.537905402543867\\
0.2	0.430580760470004\\
0.3	0.437826473235233\\
0.4	0.482191800170453\\
0.5	0.513268065213613\\
0.6	0.552839853318489\\
0.7	0.490946434867637\\
0.8	0.468609129849619\\
0.9	0.470378213142253\\
};
\addplot [color=white!55!mycolor6, forget plot]
  table[row sep=crcr]{%
0	0.84125\\
0.1	0.772344597456132\\
0.2	0.701169239529996\\
0.3	0.671673526764767\\
0.4	0.749558199829547\\
0.5	0.762731934786387\\
0.6	0.795410146681511\\
0.7	0.749803565132363\\
0.8	0.729640870150381\\
0.9	0.721621786857747\\
};
\addplot [color=mycolor6, dashdotted, line width=1.0pt, mark size=3.3pt, mark=triangle, mark options={solid, mycolor6}]
  table[row sep=crcr]{%
0	0.84125\\
0.1	0.655125\\
0.2	0.565875\\
0.3	0.55475\\
0.4	0.615875\\
0.5	0.638\\
0.6	0.674125\\
0.7	0.620375\\
0.8	0.599125\\
0.9	0.596\\
};
%\addlegendentry{PGD}

\end{axis}
\end{tikzpicture}%

%% file: fig/tikz/real_data/per_adv_DS_sen_polarity_shaded_acc.tikz
\definecolor{mycolor1}{rgb}{0.00000,0.44700,0.74100}%
\definecolor{mycolor2}{rgb}{0.49400,0.18400,0.55600}%
\definecolor{mycolor3}{rgb}{0.63500,0.07800,0.18400}%
\definecolor{mycolor4}{rgb}{0.92900,0.69400,0.12500}%
\definecolor{mycolor5}{rgb}{0.30100,0.74500,0.93300}%
\definecolor{mycolor6}{rgb}{0.85000,0.32500,0.09800}%
\begin{tikzpicture}

\begin{axis}[%
width=4.464in,
height=3.286in,
at={(0.815in,0.76in)},
scale only axis,
xmin=0,
xmax=0.95,
xlabel style={font=\color{white!15!black}},
xlabel={Percentage of adversaries},
ymin=0.2,
ymax=1,
ylabel style={font=\color{white!15!black}},
ylabel={Accuracy},
axis background/.style={fill=white},
axis x line*=bottom,
axis y line*=left,
]

\addplot[area legend, draw=none, fill=mycolor1, fill opacity=0.2, forget plot]
table[row sep=crcr] {%
x	y\\
0	0.913318806058704\\
0.1	0.669691020278959\\
0.2	0.703884554822755\\
0.3	0.857113662810811\\
0.4	0.876027056173997\\
0.5	0.842681202516557\\
0.6	0.824621615556322\\
0.7	0.792484514386858\\
0.8	0.583814244035022\\
0.9	0.493413235553206\\
0.9	0.696984844062717\\
0.8	0.783139146643114\\
0.7	0.836241230762172\\
0.6	0.87935918060291\\
0.5	0.887064746673281\\
0.4	0.896687486734585\\
0.3	0.912160191960143\\
0.2	0.943084839056021\\
0.1	0.989680854095915\\
0	0.917367331168741\\
}--cycle;
\addplot [color=white!55!mycolor1, forget plot]
  table[row sep=crcr]{%
0	0.913318806058704\\
0.1	0.669691020278959\\
0.2	0.703884554822755\\
0.3	0.857113662810811\\
0.4	0.876027056173997\\
0.5	0.842681202516557\\
0.6	0.824621615556322\\
0.7	0.792484514386858\\
0.8	0.583814244035022\\
0.9	0.493413235553206\\
};
\addplot [color=white!55!mycolor1, forget plot]
  table[row sep=crcr]{%
0	0.917367331168741\\
0.1	0.989680854095915\\
0.2	0.943084839056021\\
0.3	0.912160191960143\\
0.4	0.896687486734585\\
0.5	0.887064746673281\\
0.6	0.87935918060291\\
0.7	0.836241230762172\\
0.8	0.783139146643114\\
0.9	0.696984844062717\\
};
\addplot [color=mycolor1, line width=1.0pt, mark size=5.0pt, mark=o, mark options={solid, mycolor1}]
  table[row sep=crcr]{%
0	0.915343068613723\\
0.1	0.829685937187437\\
0.2	0.823484696939388\\
0.3	0.884636927385477\\
0.4	0.886357271454291\\
0.5	0.864872974594919\\
0.6	0.851990398079616\\
0.7	0.814362872574515\\
0.8	0.683476695339068\\
0.9	0.595199039807962\\
};
%\addlegendentry{Alg. 1 - TA + DS}

\addplot[area legend, draw=none, fill=mycolor2, fill opacity=0.2, forget plot]
table[row sep=crcr] {%
x	y\\
0	0.915983196639328\\
0.1	0.808195398832905\\
0.2	0.809278716454325\\
0.3	0.857113662810811\\
0.4	0.876027056173997\\
0.5	0.490829729244642\\
0.6	0.493739296530461\\
0.7	0.448368775959439\\
0.8	0.448895135574315\\
0.9	0.494067742647286\\
0.9	0.503771825266296\\
0.8	0.595273698192439\\
0.7	0.640929083612476\\
0.6	0.503380127354316\\
0.5	0.50692982266574\\
0.4	0.896687486734585\\
0.3	0.912160191960143\\
0.2	0.912425624413849\\
0.1	0.950076255497961\\
0	0.915983196639328\\
}--cycle;
\addplot [color=white!55!mycolor2, forget plot]
  table[row sep=crcr]{%
0	0.915983196639328\\
0.1	0.808195398832905\\
0.2	0.809278716454325\\
0.3	0.857113662810811\\
0.4	0.876027056173997\\
0.5	0.490829729244642\\
0.6	0.493739296530461\\
0.7	0.448368775959439\\
0.8	0.448895135574315\\
0.9	0.494067742647286\\
};
\addplot [color=white!55!mycolor2, forget plot]
  table[row sep=crcr]{%
0	0.915983196639328\\
0.1	0.950076255497961\\
0.2	0.912425624413849\\
0.3	0.912160191960143\\
0.4	0.896687486734585\\
0.5	0.50692982266574\\
0.6	0.503380127354316\\
0.7	0.640929083612476\\
0.8	0.595273698192439\\
0.9	0.503771825266296\\
};
\addplot [color=mycolor2, line width=1.0pt, mark size=8.6pt, mark=diamond, mark options={solid, mycolor2}]
  table[row sep=crcr]{%
0	0.915983196639328\\
0.1	0.879135827165433\\
0.2	0.860852170434087\\
0.3	0.884636927385477\\
0.4	0.886357271454291\\
0.5	0.498879775955191\\
0.6	0.498559711942388\\
0.7	0.544648929785957\\
0.8	0.522084416883377\\
0.9	0.498919783956791\\
};
%\addlegendentry{Alg. 1 - H + DS}

\addplot[area legend, draw=none, fill=mycolor3, fill opacity=0.2, forget plot]
table[row sep=crcr] {%
x	y\\
0	0.901180236047209\\
0.1	0.855556365355966\\
0.2	0.690066601112364\\
0.3	0.619444125733884\\
0.4	0.539575138551142\\
0.5	0.505540470908832\\
0.6	0.464536727905403\\
0.7	0.551932511892391\\
0.8	0.469873907598075\\
0.9	0.45371120932817\\
0.9	0.693638260565809\\
0.8	0.727925652313908\\
0.7	0.792896453900768\\
0.6	0.724221023644907\\
0.5	0.774555548295009\\
0.4	0.767566289734515\\
0.3	0.785316826456554\\
0.2	0.740739560119882\\
0.1	0.89143303252361\\
0	0.90118023604721\\
}--cycle;
\addplot [color=white!55!mycolor3, forget plot]
  table[row sep=crcr]{%
0	0.901180236047209\\
0.1	0.855556365355966\\
0.2	0.690066601112364\\
0.3	0.619444125733884\\
0.4	0.539575138551142\\
0.5	0.505540470908832\\
0.6	0.464536727905403\\
0.7	0.551932511892391\\
0.8	0.469873907598075\\
0.9	0.45371120932817\\
};
\addplot [color=white!55!mycolor3, forget plot]
  table[row sep=crcr]{%
0	0.90118023604721\\
0.1	0.89143303252361\\
0.2	0.740739560119882\\
0.3	0.785316826456554\\
0.4	0.767566289734515\\
0.5	0.774555548295009\\
0.6	0.724221023644907\\
0.7	0.792896453900768\\
0.8	0.727925652313908\\
0.9	0.693638260565809\\
};
\addplot [color=mycolor3, dashed, line width=1.0pt, mark size=5.0pt, mark=x, mark options={solid, mycolor3}]
  table[row sep=crcr]{%
0	0.90118023604721\\
0.1	0.873494698939788\\
0.2	0.715403080616123\\
0.3	0.702380476095219\\
0.4	0.653570714142828\\
0.5	0.64004800960192\\
0.6	0.594378875775155\\
0.7	0.672414482896579\\
0.8	0.598899779955991\\
0.9	0.573674734946989\\
};
%\addlegendentry{MMSR}

\addplot[area legend, draw=none, fill=mycolor4, fill opacity=0.2, forget plot]
table[row sep=crcr] {%
x	y\\
0	0.914782956591318\\
0.1	0.254706107818396\\
0.2	0.456140684688355\\
0.3	0.465067577032413\\
0.4	0.447715076321682\\
0.5	0.449726938915133\\
0.6	0.49370261021742\\
0.7	0.473383617315373\\
0.8	0.44727880890292\\
0.9	0.494179919668155\\
0.9	0.50385968825343\\
0.8	0.598450336926246\\
0.7	0.708772813970884\\
0.6	0.503336797664156\\
0.5	0.592561518776405\\
0.4	0.598174101513885\\
0.3	0.631831802843562\\
0.2	0.702170977644112\\
0.1	0.715807995002168\\
0	0.914782956591318\\
}--cycle;
\addplot [color=white!55!mycolor4, forget plot]
  table[row sep=crcr]{%
0	0.914782956591318\\
0.1	0.254706107818396\\
0.2	0.456140684688355\\
0.3	0.465067577032413\\
0.4	0.447715076321682\\
0.5	0.449726938915133\\
0.6	0.49370261021742\\
0.7	0.473383617315373\\
0.8	0.44727880890292\\
0.9	0.494179919668155\\
};
\addplot [color=white!55!mycolor4, forget plot]
  table[row sep=crcr]{%
0	0.914782956591318\\
0.1	0.715807995002168\\
0.2	0.702170977644112\\
0.3	0.631831802843562\\
0.4	0.598174101513885\\
0.5	0.592561518776405\\
0.6	0.503336797664156\\
0.7	0.708772813970884\\
0.8	0.598450336926246\\
0.9	0.50385968825343\\
};
\addplot [color=mycolor4, dashdotted, line width=1.0pt, mark size=3.5pt, mark=square, mark options={solid, mycolor4}]
  table[row sep=crcr]{%
0	0.914782956591318\\
0.1	0.485257051410282\\
0.2	0.579155831166233\\
0.3	0.548449689937988\\
0.4	0.522944588917784\\
0.5	0.521144228845769\\
0.6	0.498519703940788\\
0.7	0.591078215643129\\
0.8	0.522864572914583\\
0.9	0.499019803960792\\
};
%\addlegendentry{DS}

\addplot[area legend, draw=none, fill=mycolor5, fill opacity=0.2, forget plot]
table[row sep=crcr] {%
x	y\\
0	0.889577915583117\\
0.1	0.505697970282399\\
0.2	0.477308124402024\\
0.3	0.584855564809822\\
0.4	0.51487454559462\\
0.5	0.514678736891079\\
0.6	0.467285081340459\\
0.7	0.551643719475319\\
0.8	0.470043466504988\\
0.9	0.453755255855897\\
0.9	0.693434182031681\\
0.8	0.727796101408595\\
0.7	0.792505110290634\\
0.6	0.724433262328275\\
0.5	0.773418882632825\\
0.4	0.770502529820463\\
0.3	0.795980602423625\\
0.2	0.737774892201297\\
0.1	0.696022373786415\\
0	0.889577915583117\\
}--cycle;
\addplot [color=white!55!mycolor5, forget plot]
  table[row sep=crcr]{%
0	0.889577915583117\\
0.1	0.505697970282399\\
0.2	0.477308124402024\\
0.3	0.584855564809822\\
0.4	0.51487454559462\\
0.5	0.514678736891079\\
0.6	0.467285081340459\\
0.7	0.551643719475319\\
0.8	0.470043466504988\\
0.9	0.453755255855897\\
};
\addplot [color=white!55!mycolor5, forget plot]
  table[row sep=crcr]{%
0	0.889577915583117\\
0.1	0.696022373786415\\
0.2	0.737774892201297\\
0.3	0.795980602423625\\
0.4	0.770502529820463\\
0.5	0.773418882632825\\
0.6	0.724433262328275\\
0.7	0.792505110290634\\
0.8	0.727796101408595\\
0.9	0.693434182031681\\
};
\addplot [color=mycolor5, dashdotted, line width=1.0pt, mark size=5.0pt, mark=+, mark options={solid, mycolor5}]
  table[row sep=crcr]{%
0	0.889577915583117\\
0.1	0.600860172034407\\
0.2	0.60754150830166\\
0.3	0.690418083616723\\
0.4	0.642688537707542\\
0.5	0.644048809761952\\
0.6	0.595859171834367\\
0.7	0.672074414882977\\
0.8	0.598919783956791\\
0.9	0.573594718943789\\
};
%\addlegendentry{MV}

\addplot[area legend, draw=none, fill=mycolor6, fill opacity=0.2, forget plot]
table[row sep=crcr] {%
x	y\\
0	0.888377675535107\\
0.1	0.447633106035852\\
0.2	0.47170803163749\\
0.3	0.583595544680015\\
0.4	0.514161590024414\\
0.5	0.514507113146013\\
0.6	0.467174683777854\\
0.7	0.552101293765628\\
0.8	0.470203685556238\\
0.9	0.453880214370896\\
0.9	0.693669295531085\\
0.8	0.727995954371748\\
0.7	0.792927712035533\\
0.6	0.724743699898882\\
0.5	0.773630514379493\\
0.4	0.770335309355462\\
0.3	0.795600294487818\\
0.2	0.735413392647367\\
0.1	0.693355091603676\\
0	0.888377675535107\\
}--cycle;
\addplot [color=white!55!mycolor6, forget plot]
  table[row sep=crcr]{%
0	0.888377675535107\\
0.1	0.447633106035852\\
0.2	0.47170803163749\\
0.3	0.583595544680015\\
0.4	0.514161590024414\\
0.5	0.514507113146013\\
0.6	0.467174683777854\\
0.7	0.552101293765628\\
0.8	0.470203685556238\\
0.9	0.453880214370896\\
};
\addplot [color=white!55!mycolor6, forget plot]
  table[row sep=crcr]{%
0	0.888377675535107\\
0.1	0.693355091603676\\
0.2	0.735413392647367\\
0.3	0.795600294487818\\
0.4	0.770335309355462\\
0.5	0.773630514379493\\
0.6	0.724743699898882\\
0.7	0.792927712035533\\
0.8	0.727995954371748\\
0.9	0.693669295531085\\
};
\addplot [color=mycolor6, dashdotted, line width=1.0pt, mark size=3.3pt, mark=triangle, mark options={solid, mycolor6}]
  table[row sep=crcr]{%
0	0.888377675535107\\
0.1	0.570494098819764\\
0.2	0.603560712142428\\
0.3	0.689597919583917\\
0.4	0.642248449689938\\
0.5	0.644068813762753\\
0.6	0.595959191838368\\
0.7	0.67251450290058\\
0.8	0.599099819963993\\
0.9	0.57377475495099\\
};
%\addlegendentry{PGD}

\end{axis}
\end{tikzpicture}%

%% file: fig/tikz/real_data/per_adv_DS_dog_shaded_acc.tikz
\definecolor{mycolor1}{rgb}{0.00000,0.44700,0.74100}%
\definecolor{mycolor2}{rgb}{0.49400,0.18400,0.55600}%
\definecolor{mycolor3}{rgb}{0.63500,0.07800,0.18400}%
\definecolor{mycolor4}{rgb}{0.92900,0.69400,0.12500}%
\definecolor{mycolor5}{rgb}{0.30100,0.74500,0.93300}%
\definecolor{mycolor6}{rgb}{0.85000,0.32500,0.09800}%
\begin{tikzpicture}

\begin{axis}[%
width=4.464in,
height=3.286in,
at={(0.815in,0.76in)},
scale only axis,
xmin=0,
xmax=0.95,
xlabel style={font=\color{white!15!black}},
xlabel={Percentage of adversaries},
ymin=0,
ymax=1,
ylabel style={font=\color{white!15!black}},
ylabel={Accuracy},
axis background/.style={fill=white},
axis x line*=bottom,
axis y line*=left,
]

\addplot[area legend, draw=none, fill=mycolor1, fill opacity=0.2, forget plot]
table[row sep=crcr] {%
x	y\\
0	0.837670384138786\\
0.1	0.835967687554917\\
0.2	0.555032697190026\\
0.3	0.812786802317249\\
0.4	0.823413810725704\\
0.5	0.817939164156644\\
0.6	0.791250278926271\\
0.7	0.798266620607193\\
0.8	0.732900075784129\\
0.9	0.526730554682607\\
0.9	0.877482580385547\\
0.8	0.810098685058498\\
0.7	0.840890752379176\\
0.6	0.844933116364931\\
0.5	0.834601108457979\\
0.4	0.845483339212339\\
0.3	0.838266481449789\\
0.2	0.880407203677384\\
0.1	0.87704346486144\\
0	0.837670384138786\\
}--cycle;
\addplot [color=white!55!mycolor1, forget plot]
  table[row sep=crcr]{%
0	0.837670384138786\\
0.1	0.835967687554917\\
0.2	0.555032697190026\\
0.3	0.812786802317249\\
0.4	0.823413810725704\\
0.5	0.817939164156644\\
0.6	0.791250278926271\\
0.7	0.798266620607193\\
0.8	0.732900075784129\\
0.9	0.526730554682607\\
};
\addplot [color=white!55!mycolor1, forget plot]
  table[row sep=crcr]{%
0	0.837670384138786\\
0.1	0.87704346486144\\
0.2	0.880407203677384\\
0.3	0.838266481449789\\
0.4	0.845483339212339\\
0.5	0.834601108457979\\
0.6	0.844933116364931\\
0.7	0.840890752379176\\
0.8	0.810098685058498\\
0.9	0.877482580385547\\
};
\addplot [color=mycolor1, line width=1.0pt, mark size=5.0pt, mark=o, mark options={solid, mycolor1}]
  table[row sep=crcr]{%
0	0.837670384138786\\
0.1	0.856505576208178\\
0.2	0.717719950433705\\
0.3	0.825526641883519\\
0.4	0.834448574969021\\
0.5	0.826270136307311\\
0.6	0.818091697645601\\
0.7	0.819578686493185\\
0.8	0.771499380421313\\
0.9	0.702106567534077\\
};
%\addlegendentry{Alg. 1 - TA + DS}

\addplot[area legend, draw=none, fill=mycolor2, fill opacity=0.2, forget plot]
table[row sep=crcr] {%
x	y\\
0	0.837670384138786\\
0.1	0.839419342277563\\
0.2	0.546463185026702\\
0.3	0.812786802317249\\
0.4	0.823413810725704\\
0.5	0.112415343632257\\
0.6	0.0964470016655234\\
0.7	0.0489076354796273\\
0.8	0.111833982165033\\
0.9	0.113463747023014\\
0.9	0.373772932035227\\
0.8	0.243060689458263\\
0.7	0.351092364520373\\
0.6	0.381867744307215\\
0.5	0.260323194162043\\
0.4	0.845483339212339\\
0.3	0.838266481449789\\
0.2	0.882037434551984\\
0.1	0.867643854748459\\
0	0.837670384138786\\
}--cycle;
\addplot [color=white!55!mycolor2, forget plot]
  table[row sep=crcr]{%
0	0.837670384138786\\
0.1	0.839419342277563\\
0.2	0.546463185026702\\
0.3	0.812786802317249\\
0.4	0.823413810725704\\
0.5	0.112415343632257\\
0.6	0.0964470016655234\\
0.7	0.0489076354796273\\
0.8	0.111833982165033\\
0.9	0.113463747023014\\
};
\addplot [color=white!55!mycolor2, forget plot]
  table[row sep=crcr]{%
0	0.837670384138786\\
0.1	0.867643854748459\\
0.2	0.882037434551984\\
0.3	0.838266481449789\\
0.4	0.845483339212339\\
0.5	0.260323194162043\\
0.6	0.381867744307215\\
0.7	0.351092364520373\\
0.8	0.243060689458263\\
0.9	0.373772932035227\\
};
\addplot [color=mycolor2, line width=1.0pt, mark size=8.6pt, mark=diamond, mark options={solid, mycolor2}]
  table[row sep=crcr]{%
0	0.837670384138786\\
0.1	0.853531598513011\\
0.2	0.714250309789343\\
0.3	0.825526641883519\\
0.4	0.834448574969021\\
0.5	0.18636926889715\\
0.6	0.239157372986369\\
0.7	0.2\\
0.8	0.177447335811648\\
0.9	0.24361833952912\\
};
%\addlegendentry{Alg. 1 - H + DS}

\addplot[area legend, draw=none, fill=mycolor3, fill opacity=0.2, forget plot]
table[row sep=crcr] {%
x	y\\
0	0.827757125154895\\
0.1	0.799736687117279\\
0.2	0.767651028045597\\
0.3	0.62385255294873\\
0.4	0.481746189076892\\
0.5	0.139834579695111\\
0.6	0.217206968508376\\
0.7	0.30127320374363\\
0.8	0.387109881949142\\
0.9	0.412094274698378\\
0.9	0.436976357271882\\
0.8	0.408429151508107\\
0.7	0.389185284484375\\
0.6	0.26953404760067\\
0.5	0.174415730094232\\
0.4	0.630521468915673\\
0.3	0.720881028216078\\
0.2	0.809548476291454\\
0.1	0.8295074268852\\
0	0.827757125154895\\
}--cycle;
\addplot [color=white!55!mycolor3, forget plot]
  table[row sep=crcr]{%
0	0.827757125154895\\
0.1	0.799736687117279\\
0.2	0.767651028045597\\
0.3	0.62385255294873\\
0.4	0.481746189076892\\
0.5	0.139834579695111\\
0.6	0.217206968508376\\
0.7	0.30127320374363\\
0.8	0.387109881949142\\
0.9	0.412094274698378\\
};
\addplot [color=white!55!mycolor3, forget plot]
  table[row sep=crcr]{%
0	0.827757125154895\\
0.1	0.8295074268852\\
0.2	0.809548476291454\\
0.3	0.720881028216078\\
0.4	0.630521468915673\\
0.5	0.174415730094232\\
0.6	0.26953404760067\\
0.7	0.389185284484375\\
0.8	0.408429151508107\\
0.9	0.436976357271882\\
};
\addplot [color=mycolor3, dashed, line width=1.0pt, mark size=5.0pt, mark=x, mark options={solid, mycolor3}]
  table[row sep=crcr]{%
0	0.827757125154895\\
0.1	0.814622057001239\\
0.2	0.788599752168525\\
0.3	0.672366790582404\\
0.4	0.556133828996282\\
0.5	0.157125154894672\\
0.6	0.243370508054523\\
0.7	0.345229244114002\\
0.8	0.397769516728625\\
0.9	0.42453531598513\\
};
%\addlegendentry{MMSR}

\addplot[area legend, draw=none, fill=mycolor4, fill opacity=0.2, forget plot]
table[row sep=crcr] {%
x	y\\
0	0.833952912019826\\
0.1	0.839280785009482\\
0.2	0.575863061575058\\
0.3	0.236154635759709\\
0.4	0.229039525711699\\
0.5	0.225447320245594\\
0.6	0.241682425039302\\
0.7	0.35009953783217\\
0.8	0.112231905027642\\
0.9	0.114235030143551\\
0.9	0.371514660067106\\
0.8	0.247123733138405\\
0.7	0.44444816972669\\
0.6	0.421018938033808\\
0.5	0.373809185330614\\
0.4	0.383104216543568\\
0.3	0.361119218019721\\
0.2	0.92698700038281\\
0.1	0.869765063813319\\
0	0.833952912019826\\
}--cycle;
\addplot [color=white!55!mycolor4, forget plot]
  table[row sep=crcr]{%
0	0.833952912019826\\
0.1	0.839280785009482\\
0.2	0.575863061575058\\
0.3	0.236154635759709\\
0.4	0.229039525711699\\
0.5	0.225447320245594\\
0.6	0.241682425039302\\
0.7	0.35009953783217\\
0.8	0.112231905027642\\
0.9	0.114235030143551\\
};
\addplot [color=white!55!mycolor4, forget plot]
  table[row sep=crcr]{%
0	0.833952912019826\\
0.1	0.869765063813319\\
0.2	0.92698700038281\\
0.3	0.361119218019721\\
0.4	0.383104216543568\\
0.5	0.373809185330614\\
0.6	0.421018938033808\\
0.7	0.44444816972669\\
0.8	0.247123733138405\\
0.9	0.371514660067106\\
};
\addplot [color=mycolor4, dashdotted, line width=1.0pt, mark size=3.5pt, mark=square, mark options={solid, mycolor4}]
  table[row sep=crcr]{%
0	0.833952912019826\\
0.1	0.8545229244114\\
0.2	0.751425030978934\\
0.3	0.298636926889715\\
0.4	0.306071871127633\\
0.5	0.299628252788104\\
0.6	0.331350681536555\\
0.7	0.39727385377943\\
0.8	0.179677819083024\\
0.9	0.242874845105328\\
};
%\addlegendentry{DS}

\addplot[area legend, draw=none, fill=mycolor5, fill opacity=0.2, forget plot]
table[row sep=crcr] {%
x	y\\
0	0.81908302354399\\
0.1	0.772779323187479\\
0.2	0.608180821803557\\
0.3	0.432572101067047\\
0.4	0.432162332096161\\
0.5	0.431193918704803\\
0.6	0.434460630477201\\
0.7	0.437033761577475\\
0.8	0.429275190355799\\
0.9	0.430119016065041\\
0.9	0.452656696450448\\
0.8	0.439126296633049\\
0.7	0.454168221074321\\
0.6	0.452280385631845\\
0.5	0.447120827267935\\
0.4	0.447639402724162\\
0.3	0.485395680841255\\
0.2	0.646837765557038\\
0.1	0.808881147692322\\
0	0.81908302354399\\
}--cycle;
\addplot [color=white!55!mycolor5, forget plot]
  table[row sep=crcr]{%
0	0.81908302354399\\
0.1	0.772779323187479\\
0.2	0.608180821803557\\
0.3	0.432572101067047\\
0.4	0.432162332096161\\
0.5	0.431193918704803\\
0.6	0.434460630477201\\
0.7	0.437033761577475\\
0.8	0.429275190355799\\
0.9	0.430119016065041\\
};
\addplot [color=white!55!mycolor5, forget plot]
  table[row sep=crcr]{%
0	0.81908302354399\\
0.1	0.808881147692322\\
0.2	0.646837765557038\\
0.3	0.485395680841255\\
0.4	0.447639402724162\\
0.5	0.447120827267935\\
0.6	0.452280385631845\\
0.7	0.454168221074321\\
0.8	0.439126296633049\\
0.9	0.452656696450448\\
};
\addplot [color=mycolor5, dashdotted, line width=1.0pt, mark size=5.0pt, mark=+, mark options={solid, mycolor5}]
  table[row sep=crcr]{%
0	0.81908302354399\\
0.1	0.790830235439901\\
0.2	0.627509293680297\\
0.3	0.458983890954151\\
0.4	0.439900867410161\\
0.5	0.439157372986369\\
0.6	0.443370508054523\\
0.7	0.445600991325898\\
0.8	0.434200743494424\\
0.9	0.441387856257745\\
};
%\addlegendentry{MV}

\addplot[area legend, draw=none, fill=mycolor6, fill opacity=0.2, forget plot]
table[row sep=crcr] {%
x	y\\
0	0.830235439900867\\
0.1	0.817887916109258\\
0.2	0.777772394313492\\
0.3	0.573149510841615\\
0.4	0.447786250593418\\
0.5	0.42983007630073\\
0.6	0.423795466930376\\
0.7	0.415618407887435\\
0.8	0.410161090458488\\
0.9	0.420215068146327\\
0.9	0.446699429994937\\
0.8	0.42998760842627\\
0.7	0.43394788703202\\
0.6	0.440640716464915\\
0.5	0.444023703129257\\
0.4	0.486538408638305\\
0.3	0.717556808860987\\
0.2	0.82024495388973\\
0.1	0.829695727013418\\
0	0.830235439900867\\
}--cycle;
\addplot [color=white!55!mycolor6, forget plot]
  table[row sep=crcr]{%
0	0.830235439900867\\
0.1	0.817887916109258\\
0.2	0.777772394313492\\
0.3	0.573149510841615\\
0.4	0.447786250593418\\
0.5	0.42983007630073\\
0.6	0.423795466930376\\
0.7	0.415618407887435\\
0.8	0.410161090458488\\
0.9	0.420215068146327\\
};
\addplot [color=white!55!mycolor6, forget plot]
  table[row sep=crcr]{%
0	0.830235439900867\\
0.1	0.829695727013418\\
0.2	0.82024495388973\\
0.3	0.717556808860987\\
0.4	0.486538408638305\\
0.5	0.444023703129257\\
0.6	0.440640716464915\\
0.7	0.43394788703202\\
0.8	0.42998760842627\\
0.9	0.446699429994937\\
};
\addplot [color=mycolor6, dashdotted, line width=1.0pt, mark size=3.3pt, mark=triangle, mark options={solid, mycolor6}]
  table[row sep=crcr]{%
0	0.830235439900867\\
0.1	0.823791821561338\\
0.2	0.799008674101611\\
0.3	0.645353159851301\\
0.4	0.467162329615861\\
0.5	0.436926889714994\\
0.6	0.432218091697646\\
0.7	0.424783147459727\\
0.8	0.420074349442379\\
0.9	0.433457249070632\\
};
%\addlegendentry{PGD}

\end{axis}
\end{tikzpicture}%

%% file: fig/tikz/real_data/per_adv_DS_web_shaded_acc.tikz
\definecolor{mycolor1}{rgb}{0.00000,0.44700,0.74100}%
\definecolor{mycolor2}{rgb}{0.49400,0.18400,0.55600}%
\definecolor{mycolor3}{rgb}{0.63500,0.07800,0.18400}%
\definecolor{mycolor4}{rgb}{0.92900,0.69400,0.12500}%
\definecolor{mycolor5}{rgb}{0.30100,0.74500,0.93300}%
\definecolor{mycolor6}{rgb}{0.85000,0.32500,0.09800}%
\begin{tikzpicture}

\begin{axis}[%
width=4.464in,
height=3.286in,
at={(0.815in,0.76in)},
scale only axis,
xmin=0,
xmax=0.95,
xlabel style={font=\color{white!15!black}},
xlabel={Percentage of adversaries},
ymin=0,
ymax=1,
ylabel style={font=\color{white!15!black}},
ylabel={Accuracy},
axis background/.style={fill=white},
axis x line*=bottom,
axis y line*=left,
]

\addplot[area legend, draw=none, fill=mycolor1, fill opacity=0.2, forget plot]
table[row sep=crcr] {%
x	y\\
0	0.822345890410959\\
0.1	0.119139457772914\\
0.2	0.361475778322487\\
0.3	0.483106511585744\\
0.4	0.20350443167269\\
0.5	0.365350121432343\\
0.6	0.647382829365663\\
0.7	0.728516215360938\\
0.8	0.108505616235399\\
0.9	0.138292969777226\\
0.9	0.353226065277429\\
0.8	0.708230154590082\\
0.7	0.868016012305854\\
0.6	0.831697457102487\\
0.5	0.948332502012814\\
0.4	0.795817091131683\\
0.3	0.98330886723069\\
0.2	1.08694737819328\\
0.1	0.212631551070316\\
0	0.822345890410959\\
}--cycle;
\addplot [color=white!55!mycolor1, forget plot]
  table[row sep=crcr]{%
0	0.822345890410959\\
0.1	0.119139457772914\\
0.2	0.361475778322487\\
0.3	0.483106511585744\\
0.4	0.20350443167269\\
0.5	0.365350121432343\\
0.6	0.647382829365663\\
0.7	0.728516215360938\\
0.8	0.108505616235399\\
0.9	0.138292969777226\\
};
\addplot [color=white!55!mycolor1, forget plot]
  table[row sep=crcr]{%
0	0.822345890410959\\
0.1	0.212631551070316\\
0.2	1.08694737819328\\
0.3	0.98330886723069\\
0.4	0.795817091131683\\
0.5	0.948332502012814\\
0.6	0.831697457102487\\
0.7	0.868016012305854\\
0.8	0.708230154590082\\
0.9	0.353226065277429\\
};
\addplot [color=mycolor1, line width=1.0pt, mark size=5.0pt, mark=o, mark options={solid, mycolor1}]
  table[row sep=crcr]{%
0	0.822345890410959\\
0.1	0.165885504421615\\
0.2	0.724211578257882\\
0.3	0.733207689408217\\
0.4	0.499660761402186\\
0.5	0.656841311722578\\
0.6	0.739540143234075\\
0.7	0.798266113833396\\
0.8	0.40836788541274\\
0.9	0.245759517527328\\
};
%\addlegendentry{Alg. 1 - TA + DS}

\addplot[area legend, draw=none, fill=mycolor2, fill opacity=0.2, forget plot]
table[row sep=crcr] {%
x	y\\
0	0.822345890410959\\
0.1	0.12866266128504\\
0.2	0.361475778322487\\
0.3	0.483106511585744\\
0.4	0.396264490684183\\
0.5	0.121028504733394\\
0.6	0.0926992835862854\\
0.7	0.0779805083531809\\
0.8	0.109278643824735\\
0.9	0.114049962074847\\
0.9	0.280823765780411\\
0.8	0.22739681791669\\
0.7	0.169286736275541\\
0.6	0.224978816677567\\
0.5	0.2075052306605\\
0.4	0.878669546255131\\
0.3	0.98330886723069\\
0.2	1.08694737819328\\
0.1	0.334331468576661\\
0	0.822345890410959\\
}--cycle;
\addplot [color=white!55!mycolor2, forget plot]
  table[row sep=crcr]{%
0	0.822345890410959\\
0.1	0.12866266128504\\
0.2	0.361475778322487\\
0.3	0.483106511585744\\
0.4	0.396264490684183\\
0.5	0.121028504733394\\
0.6	0.0926992835862854\\
0.7	0.0779805083531809\\
0.8	0.109278643824735\\
0.9	0.114049962074847\\
};
\addplot [color=white!55!mycolor2, forget plot]
  table[row sep=crcr]{%
0	0.822345890410959\\
0.1	0.334331468576661\\
0.2	1.08694737819328\\
0.3	0.98330886723069\\
0.4	0.878669546255131\\
0.5	0.2075052306605\\
0.6	0.224978816677567\\
0.7	0.169286736275541\\
0.8	0.22739681791669\\
0.9	0.280823765780411\\
};
\addplot [color=mycolor2, line width=1.0pt, mark size=8.6pt, mark=diamond, mark options={solid, mycolor2}]
  table[row sep=crcr]{%
0	0.822345890410959\\
0.1	0.23149706493085\\
0.2	0.724211578257882\\
0.3	0.733207689408217\\
0.4	0.637467018469657\\
0.5	0.164266867696947\\
0.6	0.158839050131926\\
0.7	0.123633622314361\\
0.8	0.168337730870712\\
0.9	0.197436863927629\\
};
%\addlegendentry{Alg. 1 - H + DS}

\addplot[area legend, draw=none, fill=mycolor3, fill opacity=0.2, forget plot]
table[row sep=crcr] {%
x	y\\
0	0.690924657534247\\
0.1	0.486986903303558\\
0.2	0.497063747946258\\
0.3	0.403858382137848\\
0.4	0.233171168048748\\
0.5	0.291917572825139\\
0.6	0.34369721786355\\
0.7	0.390107102792667\\
0.8	0.411399376273073\\
0.9	0.429196320261814\\
0.9	0.441817626213874\\
0.8	0.426444573971933\\
0.7	0.410345215337752\\
0.6	0.382273381457973\\
0.5	0.329115220239316\\
0.4	0.3786644896972\\
0.3	0.496707015525175\\
0.2	0.528272266397867\\
0.1	0.633645950658103\\
0	0.690924657534247\\
}--cycle;
\addplot [color=white!55!mycolor3, forget plot]
  table[row sep=crcr]{%
0	0.690924657534247\\
0.1	0.486986903303558\\
0.2	0.497063747946258\\
0.3	0.403858382137848\\
0.4	0.233171168048748\\
0.5	0.291917572825139\\
0.6	0.34369721786355\\
0.7	0.390107102792667\\
0.8	0.411399376273073\\
0.9	0.429196320261814\\
};
\addplot [color=white!55!mycolor3, forget plot]
  table[row sep=crcr]{%
0	0.690924657534247\\
0.1	0.633645950658103\\
0.2	0.528272266397867\\
0.3	0.496707015525175\\
0.4	0.3786644896972\\
0.5	0.329115220239316\\
0.6	0.382273381457973\\
0.7	0.410345215337752\\
0.8	0.426444573971933\\
0.9	0.441817626213874\\
};
\addplot [color=mycolor3, dashed, line width=1.0pt, mark size=5.0pt, mark=x, mark options={solid, mycolor3}]
  table[row sep=crcr]{%
0	0.690924657534247\\
0.1	0.560316426980831\\
0.2	0.512668007172062\\
0.3	0.450282698831512\\
0.4	0.305917828872974\\
0.5	0.310516396532228\\
0.6	0.362985299660761\\
0.7	0.400226159065209\\
0.8	0.418921975122503\\
0.9	0.435506973237844\\
};
%\addlegendentry{MMSR}

\addplot[area legend, draw=none, fill=mycolor4, fill opacity=0.2, forget plot]
table[row sep=crcr] {%
x	y\\
0	0.835188356164384\\
0.1	0.137691088540391\\
0.2	0.142383030309813\\
0.3	0.0778438092302944\\
0.4	0.106049089417235\\
0.5	0.121150896505105\\
0.6	0.0926823491468137\\
0.7	0.0781438949556737\\
0.8	0.109624757690678\\
0.9	0.113426982909084\\
0.9	0.28280369933743\\
0.8	0.228106113021723\\
0.7	0.16882180425277\\
0.6	0.226955796348852\\
0.5	0.20994597496116\\
0.4	0.156747744355852\\
0.3	0.188270024165861\\
0.2	0.204630335532247\\
0.1	0.229811012560074\\
0	0.835188356164384\\
}--cycle;
\addplot [color=white!55!mycolor4, forget plot]
  table[row sep=crcr]{%
0	0.835188356164384\\
0.1	0.137691088540391\\
0.2	0.142383030309813\\
0.3	0.0778438092302944\\
0.4	0.106049089417235\\
0.5	0.121150896505105\\
0.6	0.0926823491468137\\
0.7	0.0781438949556737\\
0.8	0.109624757690678\\
0.9	0.113426982909084\\
};
\addplot [color=white!55!mycolor4, forget plot]
  table[row sep=crcr]{%
0	0.835188356164384\\
0.1	0.229811012560074\\
0.2	0.204630335532247\\
0.3	0.188270024165861\\
0.4	0.156747744355852\\
0.5	0.20994597496116\\
0.6	0.226955796348852\\
0.7	0.16882180425277\\
0.8	0.228106113021723\\
0.9	0.28280369933743\\
};
\addplot [color=mycolor4, dashdotted, line width=1.0pt, mark size=3.5pt, mark=square, mark options={solid, mycolor4}]
  table[row sep=crcr]{%
0	0.835188356164384\\
0.1	0.183751050550233\\
0.2	0.17350668292103\\
0.3	0.133056916698078\\
0.4	0.131398416886544\\
0.5	0.165548435733132\\
0.6	0.159819072747833\\
0.7	0.123482849604222\\
0.8	0.168865435356201\\
0.9	0.198115341123257\\
};
%\addlegendentry{DS}

\addplot[area legend, draw=none, fill=mycolor5, fill opacity=0.2, forget plot]
table[row sep=crcr] {%
x	y\\
0	0.787671232876712\\
0.1	0.440122082794416\\
0.2	0.388703027407898\\
0.3	0.389083448635301\\
0.4	0.393381670652209\\
0.5	0.399544196261147\\
0.6	0.411691741175102\\
0.7	0.426999260599775\\
0.8	0.422163934206746\\
0.9	0.435884817033888\\
0.9	0.445230901021144\\
0.8	0.438597467979458\\
0.7	0.43918995915145\\
0.6	0.428263027011856\\
0.5	0.425634846332144\\
0.4	0.426369554376061\\
0.3	0.421470641074462\\
0.2	0.406200082293542\\
0.1	0.488890826628899\\
0	0.787671232876712\\
}--cycle;
\addplot [color=white!55!mycolor5, forget plot]
  table[row sep=crcr]{%
0	0.787671232876712\\
0.1	0.440122082794416\\
0.2	0.388703027407898\\
0.3	0.389083448635301\\
0.4	0.393381670652209\\
0.5	0.399544196261147\\
0.6	0.411691741175102\\
0.7	0.426999260599775\\
0.8	0.422163934206746\\
0.9	0.435884817033888\\
};
\addplot [color=white!55!mycolor5, forget plot]
  table[row sep=crcr]{%
0	0.787671232876712\\
0.1	0.488890826628899\\
0.2	0.406200082293542\\
0.3	0.421470641074462\\
0.4	0.426369554376061\\
0.5	0.425634846332144\\
0.6	0.428263027011856\\
0.7	0.43918995915145\\
0.8	0.438597467979458\\
0.9	0.445230901021144\\
};
\addplot [color=mycolor5, dashdotted, line width=1.0pt, mark size=5.0pt, mark=+, mark options={solid, mycolor5}]
  table[row sep=crcr]{%
0	0.787671232876712\\
0.1	0.464506454711657\\
0.2	0.39745155485072\\
0.3	0.405277044854881\\
0.4	0.409875612514135\\
0.5	0.412589521296645\\
0.6	0.419977384093479\\
0.7	0.433094609875612\\
0.8	0.430380701093102\\
0.9	0.440557859027516\\
};
%\addlegendentry{MV}

\addplot[area legend, draw=none, fill=mycolor6, fill opacity=0.2, forget plot]
table[row sep=crcr] {%
x	y\\
0	0.899400684931507\\
0.1	0.420698386284605\\
0.2	0.350274558454768\\
0.3	0.353692347068552\\
0.4	0.368684591581651\\
0.5	0.378919851627157\\
0.6	0.404872030009934\\
0.7	0.408560688071928\\
0.8	0.421104805038263\\
0.9	0.43384007518798\\
0.9	0.44501405221496\\
0.8	0.431213325380132\\
0.7	0.424609308158755\\
0.6	0.409149832033036\\
0.5	0.39936887811276\\
0.4	0.386385140796788\\
0.3	0.374992161035481\\
0.2	0.370585445823955\\
0.1	0.487032281690831\\
0	0.899400684931507\\
}--cycle;
\addplot [color=white!55!mycolor6, forget plot]
  table[row sep=crcr]{%
0	0.899400684931507\\
0.1	0.420698386284605\\
0.2	0.350274558454768\\
0.3	0.353692347068552\\
0.4	0.368684591581651\\
0.5	0.378919851627157\\
0.6	0.404872030009934\\
0.7	0.408560688071928\\
0.8	0.421104805038263\\
0.9	0.43384007518798\\
};
\addplot [color=white!55!mycolor6, forget plot]
  table[row sep=crcr]{%
0	0.899400684931507\\
0.1	0.487032281690831\\
0.2	0.370585445823955\\
0.3	0.374992161035481\\
0.4	0.386385140796788\\
0.5	0.39936887811276\\
0.6	0.409149832033036\\
0.7	0.424609308158755\\
0.8	0.431213325380132\\
0.9	0.44501405221496\\
};
\addplot [color=mycolor6, dashdotted, line width=1.0pt, mark size=3.3pt, mark=triangle, mark options={solid, mycolor6}]
  table[row sep=crcr]{%
0	0.899400684931507\\
0.1	0.453865333987718\\
0.2	0.360430002139361\\
0.3	0.364342254052017\\
0.4	0.37753486618922\\
0.5	0.389144364869959\\
0.6	0.407010931021485\\
0.7	0.416584998115341\\
0.8	0.426159065209197\\
0.9	0.43942706370147\\
};
%\addlegendentry{PGD}

\end{axis}
\end{tikzpicture}%

%% file: fig/tikz/real_data/per_adv_DS_Adult2_shaded_acc.tikz
\definecolor{mycolor1}{rgb}{0.00000,0.44700,0.74100}%
\definecolor{mycolor2}{rgb}{0.49400,0.18400,0.55600}%
\definecolor{mycolor3}{rgb}{0.63500,0.07800,0.18400}%
\definecolor{mycolor4}{rgb}{0.92900,0.69400,0.12500}%
\definecolor{mycolor5}{rgb}{0.30100,0.74500,0.93300}%
\definecolor{mycolor6}{rgb}{0.85000,0.32500,0.09800}%
\begin{tikzpicture}

\begin{axis}[%
width=4.464in,
height=3.286in,
at={(0.815in,0.76in)},
scale only axis,
xmin=0,
xmax=0.95,
xlabel style={font=\color{white!15!black}},
xlabel={Percentage of adversaries},
ymin=0.2,
ymax=0.825506083832651,
ylabel style={font=\color{white!15!black}},
ylabel={Accuracy},
axis background/.style={fill=white},
axis x line*=bottom,
axis y line*=left,
]

\addplot[area legend, draw=none, fill=mycolor1, fill opacity=0.2, forget plot]
table[row sep=crcr] {%
x	y\\
0	0.774509803921569\\
0.1	0.499289918170808\\
0.2	0.714268585630176\\
0.3	0.725995770082979\\
0.4	0.734433911370461\\
0.5	0.722408575425818\\
0.6	0.743962880937579\\
0.7	0.696006602838557\\
0.8	0.690012922032418\\
0.9	0.634121577219435\\
0.9	0.724666301568444\\
0.8	0.7681688961494\\
0.7	0.782781275949322\\
0.6	0.786946209971512\\
0.5	0.757591424574182\\
0.4	0.762535785599236\\
0.3	0.768549684462475\\
0.2	0.737535780216743\\
0.1	0.825506083832651\\
0	0.774509803921569\\
}--cycle;
\addplot [color=white!55!mycolor1, forget plot]
  table[row sep=crcr]{%
0	0.774509803921569\\
0.1	0.499289918170808\\
0.2	0.714268585630176\\
0.3	0.725995770082979\\
0.4	0.734433911370461\\
0.5	0.722408575425818\\
0.6	0.743962880937579\\
0.7	0.696006602838557\\
0.8	0.690012922032418\\
0.9	0.634121577219435\\
};
\addplot [color=white!55!mycolor1, forget plot]
  table[row sep=crcr]{%
0	0.774509803921569\\
0.1	0.825506083832651\\
0.2	0.737535780216743\\
0.3	0.768549684462475\\
0.4	0.762535785599236\\
0.5	0.757591424574182\\
0.6	0.786946209971512\\
0.7	0.782781275949322\\
0.8	0.7681688961494\\
0.9	0.724666301568444\\
};
\addplot [color=mycolor1, line width=1.0pt, mark size=5.0pt, mark=o, mark options={solid, mycolor1}]
  table[row sep=crcr]{%
0	0.774509803921569\\
0.1	0.66239800100173\\
0.2	0.72590218292346\\
0.3	0.747272727272727\\
0.4	0.748484848484848\\
0.5	0.74\\
0.6	0.765454545454545\\
0.7	0.739393939393939\\
0.8	0.729090909090909\\
0.9	0.679393939393939\\
};
%\addlegendentry{Alg. 1 - TA + DS}

\addplot[area legend, draw=none, fill=mycolor2, fill opacity=0.2, forget plot]
table[row sep=crcr] {%
x	y\\
0	0.774509803921569\\
0.1	0.726293116778147\\
0.2	0.714268585630176\\
0.3	0.725995770082979\\
0.4	0.734433911370461\\
0.5	0.390068183195092\\
0.6	0.381813294550479\\
0.7	0.416959817297349\\
0.8	0.401379993209959\\
0.9	0.442715621161807\\
0.9	0.482132863686678\\
0.8	0.470135158305192\\
0.7	0.472737152399621\\
0.6	0.4569745842374\\
0.5	0.43174999862309\\
0.4	0.762535785599236\\
0.3	0.768549684462475\\
0.2	0.737535780216743\\
0.1	0.748196136759055\\
0	0.774509803921569\\
}--cycle;
\addplot [color=white!55!mycolor2, forget plot]
  table[row sep=crcr]{%
0	0.774509803921569\\
0.1	0.726293116778147\\
0.2	0.714268585630176\\
0.3	0.725995770082979\\
0.4	0.734433911370461\\
0.5	0.390068183195092\\
0.6	0.381813294550479\\
0.7	0.416959817297349\\
0.8	0.401379993209959\\
0.9	0.442715621161807\\
};
\addplot [color=white!55!mycolor2, forget plot]
  table[row sep=crcr]{%
0	0.774509803921569\\
0.1	0.748196136759055\\
0.2	0.737535780216743\\
0.3	0.768549684462475\\
0.4	0.762535785599236\\
0.5	0.43174999862309\\
0.6	0.4569745842374\\
0.7	0.472737152399621\\
0.8	0.470135158305192\\
0.9	0.482132863686678\\
};
\addplot [color=mycolor2, line width=1.0pt, mark size=8.6pt, mark=diamond, mark options={solid, mycolor2}]
  table[row sep=crcr]{%
0	0.774509803921569\\
0.1	0.737244626768601\\
0.2	0.72590218292346\\
0.3	0.747272727272727\\
0.4	0.748484848484848\\
0.5	0.410909090909091\\
0.6	0.419393939393939\\
0.7	0.444848484848485\\
0.8	0.435757575757576\\
0.9	0.462424242424242\\
};
%\addlegendentry{Alg. 1 - H + DS}

\addplot[area legend, draw=none, fill=mycolor3, fill opacity=0.2, forget plot]
table[row sep=crcr] {%
x	y\\
0	0.644024874911163\\
0.1	0.717663525694525\\
0.2	0.667545075359191\\
0.3	0.612184398331179\\
0.4	0.476773363296761\\
0.5	0.240127678603756\\
0.6	0.314760898788917\\
0.7	0.384655866022257\\
0.8	0.421794283172933\\
0.9	0.447916765826477\\
0.9	0.462386264476553\\
0.8	0.444872383493733\\
0.7	0.40201080064441\\
0.6	0.37251182848381\\
0.5	0.31623595775988\\
0.4	0.537772091248694\\
0.3	0.684785298638518\\
0.2	0.728498030471149\\
0.1	0.74948655710819\\
0	0.735060092409098\\
}--cycle;
\addplot [color=white!55!mycolor3, forget plot]
  table[row sep=crcr]{%
0	0.644024874911163\\
0.1	0.717663525694525\\
0.2	0.667545075359191\\
0.3	0.612184398331179\\
0.4	0.476773363296761\\
0.5	0.240127678603756\\
0.6	0.314760898788917\\
0.7	0.384655866022257\\
0.8	0.421794283172933\\
0.9	0.447916765826477\\
};
\addplot [color=white!55!mycolor3, forget plot]
  table[row sep=crcr]{%
0	0.735060092409098\\
0.1	0.74948655710819\\
0.2	0.728498030471149\\
0.3	0.684785298638518\\
0.4	0.537772091248694\\
0.5	0.31623595775988\\
0.6	0.37251182848381\\
0.7	0.40201080064441\\
0.8	0.444872383493733\\
0.9	0.462386264476553\\
};
\addplot [color=mycolor3, dashed, line width=1.0pt, mark size=5.0pt, mark=x, mark options={solid, mycolor3}]
  table[row sep=crcr]{%
0	0.689542483660131\\
0.1	0.733575041401357\\
0.2	0.69802155291517\\
0.3	0.648484848484848\\
0.4	0.507272727272727\\
0.5	0.278181818181818\\
0.6	0.343636363636364\\
0.7	0.393333333333333\\
0.8	0.433333333333333\\
0.9	0.455151515151515\\
};
%\addlegendentry{MMSR}

\addplot[area legend, draw=none, fill=mycolor4, fill opacity=0.2, forget plot]
table[row sep=crcr] {%
x	y\\
0	0.774509803921569\\
0.1	0.672510772023388\\
0.2	0.401375351654846\\
0.3	0.414929333172521\\
0.4	0.411629325602328\\
0.5	0.408164988934352\\
0.6	0.438525565611089\\
0.7	0.429391971747663\\
0.8	0.434917498257419\\
0.9	0.437270651820068\\
0.9	0.481517226967811\\
0.8	0.468112804772884\\
0.7	0.476062573706883\\
0.6	0.463292616207092\\
0.5	0.457289556520193\\
0.4	0.435643401670399\\
0.3	0.425070666827479\\
0.2	0.463416027179086\\
0.1	0.737867241783706\\
0	0.774509803921569\\
}--cycle;
\addplot [color=white!55!mycolor4, forget plot]
  table[row sep=crcr]{%
0	0.774509803921569\\
0.1	0.672510772023388\\
0.2	0.401375351654846\\
0.3	0.414929333172521\\
0.4	0.411629325602328\\
0.5	0.408164988934352\\
0.6	0.438525565611089\\
0.7	0.429391971747663\\
0.8	0.434917498257419\\
0.9	0.437270651820068\\
};
\addplot [color=white!55!mycolor4, forget plot]
  table[row sep=crcr]{%
0	0.774509803921569\\
0.1	0.737867241783706\\
0.2	0.463416027179086\\
0.3	0.425070666827479\\
0.4	0.435643401670399\\
0.5	0.457289556520193\\
0.6	0.463292616207092\\
0.7	0.476062573706883\\
0.8	0.468112804772884\\
0.9	0.481517226967811\\
};
\addplot [color=mycolor4, dashdotted, line width=1.0pt, mark size=3.5pt, mark=square, mark options={solid, mycolor4}]
  table[row sep=crcr]{%
0	0.774509803921569\\
0.1	0.705189006903547\\
0.2	0.432395689416966\\
0.3	0.42\\
0.4	0.423636363636364\\
0.5	0.432727272727273\\
0.6	0.450909090909091\\
0.7	0.452727272727273\\
0.8	0.451515151515151\\
0.9	0.459393939393939\\
};
%\addlegendentry{DS}

\addplot[area legend, draw=none, fill=mycolor5, fill opacity=0.2, forget plot]
table[row sep=crcr] {%
x	y\\
0	0.76797385620915\\
0.1	0.661631346986077\\
0.2	0.470082786941512\\
0.3	0.424627323193311\\
0.4	0.413857553471337\\
0.5	0.433294098663364\\
0.6	0.453352291258987\\
0.7	0.452946386154337\\
0.8	0.454801588038113\\
0.9	0.460997488807918\\
0.9	0.472335844525415\\
0.8	0.472471139234614\\
0.7	0.480386947178996\\
0.6	0.472708314801619\\
0.5	0.466099840730575\\
0.4	0.44311214349836\\
0.3	0.448099949533962\\
0.2	0.518396535154832\\
0.1	0.688463583824241\\
0	0.76797385620915\\
}--cycle;
\addplot [color=white!55!mycolor5, forget plot]
  table[row sep=crcr]{%
0	0.76797385620915\\
0.1	0.661631346986077\\
0.2	0.470082786941512\\
0.3	0.424627323193311\\
0.4	0.413857553471337\\
0.5	0.433294098663364\\
0.6	0.453352291258987\\
0.7	0.452946386154337\\
0.8	0.454801588038113\\
0.9	0.460997488807918\\
};
\addplot [color=white!55!mycolor5, forget plot]
  table[row sep=crcr]{%
0	0.76797385620915\\
0.1	0.688463583824241\\
0.2	0.518396535154832\\
0.3	0.448099949533962\\
0.4	0.44311214349836\\
0.5	0.466099840730575\\
0.6	0.472708314801619\\
0.7	0.480386947178996\\
0.8	0.472471139234614\\
0.9	0.472335844525415\\
};
\addplot [color=mycolor5, dashdotted, line width=1.0pt, mark size=5.0pt, mark=+, mark options={solid, mycolor5}]
  table[row sep=crcr]{%
0	0.76797385620915\\
0.1	0.675047465405159\\
0.2	0.494239661048172\\
0.3	0.436363636363636\\
0.4	0.428484848484848\\
0.5	0.44969696969697\\
0.6	0.463030303030303\\
0.7	0.466666666666667\\
0.8	0.463636363636364\\
0.9	0.466666666666667\\
};
%\addlegendentry{MV}

\addplot[area legend, draw=none, fill=mycolor6, fill opacity=0.2, forget plot]
table[row sep=crcr] {%
x	y\\
0	0.764705882352941\\
0.1	0.711714462329537\\
0.2	0.506963887107739\\
0.3	0.429390416262057\\
0.4	0.414774697993673\\
0.5	0.428703961911913\\
0.6	0.438645221780805\\
0.7	0.437086220415775\\
0.8	0.445155400794171\\
0.9	0.451868843718183\\
0.9	0.468131156281817\\
0.8	0.468783993145223\\
0.7	0.45745923412968\\
0.6	0.461960838825255\\
0.5	0.456144522936572\\
0.4	0.440982877763903\\
0.3	0.455458068586428\\
0.2	0.652575580516835\\
0.1	0.735651536589041\\
0	0.764705882352941\\
}--cycle;
\addplot [color=white!55!mycolor6, forget plot]
  table[row sep=crcr]{%
0	0.764705882352941\\
0.1	0.711714462329537\\
0.2	0.506963887107739\\
0.3	0.429390416262057\\
0.4	0.414774697993673\\
0.5	0.428703961911913\\
0.6	0.438645221780805\\
0.7	0.437086220415775\\
0.8	0.445155400794171\\
0.9	0.451868843718183\\
};
\addplot [color=white!55!mycolor6, forget plot]
  table[row sep=crcr]{%
0	0.764705882352941\\
0.1	0.735651536589041\\
0.2	0.652575580516835\\
0.3	0.455458068586428\\
0.4	0.440982877763903\\
0.5	0.456144522936572\\
0.6	0.461960838825255\\
0.7	0.45745923412968\\
0.8	0.468783993145223\\
0.9	0.468131156281817\\
};
\addplot [color=mycolor6, dashdotted, line width=1.0pt, mark size=3.3pt, mark=triangle, mark options={solid, mycolor6}]
  table[row sep=crcr]{%
0	0.764705882352941\\
0.1	0.723682999459289\\
0.2	0.579769733812287\\
0.3	0.442424242424242\\
0.4	0.427878787878788\\
0.5	0.442424242424242\\
0.6	0.45030303030303\\
0.7	0.447272727272727\\
0.8	0.456969696969697\\
0.9	0.46\\
};
%\addlegendentry{PGD}

\end{axis}
\end{tikzpicture}%